%% file: main.tex
\documentclass[11pt]{article}
\usepackage{microsoft-tech-report}

\usepackage[utf8]{inputenc} 
\usepackage[T1]{fontenc}    
\usepackage{hyperref}       
\hypersetup{colorlinks=true, urlcolor=msftblue, linkcolor=black,
            citecolor=black, filecolor=msftblue}
\usepackage{url}            
\usepackage{booktabs}       
\usepackage{amsfonts}       
\usepackage{nicefrac}       
\usepackage{microtype}      
\usepackage{xcolor}         
\usepackage{amssymb}
\usepackage{cleveref}
\usepackage{array}
\usepackage{colortbl}
\usepackage{caption}
\usepackage{subcaption}
\usepackage{multirow}
\usepackage{float}
\usepackage[inline]{enumitem}
\usepackage{xspace}
\usepackage{makecell}  
\usepackage{pgfplots}
\usepackage{tabularx}

\usepackage{figures/result-figure-style}
\usepgfplotslibrary{fillbetween,groupplots}
\pgfplotsset{compat=1.18}

\xspaceaddexceptions{+}

\def\RHO{Rho\xspace}
\def\Rho{rho}
\newcommand{\RHOB}{\RHO-base\xspace}
\newcommand{\RHOFR}{\RHO-FR3-Duo\xspace}
\newcommand{\RHOUAT}{\RHO-UR-AI-Trainer\xspace}
\newcommand{\RHOYAM}{\RHO-YAM-Box\xspace}
\newcommand{\RHORT}{\RHO-RoboTwin\xspace}
\newcommand{\thm}{ midtraining}
\newcommand{\Ethm}{Embodiment\thm\xspace}
\newcommand{\ethm}{embodiment\thm\xspace}
\newcommand{\backbonename}{Phi-Phy}
\newcommand{\phiphy}{\backbonename\xspace}

\def\GR{GR00T-N1.7\xspace}
\def\PI{$\pi_{0.5}$\xspace}
\def\MO{MolmoAct2\xspace}

\def\urait{UR AI Trainer\xspace}
\def\fr3d{FR3 Duo\xspace}
\def\yb{YAM Box\xspace}

\def\FD{FlowDAgger\xspace}

\newif\ifshownotes
\shownotesfalse      

\newif\ifshowstubs
\showstubstrue         

\ifshownotes
  \newcommand{\dontforget}[1]{\textcolor{red}{#1 $\blacksquare$\footnote{Don't forget to edit}}}
  \newcommand{\ak}[1]{\textcolor{orange}{AK: #1}}
\else
  \newcommand{\dontforget}[1]{}
  \newcommand{\ak}[1]{}
\fi

\newcommand{\reporttitle}{\RHO: A Foundation for Efficiently Adaptable VLA Models}
\newcommand{\reportdate}{September 2026}
\newcommand{\leadmark}[1]{\textsuperscript{#1}}

\newcommand{\reportauthors}{%
  Rho team\footnote{Authors are listed alphabetically. See \Cref{app:authors} for a description of each author's contributions.}\\
  Simran Bagaria, Daphne Chen, Dean Fortier, Jianlong Fu, \\ Michael Harrison,
  Tess Hellebrekers, Neel Joshi\leadmark{$\dagger$}, Andrey Kolobov\leadmark{$\spadesuit*$}, Dalton Moore, \\
  Galen Mullins\leadmark{$\ddagger$}, Michael Murray\leadmark{$\S$}, Eduardo Salinas, Reuben Tan\\[4pt]
  {\scshape\small\color{msftgray}Microsoft Research}\\[\baselineskip]
  {\footnotesize\color{msftgray}%
    Area leads: \leadmark{$*$}Data \quad \leadmark{$\S$}Training \quad \leadmark{$\ddagger$}Engineering \quad \leadmark{$\dagger$}VLM \\
    \leadmark{$\spadesuit$}Project lead
  }
}

\title{\reporttitle}
\author{\reportauthors}
\date{\reportdate}
\techreportshorttitle{\RHO}

\input{preamble}

\begin{document}

\thispagestyle{empty}
\noindent
\begin{minipage}[c]{0.5\linewidth}
  \raggedright
  \raisebox{-0.5\height}{\msftbrandmark}
\end{minipage}%
\begin{minipage}[c]{0.49\linewidth}
  \raggedleft
  {\msftdatefont\small\color{msftgray}\reportdate}
\end{minipage}\par
\vspace{0.35em}
\noindent{\color{msftline}\rule{\linewidth}{0.8pt}\par}

\vspace{1em}
\begin{center}
  {{\msfttitlefont\fontsize{17.5}{21}\selectfont\color{msftdark}
  \reporttitle\par}}
  \vspace{1.25em}
  {\normalsize\rmfamily\color{msftdark}
  \reportauthors\par}
\end{center}

\vspace{0.45em}
\begin{msfttitlebox}
  \setlength{\parindent}{0pt}
  \setlength{\parskip}{0pt}
  \input{abstract}
  \vspace{0.14cm}
  {\small
    \msftmetalabel{Project page}\url{https://microsoft.github.io/rhobotics}\par
    \msftmetalabel{Code}\url{https://github.com/microsoft/rhobotics}\par
    \msftmetalabel{Models and data}\url{https://huggingface.co/collections/microsoft/\Rho}\par
    \msftmetalabel{Date}\reportdate\par
  }
\end{msfttitlebox}
\suppressfloats[t]

\input{intro}
\input{phyphi}
\input{arch}
\input{pretraining}
\input{midtraining}
\input{adaptation}
\input{exps}
\input{rel_work}

\input{limitations}

\input{release}
\input{conclusion}
\input{acks}

\bibliographystyle{plainnat-boldtitle}
\bibliography{references}


\newpage

\input{appendix}


\end{document}

%% file: preamble.tex
\usepackage[T1]{fontenc}    
\usepackage[utf8]{inputenc} 

\newlength\savewidth

\newcolumntype{x}[1]{>{\centering\arraybackslash}p{#1pt}}
\newcolumntype{y}[1]{>{\raggedright\arraybackslash}p{#1pt}}

\definecolor{baselinecolor}{gray}{0.93}


%% file: abstract.tex
\begin{abstract}
General-purpose physical AI models must combine broad visual and linguistic capabilities
with precise control across robot embodiments and efficient adaptation to downstream tasks. We
introduce \RHO, a family of open-weights VLA models for bimanual manipulation designed for data-light task adaptation on 3 embodiments representative of dual-arm robots across research labs and the industry -- \yb, \urait, and \fr3d.
We systematically ablate \RHO's action-expert architecture and training recipe, and show in controlled simulation and physical-robot experiments that embodiment midtraining improves downstream adaptation. The resulting \RHO variants for \yb, \urait, and \fr3d match or outperform existing open-weights VLAs and achieve the strongest overall performance across the tasks, embodiments, and baselines evaluated in this report.
We further demonstrate the \RHO model family's built-in capacity for online adaptation: an internal latent policy
learns from corrective feedback to select observation-conditioned noise inputs for the frozen
flow-matching action expert. With as few as 15 corrected episodes, adapting this lightweight module
enables \RHO\ to handle task situations at the fringe of its offline finetuning distribution.
Together, these results position \RHO\ as both a strong general-purpose robotic manipulation model and a practical
foundation for adaptation. We release the base \RHO\ model and the embodiment-specific checkpoints
to facilitate \RHO's deployment in research experiments and practical industrial use cases.
\end{abstract}

%% file: intro.tex
\section{Introduction}

Physical AI implemented as vision-language-action (VLAs) and world-action models (WAMs) offers a path
toward general-purpose robot manipulation by combining
the semantic and perceptual capabilities of pretrained vision-language backbones (VLMs) with action
generation~\citep{zitkovich2023rt2,kim2024openvla,black2024pi0,pi05,gr,fang2026molmoact2,pi07}.
Recent models of this type have been exposed to increasingly broad task coverage and expressive continuous control during training. However, zero-shot success rates of existing VLA and WAMs on unseen manipulation tasks and in new environments remain low. \emph{Adaptation} is a crucial and complex step between a pretrained robot foundation model and a version of it that works in practice.


\begin{figure}[t]
  \centering
  \begin{minipage}[t]{0.32\linewidth}
    \centering
    \includegraphics[width=\linewidth]{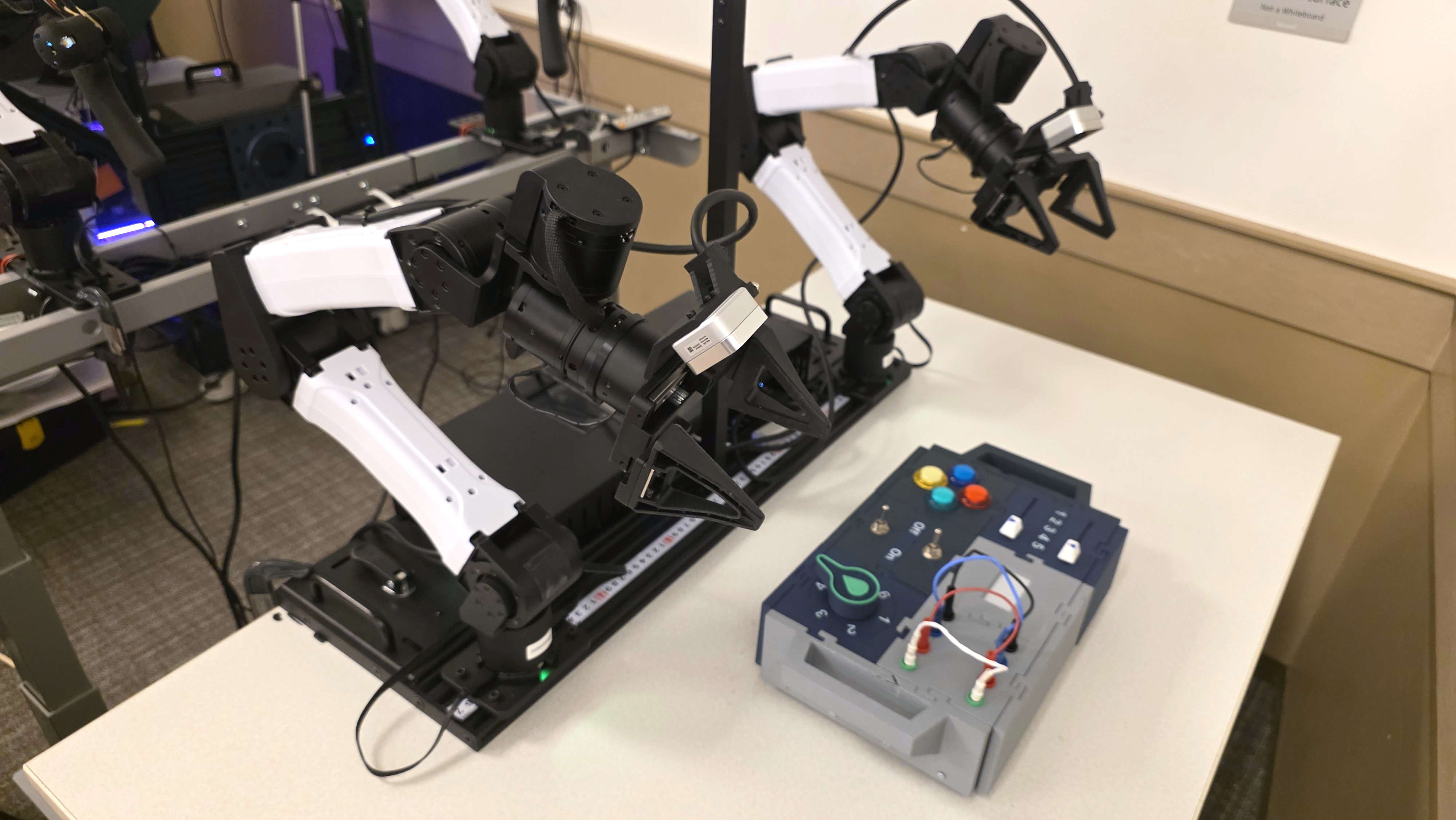}\\[3pt]
    {\vlaPanelTitleFont I$^2$RT YAM Box}
  \end{minipage}\hfill
  \begin{minipage}[t]{0.32\linewidth}
    \centering
    \includegraphics[width=\linewidth]{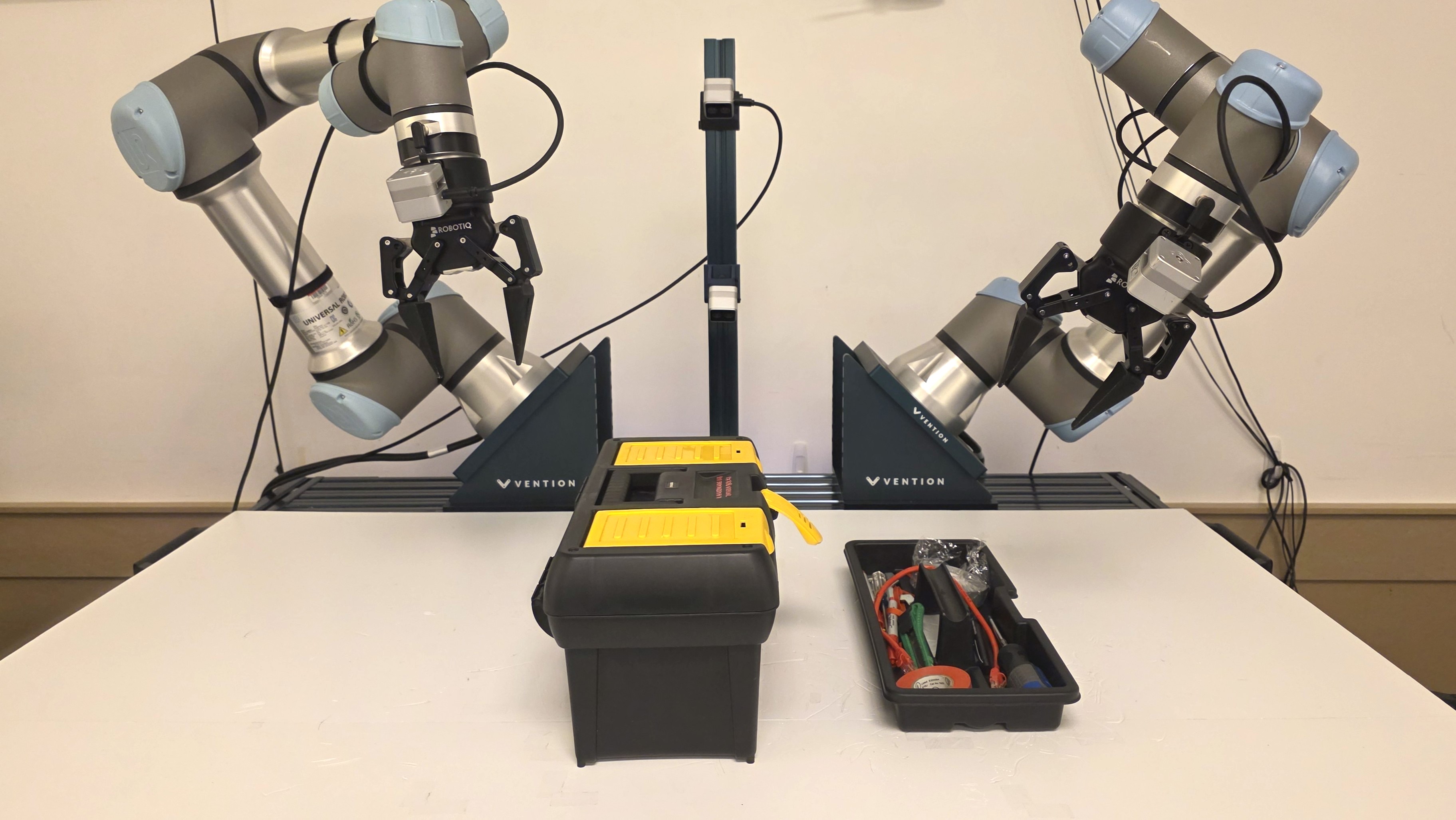}\\[3pt]
    {\vlaPanelTitleFont Universal Robots AI Trainer}
  \end{minipage}\hfill
  \begin{minipage}[t]{0.32\linewidth}
    \centering
    \includegraphics[width=\linewidth]{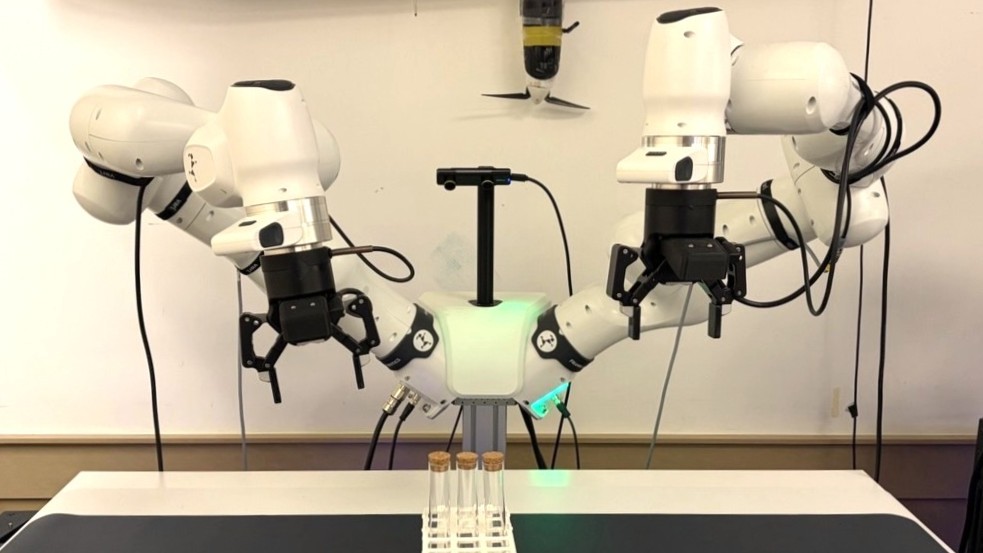}\\[3pt]
    {\vlaPanelTitleFont Franka FR3 Duo}
  \end{minipage}
  \caption{\RHO's physical dual-arm robot embodiments. YAM Box is common in research labs, FR3 Duo's usage straddles research and industry, and UR AI Trainer is an emerging standard setup for industrial bimanual manipulation tasks. Midtrained checkpoints for all three are available at \url{https://huggingface.co/collections/microsoft/\Rho}}
  \label{fig:physical-embodiments}
\end{figure}

This report introduces \RHO, an open-weights family of 5B-parameter physical AI models for bimanual robotic manipulation explicitly designed for data-efficient adaptation to new environments and tasks. \RHO's central premise is that, in order to quickly learn a new task, a physical AI model needs to master vision, language, and control of the specific robot that is expected to perform this task \emph{before} task adaptation begins. Pretraining is insufficient for this: it typically exposes the model to a broad but superficial mixture of tasks, environments, and embodiments, without focusing on sensing, kinematics, action space, and data conventions of any robot in particular. Directly teaching a pretrained model to complete an unseen task, then, involves familiarizing it with the control of the target robot platform, which makes the adaptation process data-intensive. \RHO's training recipe separates adaptation to a robot embodiment from adaptation to a task, absorbing a large part of the task adaptation burden into the \RHO models we present. The \RHO family derives from a single foundation, \RHOB, and consists of several VLAs that can serve as reusable starting points for adapting to many downstream tasks much more data-efficiently than a pretrained-only model would, as our empirical evaluation demonstrates. We also show that after offline task specialization, these VLAs can continue improving from interaction through an internal latent policy, covering blind spots left by typically small finetuning datasets.

We release the \RHO foundation \RHOB along with its variants for 3 representative robot platforms for bimanual manipulation -- I$^2$RT YAM Box, Universal Robots AI Trainer, and Franka FR3 Duo (\Cref{fig:physical-embodiments}). Each of these \RHO variants is built via a recipe we call \emph{regularized progressive adaptation}, whose
steps expose the model to various forms of robotic manipulation training while taking care to preserve useful visual, linguistic, and control capabilities gained in the previous training stages. 
As a starting point, \RHO\ uses a backbone distilled from Microsoft's Phi-family VLM~\citep{aneja2026phi4vision}. \RHO's training recipe adapts this backbone for physical-world comprehension, couples it to a flow-matching action expert that generates continuous action
chunks, and pretrains it in the end-effector-pose action space on a multi-task multi-embodiment data mixture that is dominated by demonstrations from dual-arm robot platforms but also includes robotics-relevant Web VQA data. This produces the foundation \RHOB and is followed by \RHOB's multi-task single-embodiment midtraining. The embodiment-midtraining stage is key: during it, \RHOB focuses on learning to control a target robot across a broad distribution of tasks, motions, and environments, giving rise to \RHO family's embodiment-specific VLAs \emph{\RHOYAM}, \emph{\RHOUAT}, and \emph{\RHOFR}. Offline task adaptation comes after horizontal midtraining and finetunes these VLAs to a target task. For continued adaptation after deployment, each of \RHO-family VLA's flow module includes an optional internal latent policy that maps the current observation to the flow's initial condition, replacing random sampling while leaving the action generator fixed. In our online experiments, only this lightweight internal module learns from corrective interaction, preserving the broader capabilities acquired earlier and serving as a form of regularization.

Our experiments assess whether this combination of reusable embodiment checkpoints and offline as well as online adaptation provides a
practical route from a generalist VLA to reliable task execution, both in simulation and on physical robots. We systematically study the architecture of \RHO's action expert, the composition and scheduling of its pretraining mixture, and the effect of its embodiment midtraining on its downstream task adaptation. Across the settings and open-weights baselines evaluated in this report, \RHO\ achieves the strongest overall performance.

Thus, the main contributions of our work are:
\begin{itemize}
    \item We release the \RHO model family consisting of the \RHOB foundation checkpoint and its 3 embodiment-midtrained VLA variants \RHOYAM, \RHOUAT, and \RHOFR, for researchers and practitioners to adapt to their own robotic manipulation scenarios.
    \item We systematically evaluate \RHO's architectural and multi-stage training recipe that combines
          physical grounding with heterogeneous, multi-embodiment multi-task robot data, horizontal embodiment midtraining, and vertical task adaptation. Our studies, performed in RoboTwin~\citep{chen2025robotwin}, LIBERO~\citep{liu2023libero},
          and RoboEval~\citep{wang2026roboeval} simulation environment, and on common dual-arm robot setups \yb, \urait, and \fr3d demonstrate this recipe's broad positive effects.
    \item We empirically analyze the influence of midtraining on downstream adaptation and show that it halves the amount of task finetuning data needed to achieve a given success rate threshold, compared to finetuning pretrained \RHO without midtraining. We also demonstrate that \RHO{} can continue improving after offline task specialization by adapting its internal latent policy from corrective interaction.
\end{itemize}

Together, these results establish \RHO\ as a strong open-weights VLA and, more broadly, examine how
architecture, pretraining, embodiment midtraining, and task adaptation interact in a
practical robot-learning pipeline. 

%% file: phyphi.tex
\section{Physically grounded Phi-family backbone}
\label{sec:phyphi}

\subsection{Backbone architecture}

\RHO\ begins from a vision-language model developed within the Microsoft Phi model family. The Phi family emphasizes capability relative to model size and compute through careful architecture, data curation, and staged training~\citep{aneja2026phi4vision}. This emphasis is well matched to robot control, where visual-language inference is part of an interactive control loop and must share a deployment budget with action generation.

We refer to the vision-language backbone used by \RHO\ as \emph{\phiphy}. It is a compact, 4.68-billion-parameter model with a Phi-family causal language decoder and a SigLIP~2 NaFlex vision encoder~\citep{tschannen2025siglip2}. Following the mid-fusion design studied in Phi-4-reasoning-vision-15B~\citep{aneja2026phi4vision}, it has a two-layer cross-modal projector maps visual features into the language embedding space. The projected visual ``soft tokens'' are interleaved with the text-token embeddings and processed by the language decoder. This design gives the decoder joint access to visual and linguistic context while retaining a modular visual encoder and language model.

\begin{table}[H]
  \centering
  \small
  \begin{tabular}{@{}ll@{}}
    \toprule
    Component & Specification \\
    \midrule
    Full vision-language model & 4.68B parameters \\
    Language decoder & 32 layers; width 3,072; MLP width 8,192 \\
    Decoder attention & 32 query heads; 32 key/value heads \\
    Vocabulary and positions & 100,352 tokens; 16,384 configured positions \\
    Vision encoder & SigLIP~2 SO400M NaFlex; 428M parameters \\
    Image representation & $16\!\times\!16$ patches; 256--3,600 visual tokens \\
    Cross-modal projector & MLP: 1,152 $\rightarrow$ 3,072 $\rightarrow$ 3,072, with GELU \\
    \bottomrule
  \end{tabular}
  \caption{Architecture of the \phiphy\ vision-language backbone used to initialize \RHO.}
  \label{tab:phiphy-architecture}
\end{table}

The NaFlex encoder preserves image aspect ratio and varies the number of visual tokens with image size, subject to minimum and maximum patch budgets. \phiphy\ takes features from the penultimate vision-encoder layer, removes padded visual positions, and maps the remaining features through the cross-modal projector. The resulting sequence can contain one or more images interleaved with language. For \RHO\ pretraining, camera views are resized with padding to $256\!\times\!256$ before this encoding step, giving a predictable visual-token budget for multi-camera control.

\subsection{Physical grounding data}

\begin{table}[hb]
\centering
\small
\setlength{\tabcolsep}{5pt}
\renewcommand{\arraystretch}{1.1}

\begin{tabularx}{\linewidth}{@{}
    >{\hsize=.9\hsize\linewidth=\hsize\raggedright\arraybackslash}X
    >{\hsize=0.85\hsize\linewidth=\hsize\raggedright\arraybackslash}X
    >{\hsize=1.25\hsize\linewidth=\hsize\raggedright\arraybackslash}X
@{}}
\toprule
\textbf{Name} & \textbf{Description} & \textbf{Example} \\
\midrule

Attribute Similarity
& Find the odd-one-out by font, color, or layout.
& Which screenshot uses a different font? \\
\addlinespace[0.65em]

Component Counting
& Count a UI component across screenshots.
& How many screenshots contain a card? \\
\addlinespace[0.65em]

Object Counting
& Count objects across natural images.
& How many boats in all images? \\
\addlinespace[0.65em]

Chart Type Matching
& Find the odd-one-out by chart type.
& Which chart is not a grouped bar chart? \\
\addlinespace[0.65em]

Cross-Chart Reasoning
& Combine evidence from related charts.
& What patterns emerge across the charts? \\
\addlinespace[0.65em]

Chart Difference Spotting
& Identify chart changes and their effects.
& What changed; how were profits affected? \\
\addlinespace[0.65em]

UI Difference Spotting
& Find components that were added, removed, or restyled.
& What changed between these screenshots? \\
\addlinespace[0.65em]

Multi-Image UI Grounding
& Locate and compare named UI elements.
& Mark the ``Support'' link in each image.  \\
\addlinespace[0.65em]

PIL-Style Transform Detection
& Identify PIL-style transforms  between paired images.
& What image transform occurred? \\
\addlinespace[0.35em]

Multilingual OCR Retrieval
& Localize text across multilingual documents.
& Which images contain ``bonjour''; where? \\

\bottomrule
\end{tabularx}

\caption{Multi-image data categories and example VQA questions.}
\label{tab:multi-image-categories}
\end{table}

General-purpose vision-language posttraining mixture of Phi-4-reasoning-vision-15B provides broad semantic and perceptual knowledge but does not directly emphasize the physical relationships most relevant to robot control, with the exception of PixMo collection~\cite{deitke2025molmo}. For Phi-Phy, we use additional data that is a mixture of robotics-relevant visual question answering and grounding data. The objective is to strengthen physical perception and reasoning before the model is asked to learn continuous actions from the comparatively smaller robot-data distribution. The data additions came in three categories:
\begin{itemize}
    \item High-quality publicly available robotics-relevant data: MGrounding~\cite{li2025migician}, RefSpatial~\cite{zhou2025roborefer}, XVR~\cite{jeong2026learning}, and the object-reference and spatial-reference subsets of RoboPoint~\cite{yuan2024robopoint}.
    \item Multi-image grounding, pointing, OCR, and counting data, generated by reprocessing single-image VQA data in robotics-relevant ways.
    \item Synthetic multi-image difference-spotting and odd-one-out VQA, created using synthetic-chart and synthetic-UI environments; in particular, with ground truth VQA.
\end{itemize}
Detailed descriptions of the latter two categories can be found in Table \ref{tab:multi-image-categories}.
Out of 100M SFT records in the final data mix,  roughly 6M fall in the above three categories, with an additional 30M of general caption, perception, object-existence, and UI-heavy grounding, pointing, and relational data.

%% file: arch.tex
\section{Flow-matching action architecture}
\label{sec:architecture}

\RHO\ converts the physically grounded Phi-family backbone from \Cref{sec:phyphi} into a vision-language-action model by coupling it to a continuous action expert. At each control step, the model conditions on visual observations, a language instruction, and robot state, and generates a chunk of future actions with a flow-matching objective. This section defines the model interfaces and the architectural choices evaluated later in \Cref{sec:architecture-ablations}.

\subsection{Observation and action representations}

The observation sequence combines one or more camera views with the task instruction and proprioceptive state. Each camera view occupies an image placeholder in the \phiphy\ chat sequence, so visual tokens from multiple cameras are interleaved with the language instruction before the decoder processes them. Proprioceptive state is kept outside this sequence and mapped to the action-expert width by a learned linear projection.

\RHO\ represents robot state and actions as continuous vectors with up to 32 dimensions, without requiring a particular control parameterization. The same interface can support joint-space or end-effector control, delta or absolute targets, and single- or multi-arm embodiments. The 32-dimensional ceiling leaves headroom for higher-dimensional joint-space and auxiliary controls. Dataset adapters pad smaller vectors to the common width and supply masks so that unused state dimensions, action dimensions, and chunk positions do not contribute to the loss.

The model predicts chunks of future actions rather than individual actions~\citep{zhao2023act}, which can reduce compounding error and provide ensemble-like generalization across temporal offsets~\citep{lazzati2026chunking}. The prediction horizon and control frequency are configurable properties of a training recipe rather than fixed architectural constants. The execution horizon is likewise a deployment-time choice: a controller may execute the full prediction or replan after a shorter prefix. \RHO can also be combined with real-time chunking (RTC)~\citep{black2025rtc} to update chunks asynchronously while maintaining smooth transitions between successive predictions.

\subsection{Action expert}
\label{sec:action-expert}

The action expert is trained to transform Gaussian noise into continuous robot action chunks conditioned on the backbone representation. Given a ground-truth action chunk $a$, noise $\epsilon$, and time $t$, we construct $x_t=t\epsilon+(1-t)a$ and train the model to predict the velocity $\epsilon-a$ with a mean-squared flow-matching loss. At inference, the learned field is integrated from noise at $t=1$ toward an action chunk at $t=0$; the default configuration uses 10 explicit Euler steps.

\begin{figure}[t]
  \centering
  \includegraphics[width=0.95\linewidth]{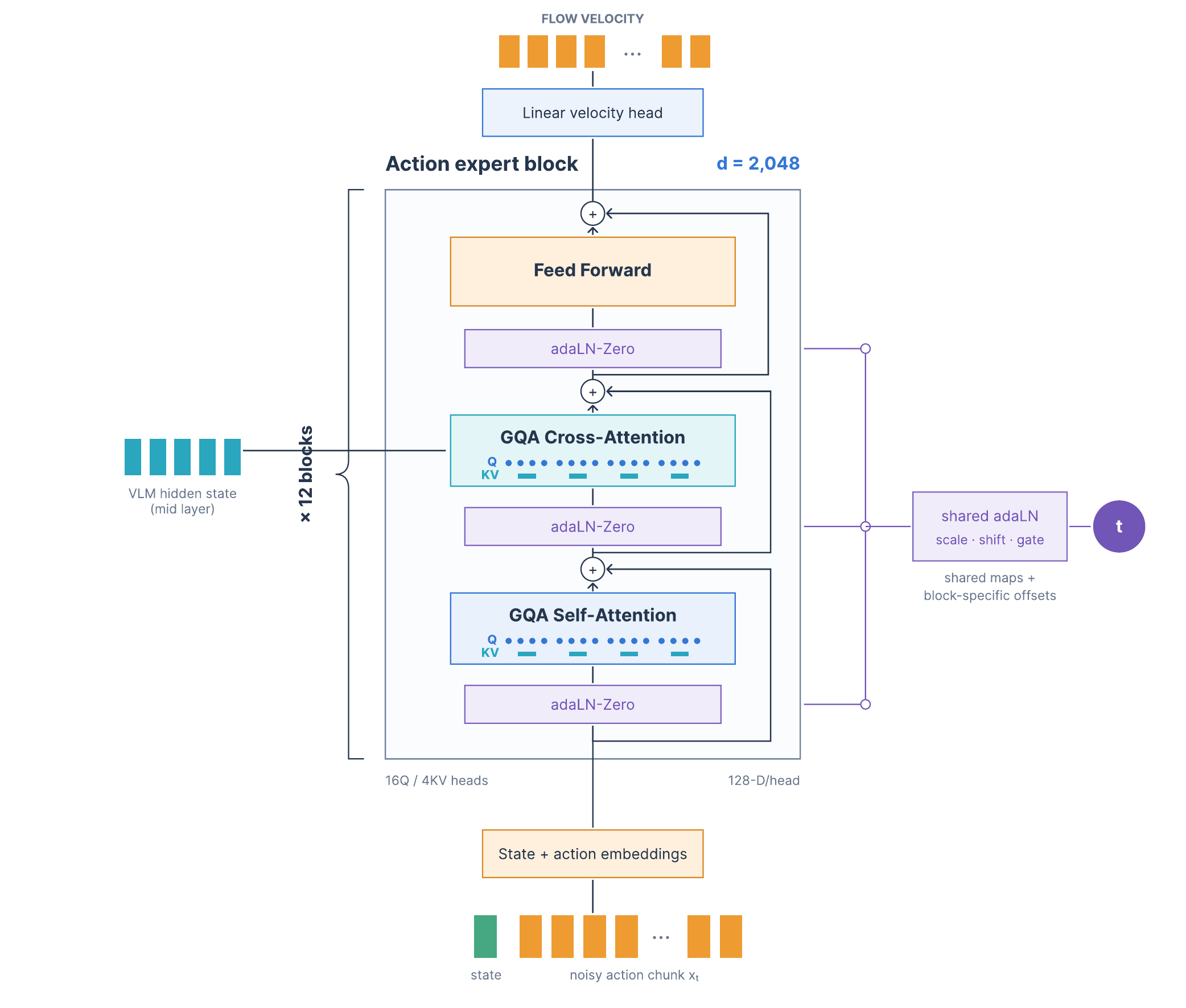}
  \caption{The \RHO\ flow-matching action expert. State and noisy-action tokens pass through 12 wide transformer blocks with grouped-query self-attention, cross-attention to the VLM context, and feed-forward updates. The flow timestep conditions the sublayers through shared adaLN-Zero maps with block-specific offsets.}
  \label{fig:action-expert}
\end{figure}

Robot state and noisy-action tokens are processed by a transformer action expert. Each block applies action-stream self-attention, cross-attention to the \phiphy\ context, and a feed-forward update. \Cref{fig:action-expert,tab:action-expert-architecture} illustrate the block structure and summarize the complete configuration.

The flow timestep conditions the expert through DiT-style adaLN-Zero, which produces timestep-dependent scale, shift, and residual-gating signals for the transformer updates~\citep{peebles2023dit}. Rather than learning a separate modulation network in every block, \RHO\ uses the parameter-efficient adaLN-single design: the base modulation maps are shared across blocks and each block learns a small zero-initialized offset~\citep{chen2024pixartalpha,reuss2025flower}. With 12 blocks, shared modulation, and grouped-query attention, the resulting action expert contains 542 million parameters. These choices preserve the 2,048-dimensional action representation and 128-dimensional attention heads while reducing capacity along dimensions for which our ablations show little benefit.

The selected shape balances per-head resolution with relational parallelism. The wide action stream gives the flow-matching expert capacity to represent high-fidelity conditional velocity fields, while sixteen 128-dimensional heads can retain fine geometric detail and model many relationships among observations, robot state, and actions in parallel. We expect this balance to matter most for contact-rich and precision manipulation, where successful action chunks occupy a narrow region of action space. The task-level RoboEval results in \Cref{sec:architecture-ablations} support this interpretation.

\begin{table}[t]
  \centering
  \small
  \begin{tabular}{@{}lp{0.64\textwidth}@{}}
    \toprule
    Component & Specification \\
    \midrule
    Action expert & 542M parameters \\
    Transformer & 12 blocks; width 2,048; MLP width 4,096 \\
    Attention & 16 query heads; 4 key/value heads; head dimension 128 \\
    Block structure & Action self-attention; cross-attention to VLM context; feed-forward network \\
    Flow conditioning & Shared adaLN-single with block-specific offsets \\
    Flow target & Linear noise--action interpolation; velocity-prediction MSE \\
    State/action interface & Continuous vectors with up to 32 dimensions; padding and loss masks \\
    Inference solver & 10 explicit Euler steps \\
    Prediction horizon & Configurable action-chunk length \\
    Backbone context & Decoder block 14; learned $3{,}072 \rightarrow 2{,}048$ projection \\
    \bottomrule
  \end{tabular}
  \caption{Configuration of the \RHO\ flow-matching action expert.}
  \label{tab:action-expert-architecture}
\end{table}

\subsection{Backbone--expert interface}

The interface between the VLM and action expert determines how visual and linguistic information reaches the continuous controller and how action-learning gradients affect the pretrained backbone. \RHO\ reads the joint image--language sequence from an intermediate \phiphy\ decoder representation. A learned linear map projects this 3,072-dimensional sequence to the 2,048-dimensional action-expert space. Every action-expert block cross-attends to this same projected sequence, while the action stream contains the projected robot-state token followed by the noisy action tokens. The VLM context is computed once and reused across all flow-integration steps during inference.

During multi-embodiment pretraining, the vision encoder, cross-modal projector, and language-decoder blocks remain trainable at a small learning rate, while the action expert and robot-specific projections use a larger learning rate. The language token-embedding table remains frozen. Vision-language batches regularize the backbone during multi-embodiment pretraining. We evaluate alternative backbone--expert interfaces, trainable parameter groups, and initialization choices in \Cref{sec:architecture-ablations,sec:pretraining-ablations}.

%% file: pretraining.tex
\section{Pretraining}
\label{sec:pretraining}

Multi-embodiment multi-task pretraining spanning diverse environments and data sources turns the grounded backbone and action expert into a general continuous-control VLA, with the goal of learning transferable action representations before adapting to a particular physical platform.

\subsection{Pretraining data mixture}

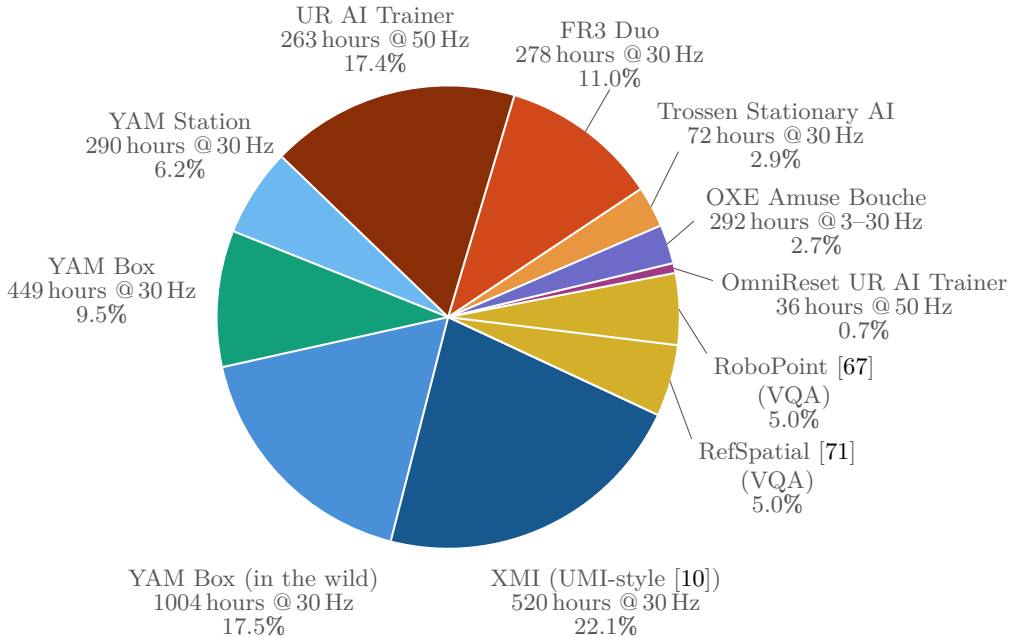
\begin{figure}[h]
  \centering
  \captionsetup{skip=\vlaCaptionSkip}
  \input{figures/fig-pretraining-mixture}
  \caption{\RHO-Tomyum mixture. Its 90\% is composed of demonstration trajectories from physical and simulated robots as well as UMI-style demonstrations collected with standard YAM grippers, and the remaining 10\% are vision-language data. All the robot/UMI-style data is bimanual, with the exception of a small subset of Open X-Embodiment Magic Soup~\cite{black2024pi0} we call \emph{OXE Amuse Bouche}. The robot data includes trajectories from policies controlling a UR AI Trainer simulated in NVIDIA Isaac Sim, trained using RL and the OmniReset framework~\cite{yin2026emergent}.}
  \label{fig:pretraining-mixture}
\end{figure}

We call our combined pretraining data mixture \emph{\RHO-Tomyum} (\Cref{fig:pretraining-mixture}). Its robot-data portion spans more than a dozen single- and dual-arm embodiments and combines publicly available datasets~\citep{oxe2024} with proprietary physical-robot and UMI-style~\cite{umi} demonstrations. It also contains a smaller set of simulated trajectories we have generated in NVIDIA Isaac Sim~\citep{isaacsim} for the UR AI Trainer embodiment using a method similar to OmniReset~\cite{yin2026emergent}. We use only a small subset of the \emph{OXE Magic Soup$++$} collection~\cite{black2024pi0}, which we call \emph{OXE Amuse Bouche} and which omits Fractal, Taco Play, DROID, Stanford Hydra, Furniture Bench, Austin Sailor, and BC-Z datasets.  The mixture covers diverse manipulation tasks, environments, camera layouts, action spaces, and control frequencies and is dominated by demonstrations from dual-arm robots, as \RHO is geared mainly toward bimanual manipulation. We map observations and actions from each robot source into the padded model interface in \Cref{sec:architecture} and apply the pretraining action standardization in \Cref{sec:pretraining-action-space}. We then draw examples using fixed source weights chosen to balance dataset scale and embodiment coverage rather than sampling a raw concatenation uniformly.

The vision-language portion of \RHO-Tomyum consists of RoboPoint~\citep{yuan2024robopoint} and RefSpatial~\citep{zhou2025roborefer}, sampled with equal probability. RoboPoint supplies spatial-affordance pointing examples together with object-reference and general visual-question-answering data. From RefSpatial, we use its single-image 2D and 3D spatial-choice and reasoning examples, including vacant-space pointing and visual-choice tasks; we omit multi-view examples and use only RGB inputs.

\subsection{Robot data standardization}
\label{sec:pretraining-action-space}

Although \RHO\ can use different continuous control parameterizations within its 32-dimensional interface, we standardize the heterogeneous \RHO-Tomyum robot datasets to a common chunk-relative delta end-effector (EEF) representation for pretraining. For each arm and future chunk position $k$, the target contains a three-dimensional translation delta, a relative orientation in the continuous 6D rotation representation~\citep{zhou2019continuity}, and an absolute gripper position. Both pose deltas are measured relative to the end-effector pose in the current observation at the start of the chunk, rather than relative to the preceding action in the chunk. This gives 10 action dimensions per arm; multi-arm actions are concatenated and then padded to width 32.

For each dataset, we construct a target spanning approximately one second at its native control rate and pad it to at most 50 steps. We normalize actions separately for every source dataset, action dimension, and chunk position: the 1st and 99th percentiles map to $-1$ and $1$, respectively, and values outside this range are clipped to $[-1.5,1.5]$. This per-position normalization accounts for the fact that chunk-relative displacement magnitudes generally grow with temporal offset. Current robot state uses per-dataset, per-dimension 1st- and 99th-percentile normalization. Masks remove padded time steps and inactive state and action dimensions from the loss. Our reported embodiment-midtraining and task-adaptation experiments reuse this convention unless otherwise noted, but it is not required by the architecture.

The \yb, \urait, and \fr3d datasets consist of long-horizon tasks requiring high-dexterity, where the errors in task demonstrations are virtually unavoidable. In our raw data, the demonstration segments containing erroneous motions were labeled as such, and we removed them: the dataset sizes provided in \Cref{fig:pretraining-mixture} correspond to the clean, processed data with the errors omitted. We note that practices on the inclusion of demonstration errors differ. E.g., \PI~\cite{pi05} uses them in pretraining but eliminates them during posttraining, and $\pi_{0.7}$~\cite{pi07} includes them while labeling them as errors.   

\subsection{Pretraining procedure}

Training initializes the vision-language backbone from \phiphy\ and randomly initializes the state and action projections and the 12-block action expert described in \Cref{sec:architecture}. We jointly optimize the vision encoder, cross-modal projector, language decoder, and action modules, while keeping the token-embedding table frozen. Robot examples contain one current observation, a task instruction, proprioceptive state, and up to three RGB camera views; images are resized with padding to $256\!\times\!256$. Throughout pretraining, embodiment midtraining, and offline task adaptation, we apply color jitter to both scene- and wrist-camera images. The scene view additionally receives mild random in-plane rotation and random cropping, while wrist views receive photometric augmentation only. Robot trajectories use the standardized representation in \Cref{sec:pretraining-action-space}.

We train with AdamW in bfloat16 using separate peak learning rates of $1\!\times\!10^{-6}$ for the pretrained vision-language modules and $1\!\times\!10^{-4}$ for the newly initialized action modules. A warmup--stable--decay schedule linearly warms up for 1,000 updates, holds the peak rates, and cosine-decays both rates by a factor of ten over the final 45,000 updates. The full pretraining run uses 48 NVIDIA B200 GPUs. Robot updates use 64 examples per GPU, giving a global robot batch size of 3,072; the longer vision-language updates use 8 examples per GPU, or 384 globally. \Cref{tab:pretraining-recipe} summarizes the recipe. We separately study the effects of component initialization, trainable parameter groups, training schedule, and data-mixture composition in \Cref{sec:pretraining-ablations}.

\subsection{Vision-language co-training}

During multi-embodiment pretraining, each optimization update uses either a robot batch or a vision-language batch. With probability $0.9$, we sample a robot batch and optimize the flow-matching objective through the backbone and action expert. With probability $0.1$, we sample RoboPoint or RefSpatial with equal probability and present the backbone with RGB image inputs, a text prompt, and a target answer. These updates bypass the robot-state and action inputs and train the backbone with autoregressive cross-entropy on visual-question-answering or pointing targets; the action expert is not used for this loss.

We multiply the vision-language cross-entropy loss by $0.02$ before backpropagation, while leaving the robot flow-matching loss unscaled. The $9{:}1$ ratio and the $0.02$ coefficient therefore control different quantities: the former determines how often each type of update occurs, and the latter controls the gradient scale of a vision-language update. This co-training keeps physically relevant spatial and visual-language supervision in the optimization mixture while the backbone adapts to continuous control. \Cref{sec:pretraining-ablations} compares this recipe with otherwise matched robot-only pretraining.

%% file: figures/fig-pretraining-mixture.tex
\definecolor{pieXmi}{HTML}{17598F}      
\definecolor{pieYamWild}{HTML}{4A90D9}
\definecolor{pieYamBox}{HTML}{12A07A}
\definecolor{pieYamSta}{HTML}{6CB8F0}
\definecolor{pieUr}{HTML}{8A2E08}       
\definecolor{pieFr3}{HTML}{D1491B}
\definecolor{pieTrossen}{HTML}{E8963F}
\definecolor{pieOxe}{HTML}{6E6AC8}      
\definecolor{pieSim}{HTML}{A03A82}      
\definecolor{pieRoboPt}{HTML}{D4B02A}   
\definecolor{pieRefSp}{HTML}{D4B02A}

\def\piedata{%
  22.05/pieXmi/{XMI (UMI-style~\cite{umi})}/520/30/999/0/0/0/0.45/-0.3,
  17.53/pieYamWild/YAM Box (in the wild)/1004/30/999/0/0.55/0/0.7/-0.60,
   9.53/pieYamBox/YAM Box/449/30/999/0/0/0/0/0,
   6.15/pieYamSta/YAM Station/290/30/999/0/0/0/0/0,
  17.42/pieUr/UR AI Trainer/263/50/999/0/0/0/0/0,
  11.03/pieFr3/FR3 Duo/278/30/999/0/0.55/4/-0.5/-0.1,
   2.87/pieTrossen/Trossen Stationary AI/72/30/2.70/2.40/0/3/0/0,
   2.73/pieOxe/OXE Amuse Bouche/292/3--30/3.35/1.27/0/1/0/0,
   0.69/pieSim/OmniReset UR AI Trainer/36/50/3.55/0.14/0/5/0/0,
   5.00/pieRoboPt/{RoboPoint~\cite{yuan2024robopoint}\\(VQA)}/0/0/3.45/-0.99/0/2/0/0,
   5.00/pieRefSp/{RefSpatial~\cite{zhou2025roborefer}\\(VQA)}/0/0/3.25/-2.12/0/2/0/0%
}

\newcommand{\pielab}[4]{%
  #1\\[-2pt]%
  \ifdim #2pt>0pt #2\,hours @\,#3\,Hz\\[-2pt]\fi
  \textbf{\pgfmathprintnumber[fixed, fixed zerofill, precision=1]{#4}\%}}

\ifdefined\piemidw\else\newlength{\piemidw}\fi
\settowidth{\piemidw}{\vlaTickFont 72\,hours @\,30\,Hz}
\begin{tikzpicture}[every node/.style={font=\vlaTickFont}]
  \def\Rad{3.05}        
  \def\Rlab{3.28}       
  \def\startang{-25}   
  \xdef\angacc{\startang}

  \foreach \pct/\col/\name/\hrs/\rate/\tx/\ty/\rex/\lead/\ldx/\ldy in \piedata {
    \pgfmathsetmacro{\sweep}{\pct*3.6}
    \pgfmathsetmacro{\astart}{\angacc}
    \pgfmathsetmacro{\aend}{\angacc-\sweep}
    \pgfmathsetmacro{\amid}{(\astart+\aend)/2}
    \pgfmathsetmacro{\rl}{\Rlab+\rex}
    \fill[\col] (0,0) -- (\astart:\Rad) arc[start angle=\astart,
                end angle=\aend, radius=\Rad] -- cycle;
    \draw[white, line width=0.8pt] (0,0) -- (\astart:\Rad);
    \ifdim\tx pt>900pt
      \ifnum\lead=1
        \draw[vlaNote, line width=0.4pt] (\amid:\Rad) --
          ([xshift=\ldx cm, yshift=\ldy cm]\amid:\rl-0.10);
      \fi
      \node[anchor={\amid+180}, align=center, text=vlaNote, inner sep=1.5pt,
            xshift=\ldx cm, yshift=\ldy cm]
        (plab) at (\amid:\rl) {\pielab{\name}{\hrs}{\rate}{\pct}};
      \ifnum\lead=4 
        \draw[vlaNote, line width=0.4pt] (\amid:\Rad) --
          ([yshift=0.03cm]plab.south);
      \fi
    \else
      \node[anchor=west, align=center, text=vlaNote, inner sep=1.5pt]
        (lab) at (\tx,\ty) {\pielab{\name}{\hrs}{\rate}{\pct}};
      \ifnum\lead=2
        \draw[vlaNote, line width=0.4pt] (\amid:\Rad) --
          ([xshift=-0.03cm, yshift=-0.03cm]lab.north west);
      \else\ifnum\lead=5
        \draw[vlaNote, line width=0.4pt] (\amid:\Rad) --
          ([xshift=1.5pt-1mm, yshift=-4.3pt]lab.north west);
      \else\ifnum\lead=3
        \draw[vlaNote, line width=0.4pt] (\amid:\Rad) --
          ($(lab.center)+(-0.5\piemidw-0.5pt-1mm,-3.2pt-1mm)$);
      \else
        \draw[vlaNote, line width=0.4pt] (\amid:\Rad) -- (\tx-0.10,\ty);
      \fi\fi\fi
    \fi
    \xdef\angacc{\aend}
  }
  \draw[white, line width=0.8pt] (0,0) -- (\startang:\Rad);

  \path let \p1=(current bounding box.east) in (-\x1,0);
\end{tikzpicture}

%% file: midtraining.tex
\section{Embodiment midtraining}
\label{sec:midtraining}

Multi-embodiment pretraining provides a generally-initialized foundation model, but deployment on a physical platform still requires learning its observation layout, kinematics, action space, control conventions, and characteristic workspace. \RHO\ separates this embodiment adaptation from task adaptation. The \ethm stage adapts the \RHO foundation to a target robot type horizontally across many tasks and motions, producing a checkpoint that can be reused for subsequent data-efficient vertical adaptation to individual downstream tasks. 

\subsection{Embodiment adaptation}

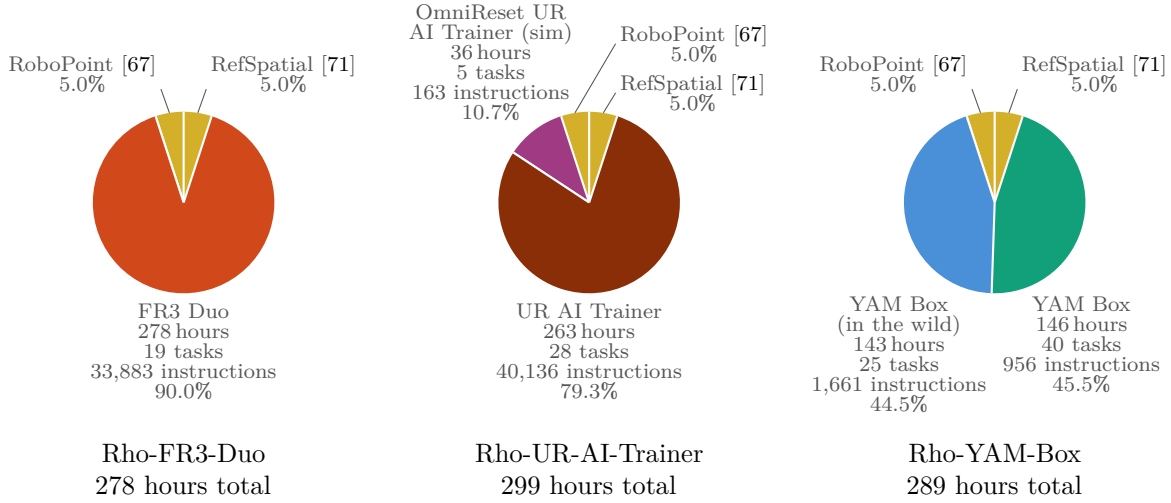
\begin{figure}[h]
  \centering
  \captionsetup{skip=\vlaCaptionSkip}
  \input{figures/fig-midtraining-mixture}
  \caption{Midtraining mixtures of the embodiment-specific \RHO variants. The count of distinct language instructions includes the number of both task and subtask descriptions.}
  \label{fig:midtraining-mixture}
\end{figure}

Each embodiment checkpoint is initialized from pretrained \RHO\ and trained for 2 epochs over a broad multitask dataset from the target platform. We refer to the resulting checkpoints as \RHOFR, \RHOUAT, and \RHOYAM. Midtraining uses the action standardization in \Cref{sec:pretraining-action-space} and preserves the pretraining co-training recipe: robot and vision-language updates are sampled in a $9{:}1$ ratio, and the vision-language loss is weighted by $0.02$. Thus, the model adapts to a platform's observations, kinematics, and control conventions while continuing to receive the spatial and visual-language supervision used during pretraining. \Cref{fig:midtraining-mixture} shows the composition of each checkpoint's midtraining mixture. Experiments studying the effect of embodiment midtraining are presented in \Cref{sec:midtraining-eval}; the adaptability of \RHOFR, \RHOUAT, and \RHOYAM to physical-robot tasks is assessed in Sections \ref{sec:yb-exps}, \ref{sec:urait-exps}, and \ref{sec:fr3d-exps}. 

We emphasize that the goal of embodiment adaptation is to familiarize \RHO with the controls and visuals from the embodiment's cameras across a diverse set of tasks rather than to have it master any task in particular. Note that the average number of hours per task in the midtraining datasets is no more than 15 (e.g., $263 / 28 \approx 9$ for UR AI Trainer). This is generally insufficient to achieve 90+\% success rates on high-precision long-horizon tasks but is helpful for representing a broad distribution of motions in several hundreds of hours of data.

\subsection{\yb}

\RHOYAM\ is a version of \RHO\ for the dual-arm \yb platform trained using a roughly 300-hour subset of the available data, sampled across 65 task families. The mixture is divided approximately evenly between demonstrations from controlled laboratory  and demonstrations collected in everyday environments (146 and 143 hours, respectively, sampled from the pretraining YAM Box datasets in \Cref{fig:pretraining-mixture} uniformly at the episode level). Its tasks range from battery loading, cable winding, terminal-block and faucet installation, and tool-head replacement to folding clothing, cleaning office equipment, cutting with scissors, assembling Lego pieces, organizing household objects, and inserting earbuds into a charging case. This combination exposes the model to precise insertion and assembly, articulated and deformable objects, tool use, and long-horizon bimanual manipulation under substantially more varied visual conditions than the laboratory data alone.

\subsection{UR AI Trainer}

\RHOUAT\ is midtrained on approximately 260 hours of physical demonstrations spanning 28 task families on the dual-arm platform, augmented with approximately 36 hours of successful simulated trajectories covering 5 tasks (see \Cref{sec:urait-descr}), collected in NVIDIA Isaac Sim~\citep{isaacsim}. The physical data includes tool and electronics packing, object sorting, cable routing and connection, storage-device insertion, garment handling, and protective wrapping. To increase repetition on precise contact and placement behaviors, we complement real-world data with simulated rollouts collected on the exact same robotic embodiment and camera setup. This simulated dataset adds three structured manipulation families: cube stacking, left- and right-arm placement, and left- and right-arm insertion. The placement tasks span high object diversity: they involve 10 unique everyday objects in over 80 distinct source-and-target combinations. To improve robustness and generalization, all simulated data incorporates domain randomization across object color, camera positions, and focal lengths.

\subsection{\fr3d}

\RHOFR\ adapts \RHO\ to a dual-arm Franka platform using about 280 hours of physical demonstrations spanning 19 task families. The collection covers bimanual and contact-rich behaviors at different time scales, including surface cleaning, folding and stacking towels, sorting utensils and other objects, pouring between containers, setting tables, packing and unpacking, and inserting cables into computer ports. These tasks require the checkpoint to learn both broad workspace motions and precise coordination between the two arms.

%% file: figures/fig-midtraining-mixture.tex
\definecolor{pieYamWild}{HTML}{4A90D9}  
\definecolor{pieYamBox}{HTML}{12A07A}
\definecolor{pieUr}{HTML}{8A2E08}
\definecolor{pieFr3}{HTML}{D1491B}
\definecolor{pieSim}{HTML}{A03A82}
\definecolor{pieRoboPt}{HTML}{D4B02A}
\definecolor{pieRefSp}{HTML}{D4B02A}

\newcommand{\midpielab}[5]{%
  #1\\[-2pt]%
  \ifdim #2pt>0pt #2\,hours\\[-2pt]#3 tasks\\[-2pt]#4 instructions\\[-2pt]\fi
  \textbf{\pgfmathprintnumber[fixed, fixed zerofill, precision=1]{#5}\%}}

\newcommand{\midpie}[2]{%
  \begin{tikzpicture}[every node/.style={font=\scriptsize}]
    \def\Rad{1.2}
    \useasboundingbox (-2.6,-2.8) rectangle (2.6,3.0);
    \xdef\angacc{#1}
    \foreach \pct/\col/\name/\hrs/\tsk/\ins/\tx/\ty/\anc/\aln/\lead/\ldy in {#2} {
      \pgfmathsetmacro{\sweep}{\pct*3.6}
      \pgfmathsetmacro{\astart}{\angacc}
      \pgfmathsetmacro{\aend}{\angacc-\sweep}
      \pgfmathsetmacro{\amid}{(\astart+\aend)/2}
      \fill[\col] (0,0) -- (\astart:\Rad) arc[start angle=\astart,
                  end angle=\aend, radius=\Rad] -- cycle;
      \draw[white, line width=0.8pt] (0,0) -- (\astart:\Rad);
      \ifnum\lead=1
        \draw[vlaNote, line width=0.4pt] (\amid:\Rad) -- (\tx,\ty);
      \fi
      \node[anchor=\anc, align=\aln, text=vlaNote, inner sep=1.5pt]
        at (\tx,\ty+\ldy) {\midpielab{\name}{\hrs}{\tsk}{\ins}{\pct}};
      \xdef\angacc{\aend}
    }
    \draw[white, line width=0.8pt] (0,0) -- (#1:\Rad);
  \end{tikzpicture}}

\newcommand{\midpiecap}[2]{%
  \par\vspace{\vlaCaptionSkip}{\small #1\\#2 hours total\par}}

\begin{subfigure}[t]{0.33\textwidth}
  \centering
  \midpie{108}{%
    5.00/pieRoboPt/{RoboPoint~\cite{yuan2024robopoint}}/0/0/0/-0.3/1.5/south east/center/1/-0.1,
    5.00/pieRefSp/{RefSpatial~\cite{zhou2025roborefer}}/0/0/0/0.3/1.5/south west/center/1/-0.1,
    90.00/pieFr3/FR3 Duo/278/19/33{,}883/0/-1.28/north/center/0/0}
  \midpiecap{\RHOFR}{278}
\end{subfigure}\hfill
\begin{subfigure}[t]{0.33\textwidth}
  \centering
  \midpie{146.63}{%
    10.73/pieSim/{OmniReset UR\\[-2pt]AI Trainer (sim)}/36/5/163/-1.30/1.03/south/center/0/0,
    5.00/pieRoboPt/{RoboPoint~\cite{yuan2024robopoint}}/0/0/0/0.35/2.1/west/center/1/0,
    5.00/pieRefSp/{RefSpatial~\cite{zhou2025roborefer}}/0/0/0/0.35/1.45/west/center/1/0,
    79.27/pieUr/UR AI Trainer/263/28/40{,}136/0/-1.28/north/center/0/0}
  \midpiecap{\RHOUAT}{299}
\end{subfigure}\hfill
\begin{subfigure}[t]{0.33\textwidth}
  \centering
  \midpie{108}{%
    5.00/pieRoboPt/{RoboPoint~\cite{yuan2024robopoint}}/0/0/0/-0.3/1.5/south east/center/1/-0.1,
    5.00/pieRefSp/{RefSpatial~\cite{zhou2025roborefer}}/0/0/0/0.3/1.5/south west/center/1/-0.1,
    45.53/pieYamBox/YAM Box/146/40/956/0.05/-1.18/north west/center/0/0,
    44.47/pieYamWild/{YAM Box\\[-2pt](in the wild)}/143/25/1{,}661/-0.05/-1.18/north east/center/0/0}
  \midpiecap{\RHOYAM}{289}
\end{subfigure}

%% file: adaptation.tex
\section{Vertical task adaptation}
\label{sec:adaptation}

\Ethm produces a reusable platform-specific checkpoint, but reliable deployment still requires vertical specialization to the target task. We study two complementary forms of vertical adaptation: offline supervised learning from task demonstrations and online learning from interaction. The online methods operate in the model's latent action space and are evaluated as deployment-time extensions of \RHO.

\subsection{Offline supervised adaptation}

Offline adaptation fine-tunes either pretrained \RHO\ or an embodiment-midtrained checkpoint on demonstrations of the target tasks. We retain the flow-matching objective, action representation, and image augmentation used during pretraining, and use the downstream-specific state and action normalization described in each evaluation protocol. We optimize the task policy with AdamW. Unlike pretraining and embodiment midtraining, this stage uses only target-task robot demonstrations; we do not interleave vision-language batches. All trainable parameters, including the backbone and action expert, use a single peak learning rate of $10^{-4}$. The optimization budget and schedule are selected for each downstream setting and reported with its evaluation protocol. The controlled experiments in \Cref{sec:midtraining-eval} hold the downstream data and optimization budget fixed while varying the initialization, isolating the contribution of embodiment midtraining.

\subsection{Online adaptation in latent action space}
\label{sec:online-adaptation}

Offline task adaptation is limited by the states represented in its demonstration
dataset: once deployed, a policy can encounter configurations in which its
learned behavior is unreliable.  \RHO{} addresses these failures through an
optional internal latent policy.  Let $G_\theta(o,z)$ denote its flow-matching
action generator, which maps an observation $o$ and latent initial condition $z$
to an action chunk.  Standard inference draws $z$ from a Gaussian distribution.
For online adaptation, \RHO{} instead samples from a learned,
observation-conditioned latent policy $\pi_\phi(o)$,
\begin{equation}
    a = G_\theta\!\left(o,\pi_\phi(o)\right).
\end{equation}
The vision-language backbone and action generator $G_\theta$ remain frozen; only
the lightweight latent policy is updated.  The latent policy is a submodule of
\RHO{}, stored and served in the same checkpoint rather than applied as an
external policy wrapper.

We train this latent policy from corrective interaction.  For an observation
$o$ and corrected action chunk $a^\star$, we invert the frozen flow to recover a
latent target $z^\star \approx G_\theta^{-1}(o,a^\star)$ and regress
$\pi_\phi(o)$ toward that target.  The corrected chunk may combine autonomous
actions with actions supplied after a human or scripted-expert takeover; in
either case, inversion translates the behavior that should have been executed
into supervision in \RHO's native latent action space.  Online learning therefore
changes which initial conditions \RHO{} supplies to its pretrained flow, while
every resulting action chunk is still generated by the unchanged offline-trained
backbone and action expert.  This provides a compact adaptation interface and
retains the behavioral structure learned from offline data.  Related work also
uses generative-policy latents for corrective supervision or reward-driven
adaptation~\citep{murray2026flowdagger,wagenmaker2025steering}; our experiments
study the corrective-supervision setting.

\Cref{sec:online-results} evaluates this procedure, known as \FD~\cite{murray2026flowdagger}, in simulation and on
\fr3d. Flow inversion, fixed-point refinement, the internal latent-policy
architecture, correction interfaces, and optimization settings are specified in
\Cref{app:online}.

%% file: exps.tex
\section{Empirical evaluation}
\label{sec:exps}

\RHO's empirical evaluation analyzes questions:

\begin{enumerate}[label=(\roman*)]
  \item Which aspects of \RHO's architectural, training, and data recipes matter for the data efficiency of its downstream adaptation?
  \item What data efficiency gains can we expect for task adaptation from \RHO's embodiment-midtrained versions?
  \item How performant are task policies derived from \RHO's midtrained checkpoints using standard offline supervised finetuning, compared to task policies finetuned from strong existing VLAs?
  \item What gains in task success rate can we expect from adapting \RHO models using online human feedback?
\end{enumerate}

We organize the evaluations to answer these questions around the main stages of \RHO's training and adaptation pipeline. We first isolate architectural and pretraining choices in Sections \ref{sec:architecture-ablations}-\ref{sec:pretraining-ablations}, showing that the shape of \RHO's action expert, the inclusion of VQA data into the pretraining mixture, and the \RHO-Tomyum pretraining mixture as a whole make a major difference for downstream task performance (Tables \ref{tab:action-expert-shape}-\ref{tab:action-expert-compression}, \Cref{fig:pretraining-adaptation}). We then measure the effect of embodiment midtraining in \Cref{sec:midtraining-eval}: when compared to pretrained-only \RHO\ finetuned on target tasks, midtrained \RHO\ tends to achieve the same success rates using \emph{half} as much data under the same compute budget (\Cref{fig:robotwin-midtraining}). Moreover, an experiment on a physical \fr3d shows that the positive effect of midtraining increases with the amount of compute, as measured by the number of midtraining epochs (\Cref{fig:fr3-midtraining}).

\begin{figure}[h]
  \centering
  \captionsetup{skip=\vlaCaptionSkip}
  \input{figures/fig-roboeval}
  \caption{Success rates of finetuned \RHO, \PI, \GR, and \MO on RoboEval tasks. \RHO outperforms all other models on \emph{Pick Book}, \emph{Stack Book Shelf}, and \emph{Rotate Valve}. On each of the remaining 5 tasks, \RHO matches the performance of the best baseline on that task. Importantly, each task's best-performing baseline is generally different, so the next-strongest model, \PI, matches \RHO's performance on only 4 out of 8 tasks.}
  \label{fig:roboeval}
\end{figure}
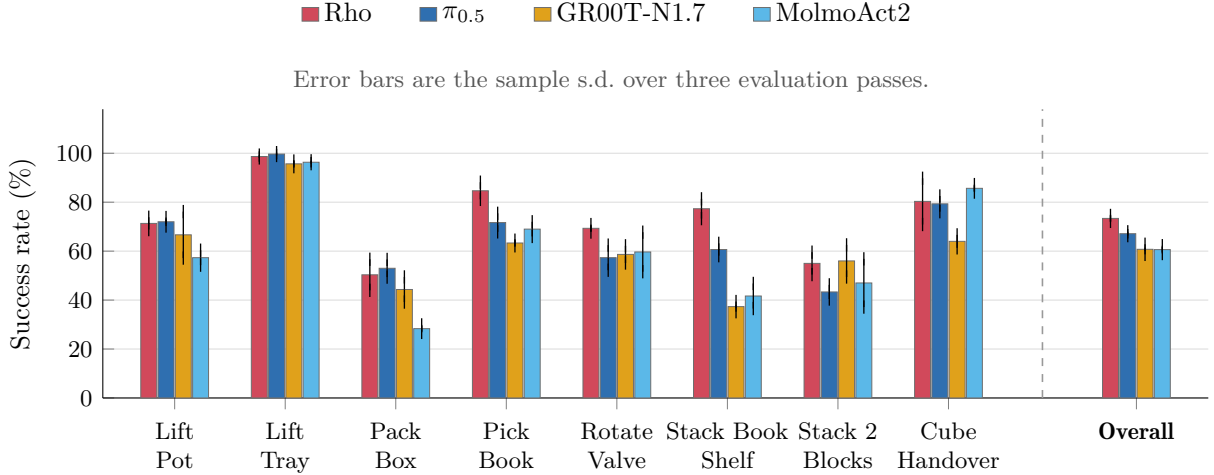

\begin{figure}[h]
  \centering
  \captionsetup{skip=\vlaCaptionSkip}
  \input{figures/fig-ur-ai-trainer}
  \caption{UR AI Trainer results. \RHOUAT achieves higher success than both baselines on both tasks. On \emph{Small-electronics cleanup}, \PI achieves similar mean task progress (80.8\% versus 82.5\%) despite lower success (40.0\% versus 60.0\%). \PI's failures tend to occur during final placement, whereas \RHO's failures more often occur during picking.}
  \label{fig:ur-ai-trainer}
\end{figure}
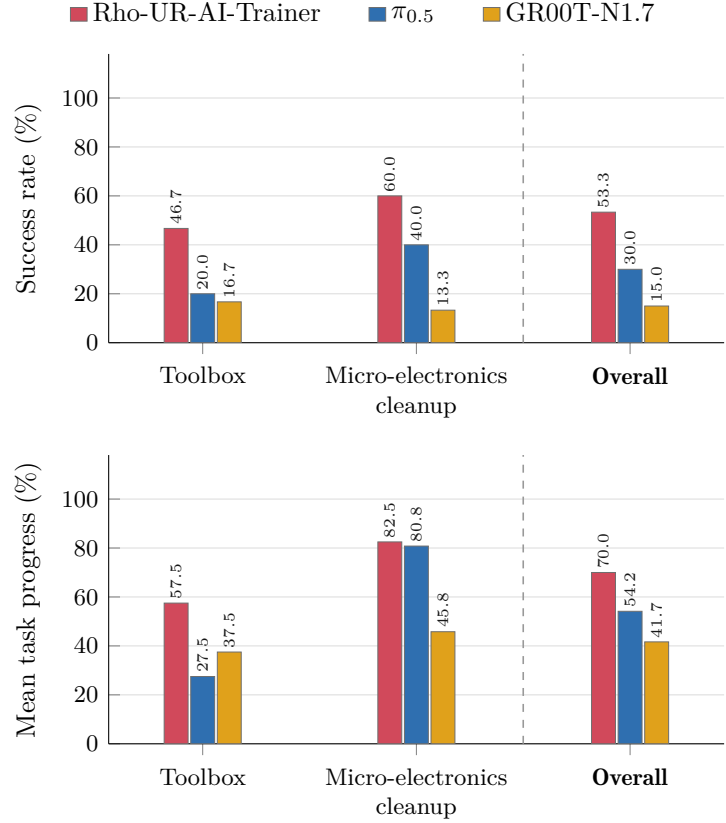

In Sections \ref{sec:simulation-results} and \ref{sec:physical-results} we analyze the performance of \RHO and its midtrained versions across several simulated environments and physical robots, showing major improvements in success rates yielded by the \RHO family over the strongest open-weights VLAs and noting the task properties that influence the size of performance gap between \RHO and its baselines (Figures \ref{fig:roboeval} - \ref{fig:fr3-duo}, \Cref{tab:libero-main}).  The final experiments, in \Cref{sec:online-results}, study online supervised adaptation of \RHO from human feedback and show the data-efficiency of combining it with \Cref{sec:physical-results}'s prior offline finetuning from a small amount of demonstrations.

Taken together, our experiments identify the most data-efficient method for adapting \RHO: finetuning \RHO starting from an embodiment-midtrained checkpoint offline via conventional offline SFT followed by rapid performance improvement through online latent-space adaptation from a few human corrections~\citep{murray2026flowdagger,wagenmaker2025steering}.

\subsection{Action architecture ablations}
\label{sec:architecture-ablations}

We evaluate the design choices introduced in \Cref{sec:architecture} under a common downstream-adaptation and evaluation protocol. Each model is pretrained for approximately one pass over a 15\% subset of the \RHO-Tomyum mixture and then adapted with its corresponding action-expert architecture. These controlled pretraining runs use 16 NVIDIA B200 GPUs, with global robot and physical-grounding batch sizes of 1,024 and 128, respectively. Downstream adaptation uses four NVIDIA B200 GPUs with a global batch size of 128. We evaluate all eight RoboEval tasks using 100 rollouts per task in each of three evaluation passes. We first isolate the action-expert shape and then reduce parameters while preserving the dimensions identified by this sweep.

\begin{table}[H]
  \centering
  \small
  \begin{tabular}{@{}lrrrrr@{}}
    \toprule
    Variant & Width & Heads & Head dim. & Blocks & Success \\
    \midrule
    1,024 / 16 heads / 16 blocks & 1,024 & 16 & 64 & 16 & 0.617 \\
    1,024 / 8 heads / 16 blocks & 1,024 & 8 & 128 & 16 & 0.640 \\
    1,024 / 8 heads / 32 blocks & 1,024 & 8 & 128 & 32 & 0.642 \\
    2,048 / 8 heads / 16 blocks & 2,048 & 8 & 256 & 16 & 0.623 \\
    2,048 / 16 heads / 16 blocks & 2,048 & 16 & 128 & 16 & 0.673 \\
    \bottomrule
  \end{tabular}
  \caption{Action-expert shape sweep on RoboEval after 40k adaptation steps. Success rates are averaged over eight tasks and three evaluation passes.}
  \label{tab:action-expert-shape}
\end{table}

\Cref{tab:action-expert-shape} identifies two important dimensions. First, a head dimension of 128 performs best at both widths: at width 1,024, eight 128-dimensional heads outperform sixteen 64-dimensional heads (0.640 versus 0.617), while at width 2,048, sixteen 128-dimensional heads outperform eight 256-dimensional heads (0.673 versus 0.623). Second, width remains important after fixing the head dimension at 128: the 2,048-wide model reaches 0.673, compared with 0.640--0.642 for the 1,024-wide models. Doubling the depth of the 1,024-wide expert from 16 to 32 blocks does not close this gap.

The sweep identifies a useful balance: 128-dimensional heads provide sufficient representational resolution for fine continuous geometry, while the wider 16-head representation allows many spatial, visual, and contact relationships to be modeled in parallel. This balance complements the flow-matching objective: the action expert regresses a conditional velocity field over continuous action chunks, and precision tasks admit only a narrow set of successful trajectories. Resolving how the field should change around this narrow set places greater demands on representational fidelity than a high-tolerance task with many successful actions.

The task-level results are consistent with this interpretation: the gains from the 2,048-wide expert are concentrated on difficult contact and precision tasks, including lifting a flat book from the table, stacking blocks, and placing a single book on a shelf.

Having identified width 2,048 and head dimension 128 as the important shape parameters, we preserve both while reducing capacity elsewhere. \Cref{tab:action-expert-compression} summarizes this compression sequence.

\begin{table}[H]
  \centering
  \small
  \begin{tabular}{@{}lrrrrr@{}}
    \toprule
    Variant & Q/KV heads & Blocks & AdaLN & AE params. & Success \\
    \midrule
    Wide baseline & 16/16 & 16 & per-block & 1.649B & 0.673 \\
    Shared AdaLN & 16/16 & 16 & shared & 882M & 0.673 \\
    Grouped-query & 16/4 & 16 & shared & 693M & 0.672 \\
    \RHO & 16/4 & 12 & shared & 542M & 0.671 \\
    \bottomrule
  \end{tabular}
  \caption{Action-expert compression on RoboEval while preserving width 2,048 and head dimension 128. Success rates are averaged over eight tasks after 40k adaptation steps. Each variant uses three evaluation passes.}
  \label{tab:action-expert-compression}
\end{table}

Sharing the AdaLN modulation maps reduces the action expert from 1.649B to 882M parameters while matching the baseline performance. Grouped-query attention retains 0.672 success with 693M parameters, and the 12-block configuration retains $0.671 \pm 0.008$ success with 542M parameters. Together, these reductions preserve the action-expert shape selected above while reducing its parameter count by approximately two thirds.

We also test routing successive action-expert blocks to different \phiphy\ hidden states. After 40k adaptation steps, layerwise cross-attention reaches $0.605 \pm 0.008$ RoboEval success, compared with $0.673 \pm 0.015$ when every block attends to the same projected hidden sequence. \RHO{} therefore uses the representation from \phiphy\ decoder block 14 throughout the action expert, yielding higher accuracy with a simpler, more memory-efficient interface.

The same architecture choices remain strong on LIBERO, where the evaluated variants are close to the benchmark ceiling. At 40k adaptation steps, the 1,024-wide baseline reaches 0.973 success, the 2,048-wide expert with shared AdaLN reaches 0.978, and with the final large-scale pretraining checkpoint, the 12-block architecture reaches 0.983. We treat LIBERO as corroborating evidence here: its saturated success rates make it less discriminative than RoboEval for selecting among action-expert variants.

\subsection{Pretraining recipe ablations}
\label{sec:pretraining-ablations}

We next evaluate how multi-embodiment pretraining affects downstream adaptation. For these controlled experiments, each pretraining run uses approximately one pass over a 15\% subset of the robot-data mixture. We then adapt the resulting checkpoints on LIBERO and RoboEval using the same data, optimization budget, and evaluation protocol for every initialization. Results are averaged over three evaluations; LIBERO uses seeded evaluations and reports success separately for each of its four suites, while RoboEval uses complete evaluation passes and averages success over all eight tasks.

\begin{figure}[h]
  \centering
  \input{figures/pretraining_adaptation}
  \vspace{0.3cm}
  \caption{Pretraining on 15\% of the dataset mixture improves downstream adaptation. Left: LIBERO suite success after 20k task-adaptation steps. Right: RoboEval success throughout adaptation. The robot+VL condition combines robot trajectories with physical-grounding examples, the robot-data-only condition removes the physical-grounding batches, and the no-pretraining condition begins task adaptation from the grounded backbone with a randomly initialized action expert. Values are means and standard deviations over three evaluations, except the robot+VL RoboEval result at 40k adaptation steps, which pools six passes from two independent evaluations.}
  \label{fig:pretraining-adaptation}
\end{figure}
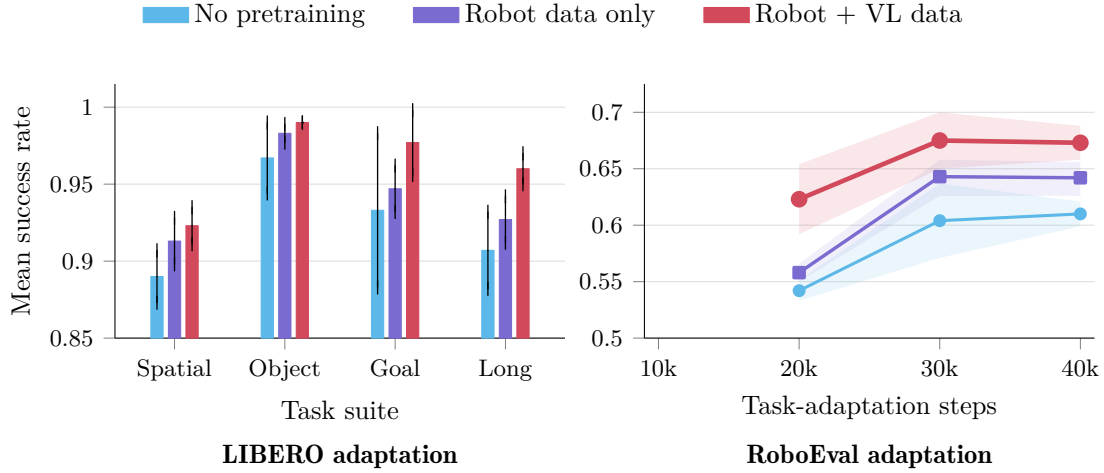

\Cref{fig:pretraining-adaptation} shows that the robot+VL pretraining condition improves downstream performance across the LIBERO suites and both adaptation speed and final performance on RoboEval. At 20k task-adaptation steps, robot+VL reaches 0.963 average LIBERO success, compared with 0.943 for robot-data-only pretraining and 0.924 without pretraining. On RoboEval, robot+VL reaches 0.623 after 20k task-adaptation steps and 0.673 after 40k, versus 0.558 and 0.652 for robot-data-only pretraining and 0.542 and 0.610 without pretraining.

The suite-level view in \Cref{fig:pretraining-adaptation} shows that physical-grounding co-training improves all four LIBERO suites at 20k adaptation steps. Relative to robot-only pretraining, success increases from 0.913 to 0.923 on LIBERO-Spatial, 0.983 to 0.990 on LIBERO-Object, 0.947 to 0.977 on LIBERO-Goal, and 0.927 to 0.960 on LIBERO-Long. LIBERO-Object is already near saturation, while the largest gains occur on Goal and Long, where adaptation must connect the instruction and scene to goal-directed, multi-stage behavior. Physical-grounding supervision therefore provides its clearest benefit on tasks with greater semantic and temporal structure.

Having established the benefit of the robot+VL mixture, we next isolate optimization of the action expert. The action modules are randomly initialized and trained at a substantially higher learning rate than the pretrained \phiphy\ backbone, making their learning rate an important control on how aggressively the model acquires continuous-control structure from the heterogeneous robot mixture. We sweep the action-expert learning rate while retaining physical-grounding data and holding the data mixture, backbone learning rate, schedule, and pretraining budget fixed.

\begin{table}[H]
  \centering
  \small
  \begin{tabular}{@{}lccc@{}}
    \toprule
    Pretraining learning rate & 20k adaptation & 30k adaptation & 40k adaptation \\
    \midrule
    $5\times10^{-5}$ & $0.601 \pm 0.028$ & $0.645 \pm 0.005$ & $0.641 \pm 0.006$ \\
    $1\times10^{-4}$ & $\mathbf{0.623} \pm 0.031$ & $\mathbf{0.675} \pm 0.035$ & $\mathbf{0.673} \pm 0.015$ \\
    $2\times10^{-4}$ & $0.586 \pm 0.010$ & $0.629 \pm 0.031$ & $0.633 \pm 0.008$ \\
    $5\times10^{-4}$ & $0.548 \pm 0.006$ & $0.637 \pm 0.021$ & $0.622 \pm 0.022$ \\
    \bottomrule
  \end{tabular}
  \caption{Effect of the action-expert learning rate during pretraining on downstream RoboEval adaptation. All rows are pretrained on the same 15\% subset of the dataset mixture and include physical-grounding data. The results are mean success and standard deviation over three evaluation passes.}
  \label{tab:pretraining-learning-rate}
\end{table}

\Cref{tab:pretraining-learning-rate} selects $10^{-4}$: it performs best at every RoboEval evaluation point and reaches 0.673 success after 40k adaptation steps, compared with 0.633 at $2\times10^{-4}$ and 0.622 at $5\times10^{-4}$. The gap persists throughout downstream adaptation, showing that the pretraining learning rate affects the quality of the transferred initialization rather than only early adaptation speed. We therefore use $10^{-4}$ for the \RHO{} pretraining recipe.

\subsection{Embodiment midtraining experiments}
\label{sec:midtraining-eval}

We study two complementary effects of horizontal midtraining. First, in a controlled RoboTwin 2.0 simulation experiment, we measure the effect of \RHO's robot midtraining on the data efficiency of offline task finetuning, showing that midtraining halves the amount of offline finetuning data needed to achieve a given success rate across the full range of finetuning data regimes, or allows the task policies finetuned from midtrained \RHO to achieve significantly higher success rates than those finetuned directly from \RHOB. Second, in a physical \fr3d experiment we keep the amount of target-task finetuning data fixed and measure the influence of the amount of midtraining \emph{compute}, as measured by the number of midtraining epochs, on real-world task-adaptation efficiency.

\subsubsection{Embodiment midtraining and data efficiency of downstream adaptation}
\label{sec:robotwin-midtraining}

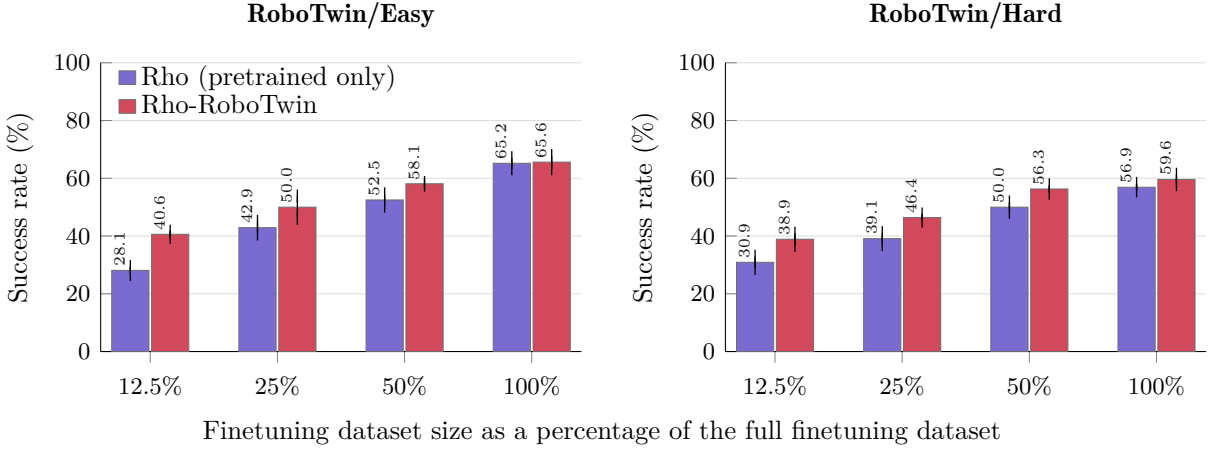
\begin{figure}[h]
  \centering
  \captionsetup{skip=\vlaCaptionSkip}
  \input{figures/fig-robotwin-midtraining}
  \caption{Embodiment midtraining on RoboTwin transfers to held-out tasks across downstream-data
  budgets. Each checkpoint is adapted for 40k steps and evaluated on the matched five-task protocol
  with 50 episodes per task. Bars are means over completed seeds (four for Easy, three for Hard) and
  error bars are sample standard deviations. The benefit of midtraining is largest in the low-data
  regime and shrinks as the adaptation set grows.}
  \label{fig:robotwin-midtraining}
\end{figure}

We use RoboTwin 2.0~\citep{chen2025robotwin} for a controlled assessment of \ethm. RoboTwin provides demonstration data for 50 tasks across five simulated robot embodiments. We select the dual-arm UR5e as the target bimanual embodiment and divide RoboTwin's data for 50 tasks on the dual-UR5e robot into a disjoint embodiment-midtraining subset of 45 tasks and a finetuning subset of the remaining 5 tasks: \emph{Hanging Mug}, \emph{Move Stapler to Pad}, \emph{Place Burger and Fries on Tray}, \emph{Microphone Handover}, and \emph{Stack Three Bowls}. We then compare task adaptation efficiency of \RHOB and a \RHO checkpoint midtrained on the 45 tasks, which we denote \emph{\RHO-RoboTwin}. 

Namely, we obtain \RHO-RoboTwin by midtraining \RHOB on the 45 tasks and, for each held-out task, construct nested demonstration subsets containing 12.5\%, 25\%, 50\%, and 100\% of the available data. For each finetuning data fraction, we combine all 5 tasks datasets into a single dataset, use these datasets for finetuning \RHOB and \RHO-RoboTwin under the same 40k-step optimization budget, and evaluate the resulting 5-task policies on the same held-out tasks under both the Easy and Hard RoboTwin protocols. The full details of the experiment protocol are provided in \Cref{app:sim-eval}.

\Cref{fig:robotwin-midtraining} reports the results. Across all finetuning data budgets until performance saturation, midtraining reduces the amount of finetuning data needed for achieving a given success rate by half: e.g., the midtrained \RHO adapted on 25\% of the finetuning data performs as well as the base \RHO adapted on 50\% of the finetuning data. For a different perspective, in the low-finetuning-data regime of 12.5\%, the midtrained \RHO performs ~30\% better than the finetuned base \RHO. These effects hold in Easy and Hard evaluation regimes, the latter of which includes distractor objects in the scene. 

Although the precise relationship between the amounts of data required by \RHOB and its embodiment-midtrained variants to achieve the same task success rate may vary depending on the variability of the task environment, the composition of the finetuning datasets, our experiments on physical robots in \Cref{sec:physical-results} provide additional evidence that benefits of embodiment midtraining hold broadly.

\subsubsection{Midtraining compute and task adaptation efficiency} 
\label{sec:fr3-midtraining-eval}

We additionally measure the effect of the number of midtraining epochs, i.e., the amount of compute midtraining uses, on task adaptation efficiency. We conduct this experiment on a physical \fr3d robot. We begin from a pretrained \RHOB\ checkpoint and either adapt it directly or first midtrain it for 1 or 2 epochs on the \fr3d mixture (\Cref{fig:midtraining-mixture}) and then adapt each of the resulting models for 50k steps on the same bimanual plug-insertion data. 

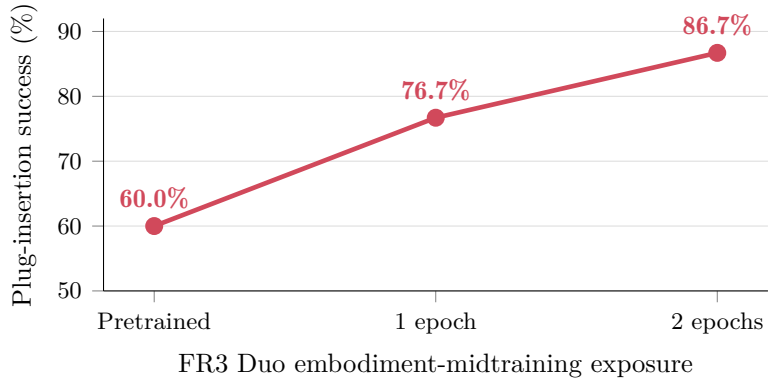
\begin{figure}[h]
  \centering
  \input{figures/fr3_midtraining_adaptation}
  \caption{\fr3d horizontal midtraining improves subsequent vertical adaptation to bimanual plug insertion. Every condition uses the same 50k-step task-adaptation budget; success reported over 30 trials.}
  \label{fig:fr3-midtraining}
\end{figure}

As shown in \Cref{fig:fr3-midtraining}, success over 30 evaluation trials increases from 18/30 (60.0\%) without horizontal midtraining to 23/30 (76.7\%) after one epoch and 26/30 (86.7\%) after two epochs. Qualitatively, much of the improvement appears on initial plug configurations near the edge of the task-training distribution, including rare or unusual positions within the workspace. We hypothesize that broad \fr3d midtraining gives the policy a stronger model of the platform's reachability and motion geometry, helping it plan through workspace regions not represented in the plug-insertion demonstrations.

\subsection{\RHO and existing VLAs: performance comparison in simulation}
\label{sec:simulation-results}

In this section, we compare the performance of task-adapted \RHOB policies and policies finetuned from the strongest existing open-weight VLAs on LIBERO~\citep{liu2023libero} and RoboEval~\citep{wang2026roboeval}. For each benchmark, we follow its standard protocol as detailed in \Cref{app:sim-eval} and compare against baselines trained with matched benchmark data.

On RoboEval, we compare task-adapted \RHOB against our matched finetuning evaluations of MolmoAct2, GR00T N1.7, and $\pi_{0.5}$. We finetune every model on the same task data for 40k optimizer steps with global batch size 128, using each model's native training objective. \RHO{} and MolmoAct2 use eight flow samples per training example, whereas the $\pi_{0.5}$ and GR00T implementations use their native single-sample objectives. The complete adaptation and evaluation protocol is given in \Cref{app:sim-eval}. \RHOB reaches 73.4\% average success on RoboEval, outperforming the next-best baseline by 6.3 percentage points. It achieves the highest task-level success on 3 of the 8 tasks, with the largest margin on \emph{Stack Single Book on Shelf}.

\begin{table}[H]
  \centering
  \small
  \setlength{\tabcolsep}{4.5pt}
  \begin{tabular}{@{}lrrrrrrrrr@{}}
    \toprule
    Model & Succ. $\uparrow$ & TP $\uparrow$ & BGVD $\downarrow$ & CPL $\downarrow$ & ECC $\downarrow$ & JPL $\downarrow$ & OPL $\downarrow$ & SCC $\downarrow$ & SC $\downarrow$ \\
    \midrule
    MolmoAct2 & 0.61 & 0.79 & \textbf{0.072} & 1.48 & 1.03 & 7.97 & 4.84 & \textbf{0.017} & \textbf{0.465} \\
    GR00T N1.7 & 0.61 & 0.80 & 0.078 & 1.66 & 1.25 & 9.64 & 5.43 & 0.095 & 0.540 \\
    $\pi_{0.5}$ & 0.67 & 0.82 & 0.074 & 1.45 & \textbf{0.99} & 7.98 & 4.82 & 0.038 & 0.482 \\
    \midrule
    \textbf{\RHO{}} & \textbf{0.73} & \textbf{0.85} & 0.073 & \textbf{1.41} & 1.02 & \textbf{7.70} & \textbf{4.69} & 0.035 & 0.477 \\
    \bottomrule
  \end{tabular}

  \vspace{4pt}
  \begin{minipage}{0.98\linewidth}
  \scriptsize
  Succ.: success rate; TP: task progression; BGVD: bimanual gripper vertical difference (m); CPL: Cartesian path length (m); ECC: environment collision count; JPL: joint path length (rad); OPL: orientation path length (rad); SCC: self-collision count; SC: slip count.
  \end{minipage}
  \caption{Aggregate RoboEval outcome and behavioral metrics after 40k task-adaptation steps with global batch size 128. \RHO outperforms the baselines on 5 out of the 9 metrics, with the next-strongest model across these metrics, \MO, winning on 3.  Every row is the mean of 3 800-episode evaluations across the 8 task variations. \emph{Joint Path length (JPL)} measures the sum of joint path lengths traced by the two arms. Arrows indicate the preferred direction of each metric, e.g., ``$\downarrow$'' means that lower is better.}
  \label{tab:roboeval-behavioral}
\end{table}

Beyond binary success on RoboEval tasks, \Cref{tab:roboeval-behavioral} reports task progression and behavioral metrics for coordination, motion efficiency, and safety. In addition to mean success rate, \RHO{} achieves the highest task progression and the shortest Cartesian, joint, and orientation paths; $\pi_{0.5}$ has the fewest environment collisions, while MolmoAct2 has the lowest bimanual gripper vertical difference, self-collision count, and slip count. We provide the full task-level breakdown in \Cref{app:roboeval-behavioral}.

For LIBERO, we evaluate a policy initialized from \RHOB and adapted in two stages. We first finetune the full model on the official LIBERO demonstrations using offline supervised adaptation. We then freeze the backbone and action expert and online-adapt only the internal latent policy, using the benchmark's training initial states and corrective supervision from a scripted LIBERO expert, as described in \Cref{sec:online-adaptation}. We compare the resulting policy against representative open VLA baselines with publicly reported LIBERO results in \Cref{tab:libero-main}. For OpenVLA, $\pi_0$, and \MO, we cite their success rates from the original model reports~\cite{kim2024openvla,black2024pi0,lee2025molmoact}. The $\pi_{0.5}$ values come from the official OpenPi LIBERO release~\citep{pi2026libero}. The GR00T N1.7 results come from NVIDIA's official LeRobot evaluation~\citep{nvidia2026groot17}. The adapted \RHOB policy reaches 97.9\% average success across LIBERO's 4 task suites, the strongest aggregate result among the compared models. Its margin comes almost entirely from LIBERO-Long, the hardest suite, where it reaches 95.9\% and exceeds the next-best model by 2.7 percentage points; on the three shorter-horizon suites it is within 0.5 points of the best reported value.

\begin{table}[H]
  \centering
  \small
  \setlength{\tabcolsep}{5.2pt}
  \begin{tabular}{@{}lccccc@{}}
    \toprule
    Model & Spatial & Object & Goal & Long & Average \\
    \midrule
    OpenVLA~\citep{kim2024openvla} & 84.7 & 88.4 & 79.2 & 53.7 & 76.5 \\
    $\pi_0$~\citep{black2024pi0} & 96.8 & 98.8 & 95.8 & 85.2 & 94.2 \\
    MolmoAct-7B-D~\citep{lee2025molmoact} & 87.0 & 95.4 & 87.6 & 77.2 & 86.6 \\
    GR00T N1.7~\citep{nvidia2026groot17} & 95.0 & \textbf{100.0} & \textbf{98.0} & 93.0 & 96.5 \\
    $\pi_{0.5}$~\citep{pi2026libero} & \textbf{98.8} & 98.2 & \textbf{98.0} & 92.4 & 96.9 \\
    MolmoAct2~\citep{fang2026molmoact2} & 97.8 & \textbf{100.0} & 97.8 & 93.2 & 97.2 \\
    \midrule
    \textbf{\RHO} & 98.3 & 99.6 & 97.6 & \textbf{95.9} & \textbf{97.9} \\
    \bottomrule
  \end{tabular}
  \caption{\RHO's mean performance and published success rates of several baselines on the 4 task suites of LIBERO~\cite{liu2023libero}. \RHO outperforms the baselines in overall success rate aggregated across all 4 task suites and in success rate on the most challenging suite, \emph{LIBERO-long}.  Baseline values are taken from the corresponding model report or official release as cited in column 1; evaluation and adaptation protocols may differ across the reports. For \RHO, the mean success rate is over 3 seeds. All values are percentages. \textbf{Bold} denotes the best result in each column.}
  \label{tab:libero-main}
\end{table}

\subsection{\RHO and existing VLAs: performance comparison on physical robots}
\label{sec:physical-results}

In the central set of experiments of this report, we evaluate task adaptation of \RHO-family models on three representative stationary dual-arm robot embodiments: \fr3d, \urait, and \yb (\Cref{fig:physical-embodiments}). Unlike the evaluations in \Cref{sec:simulation-results}, which focused on adapting \RHOB, the physical-robot experiments examine the adaptability of \RHOYAM, \RHOUAT, and \RHOFR\ -- the \RHOB derivatives midtrained for the robots we use. On each embodiment, we compare the corresponding \RHO variant against \PI and \GR; on \yb, we additionally evaluate against \MO, since this model has significant amounts of data from a dual-arm YAM robot similar to \yb in its training mixture (720 hours). Each model is adapted independently on the same task-specific demonstrations before evaluation. The comparisons are carried out according to the parallel A/B evaluation~\cite{kressgazit2024robotlearningempiricalscience}, which minimizes the effect of environment drift and the choice of initial configurations of the task environments on the experiment outcome. Robot specifications and full evaluation protocols are provided in
\Cref{app:robots,app:real-eval}.

The evaluations span advanced stacking and pick-and-place tasks, packing and sorting, insertion, and interaction with deformable as well as articulated objects -- all requiring precision during execution. Together, these tasks reflect realistic manipulation challenges and probe long-horizon bimanual coordination, precise control, generalization across environment configurations, and a broad range of manipulation skills.

\subsubsection{\yb}
\label{sec:yb-exps}

\begin{figure}[t]
  \centering
  \captionsetup{skip=\vlaCaptionSkip}
  \input{figures/fig-yam-box-tasks}
  \caption{The 6 BusyBox task categories used in the \yb experiments. \emph{Bimanual} tasks require two-arm coordination; \emph{dual-arm} tasks require choosing the correct arm for task completion.}
  \label{fig:yam-box-tasks}
\end{figure}
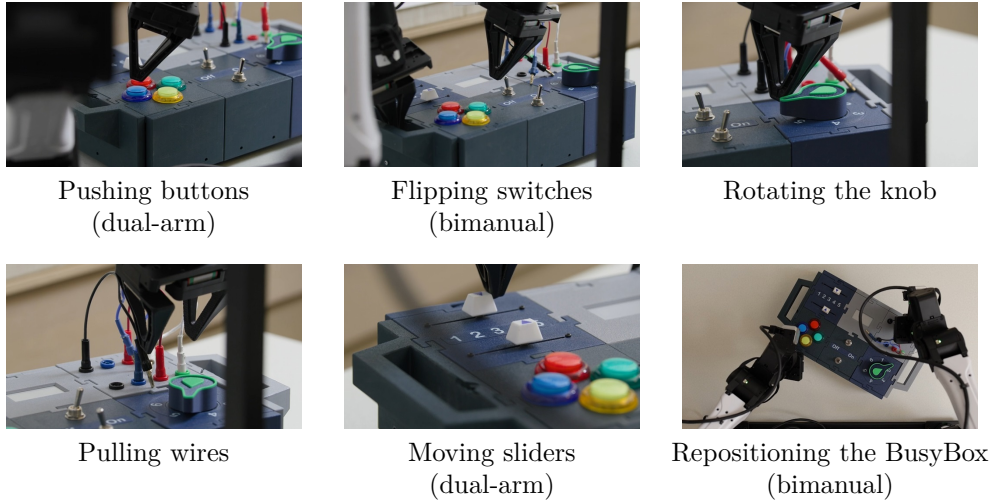

The \yb study considers a high-data finetuning regime: we finetune \PI, \GR, \MO, and \RHOYAM\ on $\sim 2000$ demonstrations representing 6 task categories from the BusyBox~\citep{fortier2026busybox} manipulation benchmark (\Cref{fig:yam-box-tasks}), which we describe here for completeness:

\begin{itemize}
    \item \textbf{Pushing buttons.} The robot must press one of four colored, illuminated buttons with a specified gripper (e.g., ``Push the blue button with the left gripper'').
    \item \textbf{Flipping switches.} The robot must flip a specified switch on or off using the specified arm (left or right), as described in the task instruction. Because flipping requires considerable force, one arm must pin the BusyBox in place while the other operates the switch, making the task bimanual.
    \item \textbf{Moving sliders.} The robot must move the top or the bottom slider, as dictated by the task instruction, to a specified position between 1 and 5. The task instruction doesn't specify with which arm this should be done, so the task involves selecting the correct slider, choosing the arm, and positioning the slider accurately.
    \item \textbf{Rotating the knob.} The robot must turn the knob to a specified position between 1 and 6, which can be made difficult by occlusions.
    \item \textbf{Pulling wires.} The robot must grasp and unplug a wire of a specified color (red, black, blue, or white) from the wire module.
    \item \textbf{Repositioning the BusyBox.} The robot must rotate the entire BusyBox clockwise or counter-clockwise, or slide it left, right, closer, or away. The task variants involving rotations are bimanual.
\end{itemize}

The full finetuning dataset for \yb that we used in this experiment is available at \url{https://huggingface.co/datasets/microsoft/BusyBox}.

We pool the data from all tasks and finetune \RHOYAM\ on the entire dataset for 50k steps with a global batch size of 128. The baselines were finetuned with their recommended finetuning settings in their respective repositories or reports OpenPi~\cite{openpi2024}, GR00T N1.7~\cite{gr00tn17_2025}, and \MO~\cite{fang2026molmoact2}. We evaluate each finetuned model's success rate over 60 rollouts, 10 rollouts per task category. Each of a task's 10 rollouts uses a different instruction (e.g., \emph{``Flip the bottom switch on with the left gripper''} vs. \emph{``Flip the bottom switch off with the right gripper''}) and a different initial configuration of the BusyBox, which determines the initial states of all of the BusyBox controls and the position and rotation of the BusyBox with respect to the robot. 

\begin{figure}[h!]
  \centering
  \captionsetup{skip=\vlaCaptionSkip}
  \input{figures/fig-yam-box}
  \caption{\yb BusyBox results. After task-specific finetuning, \RHOYAM achieves the same 90\% mean success rate as \PI and a higher mean success rate than \GR and \MO across 60 matched evaluation rollouts.}
  \label{fig:yam-box}
\end{figure}
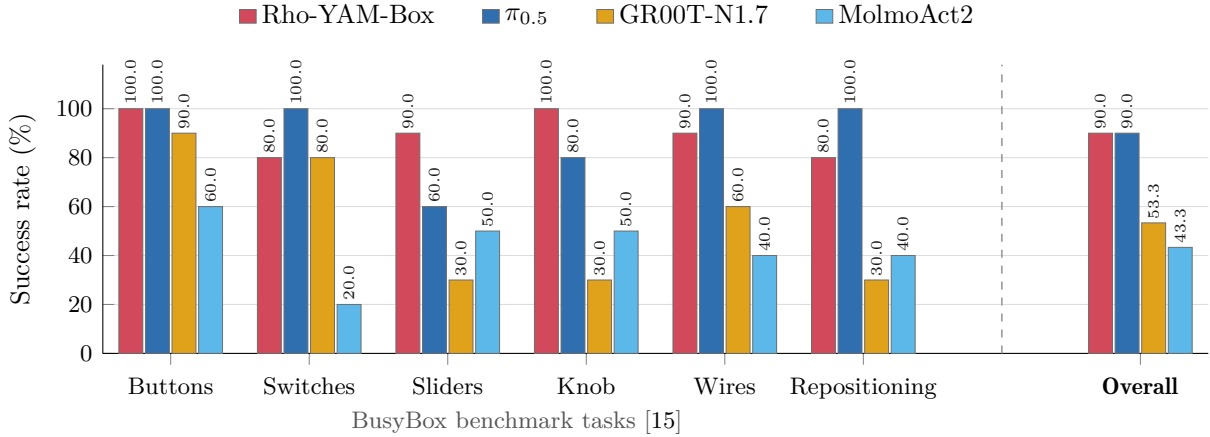

Despite a lot of variability in the initial states of the BusyBox tasks as well as the precision and dexterity of manipulation required to complete them, each task generally takes an experienced teleoperator less than 15 seconds to complete. During evaluation, we set the task horizon to 900 steps, corresponding to 30 seconds on a \yb operating at 30 Hz. Thus, the tasks are fairly short-horizon, and the 2000 high-quality demonstrations across all tasks constitute a sufficiently large dataset to learn the tasks well.

\paragraph{Results analysis.}

As shown in \Cref{fig:yam-box}, \RHOYAM achieves 90\% mean success rate across the BusyBox suite, the same observed mean as \PI and a higher observed mean than \GR and \MO. This high-data experiment shows that, when task-specific finetuning data is plentiful, \RHOYAM can achieve robust adaptation performance comparable to the highest-performing open-weights baselines evaluated here. \PI and \MO also have prior training exposure to related dual-arm robot embodiments, making them strong comparisons in this setting.

\subsubsection{\urait}
\label{sec:urait-exps}

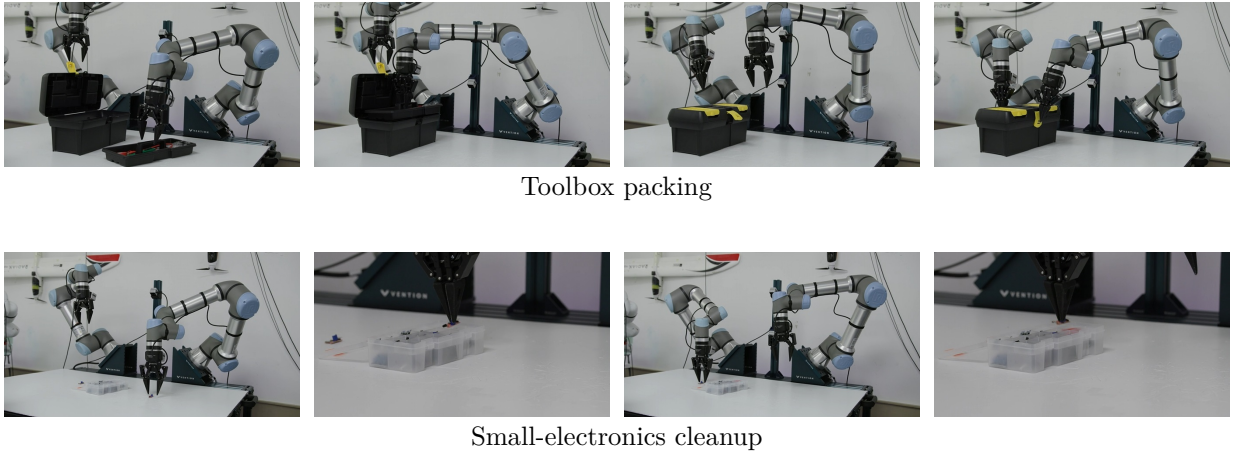
\begin{figure}[t]
  \centering
  \captionsetup{skip=\vlaCaptionSkip, justification=centering}
  \input{figures/fig-ur-ai-trainer-tasks}
  \caption{The \urait tasks and their stages. Each row shows task progression from left to right.}
  \label{fig:ur-ai-trainer-tasks}
\end{figure}

\paragraph{Tasks and finetuning protocol.}
In the experiments on UR AI Trainer, we evaluate \RHO's performance on adaptation to multi-stage tasks in low-data finetuning regime. We adapt \RHO, \PI, and \GR to two target tasks (\Cref{fig:ur-ai-trainer-tasks}), training a separate policy for each task. \RHO's adaptation begins from the UR AI Trainer-midtrained checkpoint, \RHOUAT, which captures the platform's observations, kinematics, and control conventions.

\begin{itemize}
    \item \textbf{Toolbox packing} ($\sim 150$ demonstrations). The workspace contains an open toolbox and a tray filled with materials and tools. The robot must coordinate both arms across 4 task stages: lifting the tray, fitting it into the toolbox without tilting it or spilling its contents, closing the lid, and engaging the latch. Precision is paramount: without careful alignment of the tray with the toolbox sides, the toolbox's lid won't close. For the purpose of calculating task progress, completing each stage earns the score of 0.25. The execution horizon during evaluation was 1080 steps (60 seconds at 18 Hz), but all models finished their rollouts earlier.
    \item \textbf{Small-electronics cleanup} ($\sim 160$ demonstrations). Two small electronic components (roughly, 1.5 inches long and 0.5 inches wide) are arbitrarily positioned on a tabletop in  the beside a compartmentalized tray. All of the tray's compartments but 2 are filled with other small electronics. The robot must pick and place each component on the table into an empty compartment of the tray. Despite being a pick-and-place task it presents difficulty due to the electronics pieces having small grippable surface, comparable in size to \urait's gripper tips, and the tray compartments being small as well. As in \emph{Toolbox packing}, each stage (picking up and placing each component into an empty compartment) yields 0.25 points toward task progress. The task horizon during execution was 1620 steps (90 seconds at 18 Hz). 
\end{itemize}


\paragraph{Results analysis.}

The finetuned \RHOUAT has the highest observed success rate on both tasks: 46.7\% on \emph{Toolbox packing}, compared with 20.0\% for \PI and 16.7\% for \GR, and 60.0\% on \emph{Small-electronics cleanup}, compared with 40.0\% and 13.3\%, respectively. These results are consistent with the motivation for separating embodiment adaptation from task adaptation, particularly on long-horizon, precision-demanding tasks with limited task-specific demonstrations.

At the same time, the long-horizon precision-demanding nature of both tasks makes them hard to learn from only 150-200 demonstrations even for \RHOUAT, as indicated by all models' absolute success rates. Task progress provides an additional perspective on relative performance. On \emph{Small-electronics cleanup}, \RHOUAT and \PI achieve similar observed mean task progress (82.5\% and 80.8\%, respectively), although \PI's failures tend to happen in the task's concluding stages (placing electronics into the correct compartment).

\subsubsection{\fr3d}
\label{sec:fr3d-exps}

\begin{figure}[t]
  \centering
  \captionsetup{skip=\vlaCaptionSkip}
  \input{figures/fig-fr3-duo-tasks}
  \caption{The \fr3d tasks and their stages. The Test-tube and Tumbler tasks are evaluated in both directions: their frames read from left to right for the task on the upper arrow and from right to left for the task on the lower arrow.}
  \label{fig:fr3-duo-tasks}
\end{figure}
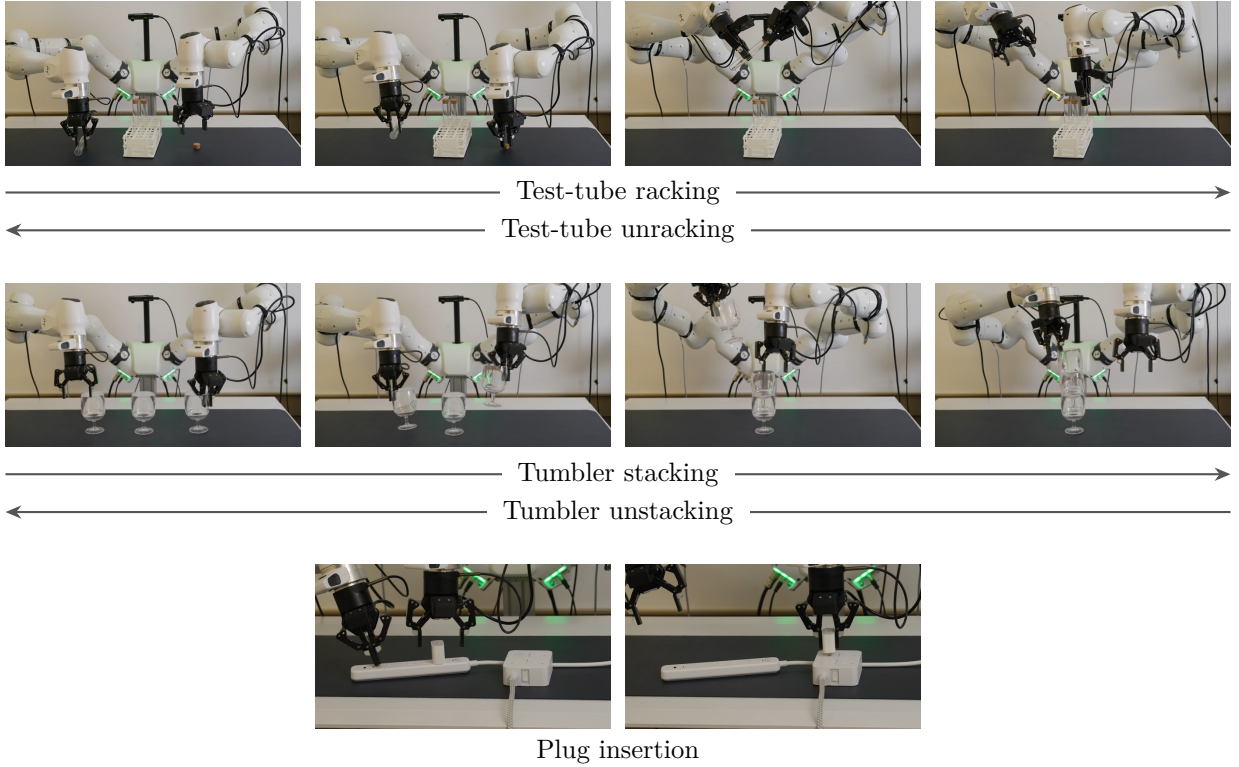

\paragraph{Tasks and finetuning protocol.}
On \fr3d, we evaluate task adaptation across three task families selected to
exercise complementary capabilities of the platform: long-horizon bimanual
coordination, object--object alignment and stability, and precise small-object
contact. Plug insertion contributes one evaluation variant, while tumbler and
test-tube manipulation are each evaluated in both directions, yielding five
variants from three finetuning datasets (\Cref{fig:fr3-duo-tasks}):
\begin{itemize}
    \item \textbf{Plug insertion} ($\sim$150 demonstrations) requires
    one arm to stabilize a power strip while the other removes, reorients, and
    inserts the adapter into a second strip. This tests distinct bimanual roles
    over a long sequence ending in precise alignment. For measuring task progress, we break the task into 2 stages (picking up the plug and inserting it correctly) and assign 0.5 points for each.
    \item \textbf{Tumbler stacking and unstacking} use $\sim$100
    demonstrations, divided evenly between the two directions. The tall, easily
    disturbed objects test coordinated alignment and stability during stacking,
    as well as controlled separation and placement during unstacking. Both forward and backward task flavors consist of 4 stages (picking up each of 2 tumblers and placing it, either on the tabletop or on the tumbler below it), earning 0.25 points per stage.
    \item \textbf{Test-tube racking and unracking} use $\sim$300
    demonstrations, again divided evenly. These variants combine small-object
    handling and cork manipulation with constrained insertion into or extraction
    from a rack. These tasks, too, consist of 4 stages of picking up/placing the tube and the cork, inserting/removing the cork into/from the tube, and putting the closed tube into the rack/removing it from the rack, with 0.25 task progress points for each stage.
\end{itemize}

The paired tumbler and test-tube variants let us compare precision assembly with
the generally less constrained reverse behavior while keeping the objects and
workspace fixed. For each dataset, the finetuned \RHO is initialized from the
\fr3d-midtrained checkpoint \RHOFR, while \PI{} and \GR{} begin from their respective
pretrained checkpoints. We adapt all three models on the same demonstrations for
50k optimizer steps with a global batch size of 64. All five task variants use a maximum rollout horizon of 30 seconds, equivalent
to 900 control steps at the 30-Hz control rate.

For each task variant and model, we evaluate 30 rollouts from distinct initial
configurations. The matched configuration set varies the positions and orientations
of the adapter and power strips for plug insertion, the starting tumbler poses for
stacking and unstacking, and the tube, cork, and rack poses for racking and
unracking. Every model is evaluated on the same configurations.

\begin{figure}[h]
  \centering
  \captionsetup{skip=\vlaCaptionSkip}
  \input{figures/fig-fr3-duo}
  \caption{\fr3d task-adaptation results. Each model is adapted independently
  on each task dataset and evaluated over 30 distinct, matched initial configurations per variant.
  We report binary success rate and mean task progress; Overall is the unweighted
  mean across the five task variants.}
  \label{fig:fr3-duo}
\end{figure}
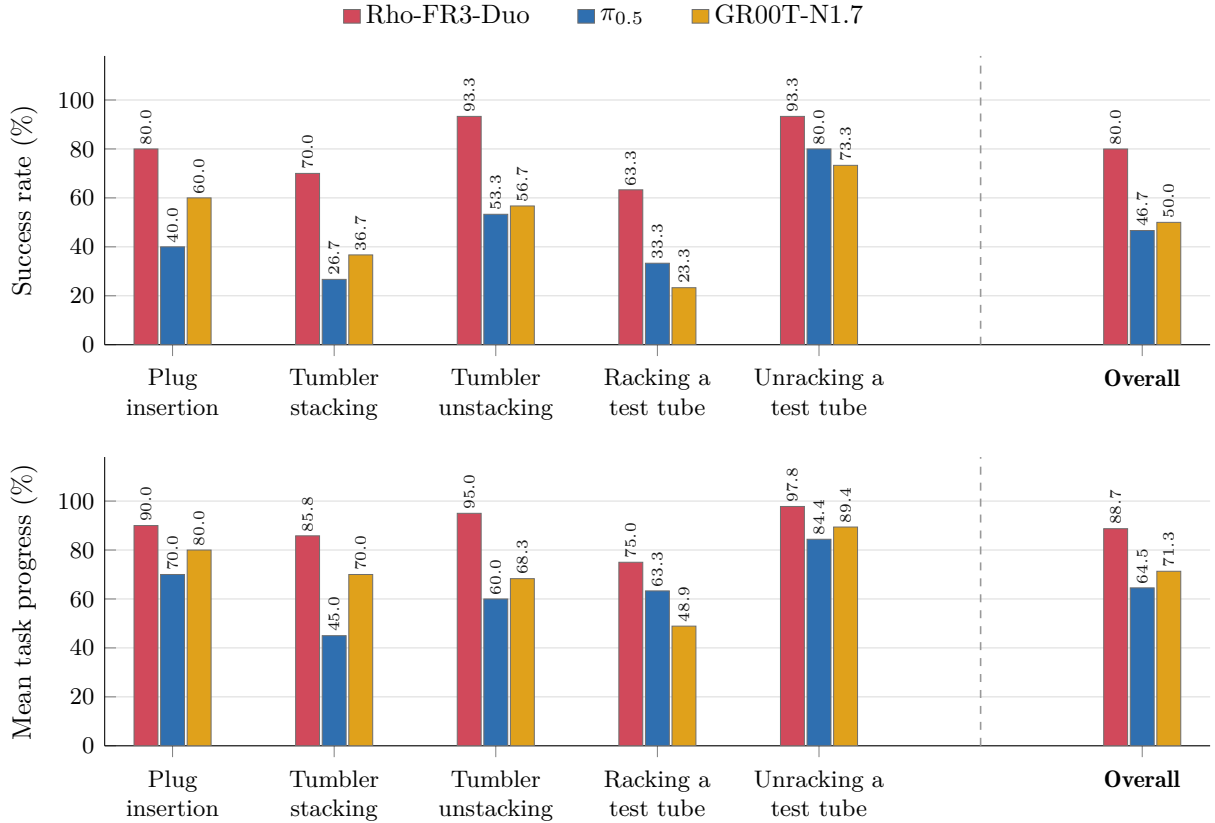

\paragraph{Results analysis.}

As shown in \Cref{fig:fr3-duo}, \RHOFR{} achieves 80.0\% mean success and
88.7\% mean task progress across the five variants, compared with 50.0\%
success and 71.3\% progress for \GR{} and 46.7\% success and 64.5\% progress
for \PI{}. More importantly, the per-task success rates reflects the physical
difficulty of the suite. Stacking is harder than unstacking and racking is
harder than unracking for all 3 models: the forward task variants require the
policy to establish and maintain precise alignment under contact, whereas the
reverse tasks provide more room for error once the objects have been separated. \RHOFR{}
has its largest observed margins on these alignment-sensitive behaviors, exceeding
the highest baseline point estimate by 33.3 percentage points on \emph{Tumbler Stacking} and by
30.0 points on test-tube racking. Its point estimate is also 20.0 points higher on plug insertion,
where success depends on coordinating distinct roles for the two arms over a
long sequence.

Task progress adds useful context to the remaining failures. On \emph{Tumbler Stacking},
for example, \RHOFR{} reaches 85.8\% mean progress despite 70.0\% binary success;
on test-tube racking it reaches 75.0\% progress and 63.3\% success. This pattern
is consistent with many unsuccessful rollouts reaching the later alignment or
placement stages rather than failing at the initial grasp. \RHOFR{} has higher observed mean
progress and success than both baselines on every variant, suggesting that the
gain is not limited to crossing the final success threshold.

Together with the controlled midtraining study in
\Cref{sec:fr3-midtraining-eval}, these results illustrate the intended role of
embodiment midtraining. A task-specific demonstration set necessarily covers
only a narrow portion of the robot's reachable workspace. A model adapted
directly from a general pretrained checkpoint must therefore learn the task
while also inferring the platform's reachability and motion geometry from that
limited slice of experience. \RHOFR{}, in contrast, has already encountered a
broader range of \fr3d poses and motions during embodiment midtraining. This
provides a stronger starting point when evaluation places objects near the edge
of, or outside the configurations represented in, the task-finetuning data.
The advantage is consistent with \RHOFR's gains across the pose-varied
evaluation configurations, particularly on behaviors that require traversing
the workspace before establishing precise contact.

Like the tasks from the \urait experiments in \Cref{sec:urait-exps}, the \fr3d tasks are generally difficult to master from as few as 150 demonstrations collected offline from a fixed initial-state distribution. In the next section, we show how online adaptation using human feedback provided online for \RHO's task policies learned during offline finetuning can be a much more data-efficient way of achieving task proficiency than offline finetuning alone.

\subsection{Online-adaptation results}
\label{sec:online-results}

The final experiments begin from offline task-adapted \RHO\ policies and measure how much further they improve through online adaptation in latent space (\Cref{sec:online-adaptation}). We enable \RHO's internal latent policy, a learned sampler that predicts the initial condition of its frozen flow-matching action expert and is stored in the same model checkpoint. This steers the generated action chunks without changing the pretrained model parameters~\citep{murray2026flowdagger,wagenmaker2025steering}. In all experiments below, we use \FD~\cite{murray2026flowdagger} as the online adaptation method, and the feedback signal is corrective supervision. On the physical robot a human operator takes over with a SpaceMouse; in simulation a scripted expert with access to privileged simulator state takes over. We assemble the actions actually executed during each query period into a complete chunk, traverse the discretized flow in reverse, and apply fixed-point refinement at each inverse step. \RHO's latent policy is trained to predict the resulting initial-condition targets~\citep{murray2026flowdagger}. Full algorithmic details and hyperparameters are given in \Cref{app:online}. 
We evaluate this stage in simulation on Meta-World~\citep{yu2020metaworld} and on two \fr3d tasks.

\paragraph{Simulation.} To obtain starting policies with substantial room for improvement, we train \RHO\ with a deliberately limited budget on the full 50-task Meta-World suite and then adapt it in latent space on six challenging manipulation tasks that exercise precise insertion and extraction, tool use, and object placement, using 50 additional episodes per task supervised by Meta-World's privileged scripted expert. \Cref{tab:flowdagger-metaworld} reports success over 100 evaluation rollouts per task. Latent-space adaptation raises average success from 57.5\% to 73.3\%. The largest gains are on \emph{Assembly} (45\% to 78\%) and \emph{Stick pull} (58\% to 78\%), and on \emph{Hammer} adaptation closes the remaining gap from 92\% to 100\%. Pick Out of Hole remains the hardest task, improving from 22\% to 37\%, while the gain on \emph{Basketball} is small (+6 points).

\begin{table}[H]
  \centering
  \small
  \begin{tabular}{@{}lccc@{}}
    \toprule
    Task & Offline & + Online & $\Delta$ \\
    \midrule
    Assembly         & 45.0 & 78.0  & +33.0 \\
    Stick pull       & 58.0 & 78.0  & +20.0 \\
    Pick out of hole & 22.0 & 37.0  & +15.0 \\
    Hand insert      & 73.0 & 86.0  & +13.0 \\
    Hammer           & 92.0 & 100.0 & +8.0 \\
    Basketball       & 55.0 & 61.0  & +6.0 \\
    \midrule
    Average          & 57.5 & 73.3  & +15.8 \\
    \bottomrule
  \end{tabular}
  \caption{Online latent-space adaptation on Meta-World. The offline policy is \RHO\ trained with a limited budget on all 50 Meta-World tasks; the online column adapts it to each task in latent space from 50 episodes supervised by a privileged scripted expert. Success is measured over 100 rollouts per task; all values are percentages.}
  \label{tab:flowdagger-metaworld}
\end{table}

\paragraph{Physical robot.} On \fr3d, we start from \RHOFR\ policies adapted offline to the test-tube assembly and plug-insertion tasks from 150 demonstrations each, adapt them online from 15 additional human-supervised episodes, and evaluate each on a hard subset of initial configurations. Each subset consists of 10 configurations near the edge of the task-training distribution, with objects placed at the edge of the workspace or in unusual orientations; these are the same kinds of configurations on which \fr3d midtraining helped most in \Cref{sec:fr3-midtraining-eval}. The online corrections are collected on these configurations, so this experiment measures how quickly the policy can be adapted to a known weak region of its state distribution. \Cref{tab:flowdagger-real} reports success rate (SR) and task progress (TP) over 30 trials (three per configuration). On test-tube assembly, online adaptation raises success from 9/30 to 21/30, more than doubling the success rate. On plug insertion, where the offline policy is already stronger, it raises success from 20/30 to 28/30, cutting failures from ten to two. In both cases, online supervision amounting to a tenth of the offline demonstration budget substantially improves the policy on exactly the configurations where it struggles most.

\begin{table}[H]
  \centering
  \small
  \begin{tabular}{@{}lcccc@{}}
    \toprule
    & \multicolumn{2}{c}{Offline} & \multicolumn{2}{c}{+ Online} \\
    \cmidrule(lr){2-3} \cmidrule(lr){4-5}
    Task & SR & TP & SR & TP \\
    \midrule
    Test-tube assembly & 30.0 & 56.7 & \textbf{70.0} & \textbf{81.7} \\
    Plug insertion     & 66.7 & 83.3 & \textbf{93.3} & \textbf{96.7} \\
    \bottomrule
  \end{tabular}
  \caption{Online latent-space adaptation on \fr3d. Offline policies are adapted from the \RHOFR\ checkpoint on 150 demonstrations per task; online adaptation uses 15 additional human-supervised episodes. Each task is evaluated over 30 trials, three on each of 10 hard initial configurations. SR is success rate and TP is task progress; both are percentages.}
  \label{tab:flowdagger-real}
\end{table}

%% file: figures/pretraining_adaptation.tex
\colorlet{rhoPlotBlue}{vlaRho}
\colorlet{rhoPlotTeal}{vlaRhoPre}
\colorlet{rhoPlotSlate}{vlaMolmo}
\renewcommand{\vlaLegendCols}{3}

\begin{tikzpicture}
  \begin{groupplot}[
    group style={group size=2 by 1, horizontal sep=1.05cm, group name=pa},
    width=0.465\textwidth,
    height=0.305\textwidth,
    axis x line*=bottom,
    axis y line*=left,
    tick align=outside,
    tick label style={font=\vlaTickFont},
    label style={font=\vlaAxisLabelFont},
    title style={font=\vlaPanelTitleFont, yshift=3pt},
    ymajorgrids=true,
    xmajorgrids=false,
    grid style={vlaGrid, line width=0.4pt},
    axis background/.style={fill=white},
    every axis plot/.append style={line cap=round, line join=round},
  ]
    \nextgroupplot[
      ybar,
      bar width=4.6pt,
      ymin=0.85,
      ymax=1.015,
      ytick={0.85,0.90,0.95,1.00},
      ylabel={Mean success rate},
      xlabel={Task suite},
      symbolic x coords={Spatial,Object,Goal,Long},
      xtick=data,
      enlarge x limits=0.18,
      legend to name=pretraining-legend,
      vla legend,
      error bars/y dir=both,
      error bars/y explicit,
      error bars/error bar style={black, line width=0.5pt},
      error bars/error mark options={mark size=\vlaErrorBarWidth,
                                     line width=0.5pt, black},
    ]
      \addlegendimage{area legend,draw=rhoPlotSlate,fill=rhoPlotSlate}
      \addlegendentry{No pretraining}
      \addlegendimage{area legend,draw=rhoPlotTeal,fill=rhoPlotTeal}
      \addlegendentry{Robot data only}
      \addlegendimage{area legend,draw=rhoPlotBlue,fill=rhoPlotBlue}
      \addlegendentry{Robot + VL data}

      \addplot[draw=rhoPlotSlate, fill=rhoPlotSlate]
        coordinates {
          (Spatial,0.890) +- (0,0.017)
          (Object,0.967) +- (0,0.023)
          (Goal,0.933) +- (0,0.050)
          (Long,0.907) +- (0,0.025)
        };
      \addplot[draw=rhoPlotTeal, fill=rhoPlotTeal]
        coordinates {
          (Spatial,0.913) +- (0,0.015)
          (Object,0.983) +- (0,0.006)
          (Goal,0.947) +- (0,0.015)
          (Long,0.927) +- (0,0.015)
        };
      \addplot[draw=rhoPlotBlue, fill=rhoPlotBlue]
        coordinates {
          (Spatial,0.923) +- (0,0.012)
          (Object,0.990) +- (0,0.000)
          (Goal,0.977) +- (0,0.021)
          (Long,0.960) +- (0,0.010)
        };
    \nextgroupplot[
      xmin=9,
      xmax=41,
      ymin=0.50,
      ymax=0.725,
      xtick={10,20,30,40},
      xticklabels={10k,20k,30k,40k},
      ytick={0.50,0.55,0.60,0.65,0.70},
      xlabel={Task-adaptation steps},
      ylabel={},
    ]
      \addplot[draw=none, name path=robo-scratch-upper, forget plot]
        coordinates {(20,0.551) (30,0.637) (40,0.621)};
      \addplot[draw=none, name path=robo-scratch-lower, forget plot]
        coordinates {(20,0.533) (30,0.571) (40,0.599)};
      \addplot[rhoPlotSlate, fill opacity=0.12, forget plot]
        fill between[of=robo-scratch-upper and robo-scratch-lower];
      \addplot[color=rhoPlotSlate, mark=*, mark options={fill=rhoPlotSlate},
        line width=1.15pt, mark size=1.85pt, forget plot]
        coordinates {(20,0.542) (30,0.604) (40,0.610)};

      \addplot[draw=none, name path=robo-robot-upper, forget plot]
        coordinates {(20,0.566) (30,0.658) (40,0.656)};
      \addplot[draw=none, name path=robo-robot-lower, forget plot]
        coordinates {(20,0.550) (30,0.626) (40,0.626)};
      \addplot[rhoPlotTeal, fill opacity=0.10, forget plot]
        fill between[of=robo-robot-upper and robo-robot-lower];
      \addplot[color=rhoPlotTeal, mark=square*, mark options={fill=rhoPlotTeal},
        line width=1.3pt, mark size=1.9pt, forget plot]
        coordinates {(20,0.558) (30,0.643) (40,0.642)};

      \addplot[draw=none, name path=robo-rho-upper, forget plot]
        coordinates {(20,0.654) (30,0.700) (40,0.688)};
      \addplot[draw=none, name path=robo-rho-lower, forget plot]
        coordinates {(20,0.592) (30,0.650) (40,0.658)};
      \addplot[rhoPlotBlue, fill opacity=0.13, forget plot]
        fill between[of=robo-rho-upper and robo-rho-lower];
      \addplot[color=rhoPlotBlue, mark=*, mark options={fill=rhoPlotBlue},
        line width=1.75pt, mark size=2.25pt, forget plot]
        coordinates {(20,0.623) (30,0.675) (40,0.673)};
  \end{groupplot}
  \node[anchor=north, font=\vlaPanelTitleFont] at ([yshift=-36pt]pa c1r1.south)
    {LIBERO adaptation};
  \node[anchor=north, font=\vlaPanelTitleFont] at ([yshift=-36pt]pa c2r1.south)
    {RoboEval adaptation};
  \node[anchor=south] at ([yshift=0.48cm]current bounding box.north)
    {\ref*{pretraining-legend}};
\end{tikzpicture}

%% file: figures/fig-robotwin-midtraining.tex
%
%
%

\newcommand{\rtEasyMidA}{40.6}  \newcommand{\rtEasyMidAsd}{1.01}
\newcommand{\rtEasyMidB}{50.0}  \newcommand{\rtEasyMidBsd}{3.82}
\newcommand{\rtEasyMidC}{58.1}  \newcommand{\rtEasyMidCsd}{0.38}
\newcommand{\rtEasyMidD}{65.6}  \newcommand{\rtEasyMidDsd}{2.19}

\newcommand{\rtEasyPreA}{28.1}  \newcommand{\rtEasyPreAsd}{1.32}
\newcommand{\rtEasyPreB}{42.9}  \newcommand{\rtEasyPreBsd}{2.15}
\newcommand{\rtEasyPreC}{52.5}  \newcommand{\rtEasyPreCsd}{2.05}
\newcommand{\rtEasyPreD}{65.2}  \newcommand{\rtEasyPreDsd}{1.88}

\newcommand{\rtHardMidA}{38.9}  \newcommand{\rtHardMidAsd}{2.05}
\newcommand{\rtHardMidB}{46.4}  \newcommand{\rtHardMidBsd}{1.20}
\newcommand{\rtHardMidC}{56.3}  \newcommand{\rtHardMidCsd}{1.40}
\newcommand{\rtHardMidD}{59.6}  \newcommand{\rtHardMidDsd}{1.74}

\newcommand{\rtHardPreA}{30.9}  \newcommand{\rtHardPreAsd}{2.05}
\newcommand{\rtHardPreB}{39.1}  \newcommand{\rtHardPreBsd}{2.01}
\newcommand{\rtHardPreC}{50.0}  \newcommand{\rtHardPreCsd}{1.74}
\newcommand{\rtHardPreD}{56.9}  \newcommand{\rtHardPreDsd}{1.22}

\renewcommand{\vlaLegendCols}{1}   

\pgfplotsset{
  robotwin midtraining axis/.style={
    vla bar axis,
    vla error bars,
    width=\vlaFigWidth,
    ymax=100,                
    title style/.append style={yshift=-8pt},
    ylabel style/.append style={yshift=-10pt},
    bar width=14pt,
    enlarge x limits=0.13,
    every node near coord/.append style={yshift=4pt},
    symbolic x coords={dtwelve,dtwentyfive,dfifty,dhundred},
    xtick={dtwelve,dtwentyfive,dfifty,dhundred},
    xticklabels={12.5\%, 25\%, 50\%, 100\%},
    ylabel={Success rate (\%)},
  },
}

\begin{minipage}[t]{0.49\linewidth}
\centering
\begin{tikzpicture}
\begin{axis}[
  robotwin midtraining axis,
  title={RoboTwin/Easy},
  vla legend,
  legend cell align=left,
  legend style={at={(0.02,0.9)}, anchor=west, row sep=-2pt,
                inner ysep=1pt},
]
\addplot[vla series=vlaRhoPre] coordinates {
  (dtwelve,\rtEasyPreA)     +- (0,\rtEasyPreAsd)
  (dtwentyfive,\rtEasyPreB) +- (0,\rtEasyPreBsd)
  (dfifty,\rtEasyPreC)      +- (0,\rtEasyPreCsd)
  (dhundred,\rtEasyPreD)    +- (0,\rtEasyPreDsd)};
\addlegendentry{\RHO (pretrained only)}
\addplot[vla series=vlaRho] coordinates {
  (dtwelve,\rtEasyMidA)     +- (0,\rtEasyMidAsd)
  (dtwentyfive,\rtEasyMidB) +- (0,\rtEasyMidBsd)
  (dfifty,\rtEasyMidC)      +- (0,\rtEasyMidCsd)
  (dhundred,\rtEasyMidD)    +- (0,\rtEasyMidDsd)};
\addlegendentry{\RHORT}
\end{axis}
\end{tikzpicture}
\end{minipage}\hfill
\begin{minipage}[t]{0.49\linewidth}
\centering
\begin{tikzpicture}
\begin{axis}[
  robotwin midtraining axis,
  title={RoboTwin/Hard},
]
\addplot[vla series=vlaRhoPre] coordinates {
  (dtwelve,\rtHardPreA)     +- (0,\rtHardPreAsd)
  (dtwentyfive,\rtHardPreB) +- (0,\rtHardPreBsd)
  (dfifty,\rtHardPreC)      +- (0,\rtHardPreCsd)
  (dhundred,\rtHardPreD)    +- (0,\rtHardPreDsd)};
\addplot[vla series=vlaRho] coordinates {
  (dtwelve,\rtHardMidA)     +- (0,\rtHardMidAsd)
  (dtwentyfive,\rtHardMidB) +- (0,\rtHardMidBsd)
  (dfifty,\rtHardMidC)      +- (0,\rtHardMidCsd)
  (dhundred,\rtHardMidD)    +- (0,\rtHardMidDsd)};
\end{axis}
\end{tikzpicture}
\end{minipage}

\vspace{1pt}
{\centering\vlaAxisLabelFont Finetuning dataset size as a percentage of the full finetuning dataset\par}

%% file: figures/fr3_midtraining_adaptation.tex
\colorlet{fr3PlotBlue}{vlaRho}
\begin{tikzpicture}
  \begin{axis}[
    width=0.64\textwidth,
    height=0.32\textwidth,
    xmin=-0.18,
    xmax=2.18,
    ymin=50,
    ymax=92,
    xtick={0,1,2},
    xticklabels={Pretrained,1 epoch,2 epochs},
    ytick={50,60,70,80,90},
    xlabel={\fr3d embodiment-midtraining exposure},
    ylabel={Plug-insertion success (\%)},
    axis x line*=bottom,
    axis y line*=left,
    tick align=outside,
    tick label style={font=\vlaTickFont},
    label style={font=\vlaAxisLabelFont},
    ymajorgrids=true,
    xmajorgrids=false,
    grid style={vlaGrid, line width=0.4pt},
    axis background/.style={fill=white},
    nodes near coords,
    point meta=explicit symbolic,
    every node near coord/.append style={
      font=\small\bfseries,
      text=fr3PlotBlue,
      anchor=south,
      yshift=2pt,
    },
  ]
    \addplot[
      color=fr3PlotBlue,
      mark=*,
      mark options={fill=fr3PlotBlue},
      line width=1.75pt,
      mark size=2.5pt,
    ] coordinates {
      (0,60.0) [60.0\%]
      (1,76.7) [76.7\%]
      (2,86.7) [86.7\%]
    };
  \end{axis}
\end{tikzpicture}

%% file: figures/fig-roboeval.tex
%
%
%

\newcommand{\reRhoPot}{71.33}      \newcommand{\reRhoPotE}{2.52}
\newcommand{\reRhoTray}{98.67}     \newcommand{\reRhoTrayE}{0.58}
\newcommand{\reRhoPack}{50.33}     \newcommand{\reRhoPackE}{6.35}
\newcommand{\reRhoBook}{84.67}     \newcommand{\reRhoBookE}{3.51}
\newcommand{\reRhoValve}{69.33}    \newcommand{\reRhoValveE}{1.53}
\newcommand{\reRhoShelf}{77.33}    \newcommand{\reRhoShelfE}{4.04}
\newcommand{\reRhoBlocks}{55.00}   \newcommand{\reRhoBlocksE}{4.58}
\newcommand{\reRhoCube}{80.33}     \newcommand{\reRhoCubeE}{9.45}

\newcommand{\rePiPot}{72.00}       \newcommand{\rePiPotE}{1.73}
\newcommand{\rePiTray}{99.67}      \newcommand{\rePiTrayE}{0.58}
\newcommand{\rePiPack}{53.00}      \newcommand{\rePiPackE}{3.61}
\newcommand{\rePiBook}{71.67}      \newcommand{\rePiBookE}{3.79}
\newcommand{\rePiValve}{57.33}     \newcommand{\rePiValveE}{5.13}
\newcommand{\rePiShelf}{60.67}     \newcommand{\rePiShelfE}{2.52}
\newcommand{\rePiBlocks}{43.33}    \newcommand{\rePiBlocksE}{2.89}
\newcommand{\rePiCube}{79.33}      \newcommand{\rePiCubeE}{3.21}

\newcommand{\reGrPot}{66.67}       \newcommand{\reGrPotE}{9.50}
\newcommand{\reGrTray}{95.67}      \newcommand{\reGrTrayE}{1.15}
\newcommand{\reGrPack}{44.33}      \newcommand{\reGrPackE}{5.13}
\newcommand{\reGrBook}{63.33}      \newcommand{\reGrBookE}{1.15}
\newcommand{\reGrValve}{58.67}     \newcommand{\reGrValveE}{3.51}
\newcommand{\reGrShelf}{37.33}     \newcommand{\reGrShelfE}{2.08}
\newcommand{\reGrBlocks}{56.00}    \newcommand{\reGrBlocksE}{6.56}
\newcommand{\reGrCube}{64.00}      \newcommand{\reGrCubeE}{2.65}

\newcommand{\reMoPot}{57.33}       \newcommand{\reMoPotE}{3.06}
\newcommand{\reMoTray}{96.33}      \newcommand{\reMoTrayE}{0.58}
\newcommand{\reMoPack}{28.33}      \newcommand{\reMoPackE}{1.53}
\newcommand{\reMoBook}{69.00}      \newcommand{\reMoBookE}{3.00}
\newcommand{\reMoValve}{59.67}     \newcommand{\reMoValveE}{8.08}
\newcommand{\reMoShelf}{41.67}     \newcommand{\reMoShelfE}{5.13}
\newcommand{\reMoBlocks}{47.00}    \newcommand{\reMoBlocksE}{9.85}
\newcommand{\reMoCube}{85.67}      \newcommand{\reMoCubeE}{1.53}

\pgfmathsetmacro{\reRhoAll}{(\reRhoPot+\reRhoTray+\reRhoPack+\reRhoBook
  +\reRhoValve+\reRhoShelf+\reRhoBlocks+\reRhoCube)/8}
\pgfmathsetmacro{\rePiAll}{(\rePiPot+\rePiTray+\rePiPack+\rePiBook
  +\rePiValve+\rePiShelf+\rePiBlocks+\rePiCube)/8}
\pgfmathsetmacro{\reGrAll}{(\reGrPot+\reGrTray+\reGrPack+\reGrBook
  +\reGrValve+\reGrShelf+\reGrBlocks+\reGrCube)/8}
\pgfmathsetmacro{\reMoAll}{(\reMoPot+\reMoTray+\reMoPack+\reMoBook
  +\reMoValve+\reMoShelf+\reMoBlocks+\reMoCube)/8}

\newcommand{\reRhoAllE}{1.21}
\newcommand{\rePiAllE}{0.78}
\newcommand{\reGrAllE}{2.07}
\newcommand{\reMoAllE}{1.60}

\renewcommand{\vlaLegendCols}{4}
\renewcommand{\vlaBarWidth}{6pt}
\renewcommand{\vlaBarSep}{0.5pt}

\newcommand{\reOverallX}{9.7}
\pgfmathsetmacro{\reDividerX}{(8+\reOverallX)/2}

\vlalegendblock{roboevallegend}
\vlanote{Error bars are the sample s.d.\ over three evaluation passes.}

\begin{tikzpicture}
\begin{axis}[
  vla bar axis,
  vla error bars,
  vla hide values,                 
  vla legend,
  legend to name=roboevallegend,
  ylabel={Success rate (\%)},
  xtick={1,2,3,4,5,6,7,8,\reOverallX},
  enlarge x limits={abs=0.65},
  xticklabels={Lift\\Pot, Lift\\Tray, Pack\\Box, Pick\\Book, Rotate\\Valve,
               Stack Book\\Shelf, Stack 2\\Blocks, Cube\\Handover,
               \textbf{Overall}},
]
\addplot[vla series=vlaRho] coordinates {
  (1,\reRhoPot)       +- (0,\reRhoPotE)
  (2,\reRhoTray)     +- (0,\reRhoTrayE)
  (3,\reRhoPack)     +- (0,\reRhoPackE)
  (4,\reRhoBook)     +- (0,\reRhoBookE)
  (5,\reRhoValve)   +- (0,\reRhoValveE)
  (6,\reRhoShelf)   +- (0,\reRhoShelfE)
  (7,\reRhoBlocks) +- (0,\reRhoBlocksE)
  (8,\reRhoCube)     +- (0,\reRhoCubeE)
  (\reOverallX,\reRhoAll)   +- (0,\reRhoAllE)};
\addlegendentry{\RHO}
\addplot[vla series=vlaPi] coordinates {
  (1,\rePiPot)       +- (0,\rePiPotE)
  (2,\rePiTray)     +- (0,\rePiTrayE)
  (3,\rePiPack)     +- (0,\rePiPackE)
  (4,\rePiBook)     +- (0,\rePiBookE)
  (5,\rePiValve)   +- (0,\rePiValveE)
  (6,\rePiShelf)   +- (0,\rePiShelfE)
  (7,\rePiBlocks) +- (0,\rePiBlocksE)
  (8,\rePiCube)     +- (0,\rePiCubeE)
  (\reOverallX,\rePiAll)   +- (0,\rePiAllE)};
\addlegendentry{\PI}
\addplot[vla series=vlaGroot] coordinates {
  (1,\reGrPot)       +- (0,\reGrPotE)
  (2,\reGrTray)     +- (0,\reGrTrayE)
  (3,\reGrPack)     +- (0,\reGrPackE)
  (4,\reGrBook)     +- (0,\reGrBookE)
  (5,\reGrValve)   +- (0,\reGrValveE)
  (6,\reGrShelf)   +- (0,\reGrShelfE)
  (7,\reGrBlocks) +- (0,\reGrBlocksE)
  (8,\reGrCube)     +- (0,\reGrCubeE)
  (\reOverallX,\reGrAll)   +- (0,\reGrAllE)};
\addlegendentry{\GR}
\addplot[vla series=vlaMolmo] coordinates {
  (1,\reMoPot)       +- (0,\reMoPotE)
  (2,\reMoTray)     +- (0,\reMoTrayE)
  (3,\reMoPack)     +- (0,\reMoPackE)
  (4,\reMoBook)     +- (0,\reMoBookE)
  (5,\reMoValve)   +- (0,\reMoValveE)
  (6,\reMoShelf)   +- (0,\reMoShelfE)
  (7,\reMoBlocks) +- (0,\reMoBlocksE)
  (8,\reMoCube)     +- (0,\reMoCubeE)
  (\reOverallX,\reMoAll)   +- (0,\reMoAllE)};
\addlegendentry{\MO}

\draw[vlaRule, dashed, line width=0.6pt]
  (axis cs:\reDividerX,0) -- (axis cs:\reDividerX,\vlaYMax);
\end{axis}
\end{tikzpicture}

%% file: figures/fig-yam-box-tasks.tex
%
%

\vlasetphotowidth{4}%
\vlaphotolabeled{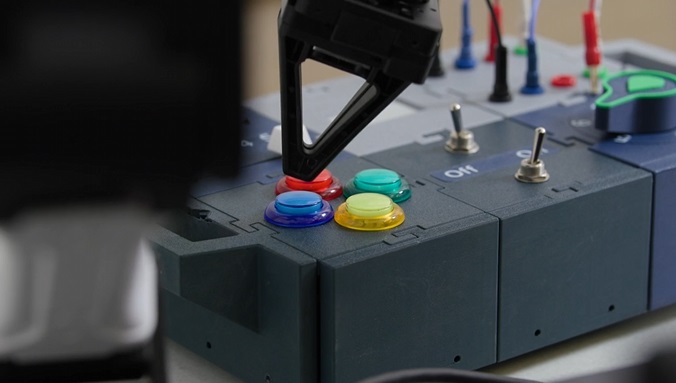}{Pushing buttons\\(dual-arm)}\hspace{\vlaPhotoGapWide}%
\vlaphotolabeled{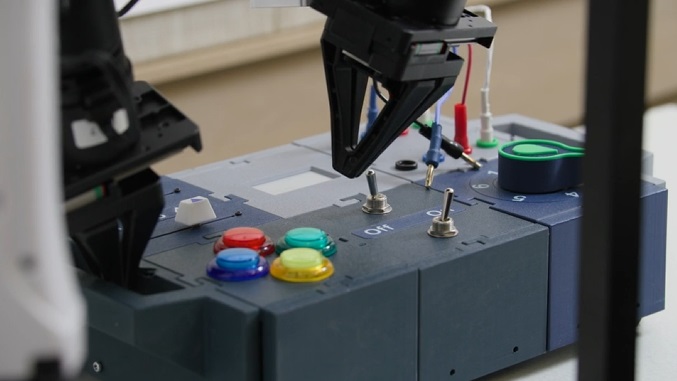}{Flipping switches\\(bimanual)}\hspace{\vlaPhotoGapWide}%
\vlaphotolabeled{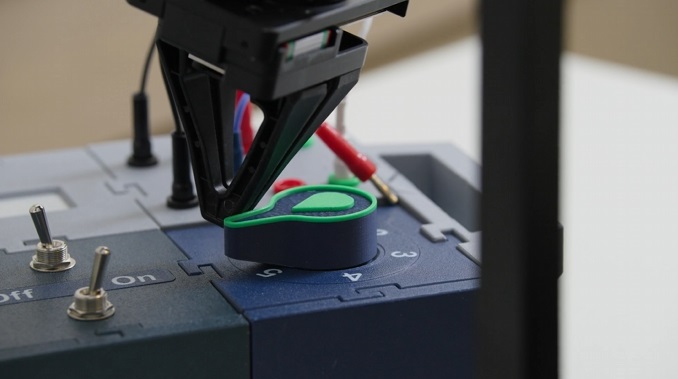}{Rotating the knob}%
\par\vspace{\vlaPhotoRowSkip}%
\vlaphotolabeled{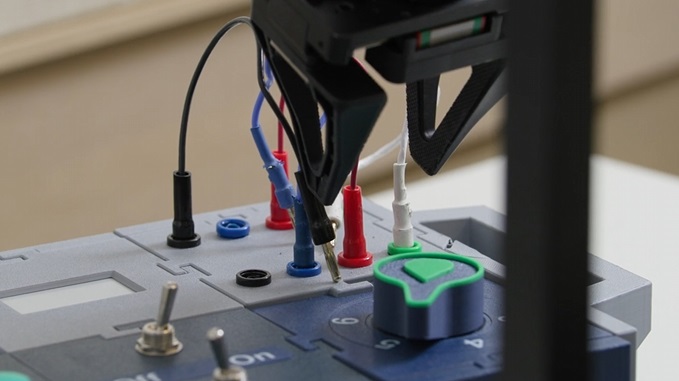}{Pulling wires}\hspace{\vlaPhotoGapWide}%
\vlaphotolabeled{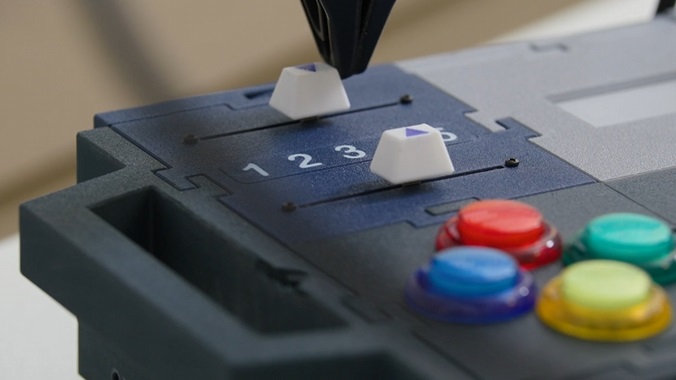}{Moving sliders\\(dual-arm)}\hspace{\vlaPhotoGapWide}%
\vlaphotolabeled{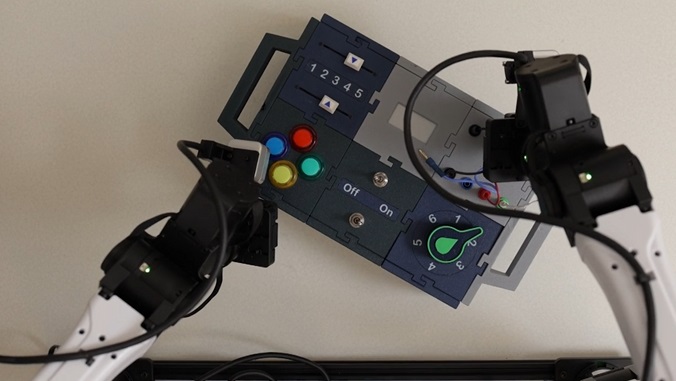}{Repositioning the BusyBox\\(bimanual)}%
\par

%% file: figures/fig-yam-box.tex
%
%

\newcommand{\ybRhoButton}{100.0}  \newcommand{\ybPiButton}{100.0}  \newcommand{\ybGrButton}{90.0}  \newcommand{\ybMoButton}{60.0}
\newcommand{\ybRhoSwitch}{80.0}   \newcommand{\ybPiSwitch}{100.0}  \newcommand{\ybGrSwitch}{80.0}  \newcommand{\ybMoSwitch}{20.0}
\newcommand{\ybRhoSlider}{90.0}   \newcommand{\ybPiSlider}{60.0}   \newcommand{\ybGrSlider}{30.0}  \newcommand{\ybMoSlider}{50.0}
\newcommand{\ybRhoKnob}{100.0}    \newcommand{\ybPiKnob}{80.0}     \newcommand{\ybGrKnob}{30.0}    \newcommand{\ybMoKnob}{50.0}
\newcommand{\ybRhoWire}{90.0}     \newcommand{\ybPiWire}{100.0}    \newcommand{\ybGrWire}{60.0}    \newcommand{\ybMoWire}{40.0}
\newcommand{\ybRhoRepos}{80.0}    \newcommand{\ybPiRepos}{100.0}   \newcommand{\ybGrRepos}{30.0}   \newcommand{\ybMoRepos}{40.0}

\pgfmathsetmacro{\ybRhoAll}{(\ybRhoButton+\ybRhoSwitch+\ybRhoSlider+\ybRhoKnob+\ybRhoWire+\ybRhoRepos)/6}
\pgfmathsetmacro{\ybPiAll} {(\ybPiButton +\ybPiSwitch +\ybPiSlider +\ybPiKnob +\ybPiWire +\ybPiRepos )/6}
\pgfmathsetmacro{\ybGrAll} {(\ybGrButton +\ybGrSwitch +\ybGrSlider +\ybGrKnob +\ybGrWire +\ybGrRepos )/6}
\pgfmathsetmacro{\ybMoAll} {(\ybMoButton +\ybMoSwitch +\ybMoSlider +\ybMoKnob +\ybMoWire +\ybMoRepos )/6}

\renewcommand{\vlaLegendCols}{4}

\vlalegendblock{yamboxlegend}

\begin{tikzpicture}
\begin{axis}[
  vla bar axis,
  vla legend,
  legend to name=yamboxlegend,
  clip=false,                        
  ylabel={Success rate (\%)},
  symbolic x coords={button,switch,slider,knob,wire,repos,gap,overall},
  xmin=button, xmax=overall,
  xtick={button,switch,slider,knob,wire,repos,overall},
  xticklabels={Buttons, Switches, Sliders, Knob, Wires, Repositioning, \textbf{Overall}},
]
\addplot[vla series=vlaRho] coordinates {
  (button,\ybRhoButton) (switch,\ybRhoSwitch) (slider,\ybRhoSlider)
  (knob,\ybRhoKnob) (wire,\ybRhoWire) (repos,\ybRhoRepos) (overall,\ybRhoAll)};
\addlegendentry{\RHOYAM}
\addplot[vla series=vlaPi] coordinates {
  (button,\ybPiButton) (switch,\ybPiSwitch) (slider,\ybPiSlider)
  (knob,\ybPiKnob) (wire,\ybPiWire) (repos,\ybPiRepos) (overall,\ybPiAll)};
\addlegendentry{\PI}
\addplot[vla series=vlaGroot] coordinates {
  (button,\ybGrButton) (switch,\ybGrSwitch) (slider,\ybGrSlider)
  (knob,\ybGrKnob) (wire,\ybGrWire) (repos,\ybGrRepos) (overall,\ybGrAll)};
\addlegendentry{\GR}
\addplot[vla series=vlaMolmo] coordinates {
  (button,\ybMoButton) (switch,\ybMoSwitch) (slider,\ybMoSlider)
  (knob,\ybMoKnob) (wire,\ybMoWire) (repos,\ybMoRepos) (overall,\ybMoAll)};
\addlegendentry{\MO}

\vladivider{gap}
\vlagrouplabel{button}{repos}{BusyBox benchmark tasks~\citep{fortier2026busybox}}
\end{axis}
\end{tikzpicture}

%% file: figures/fig-ur-ai-trainer-tasks.tex
%
%

\vlasetphotowidth{4}%
\vlaphoto{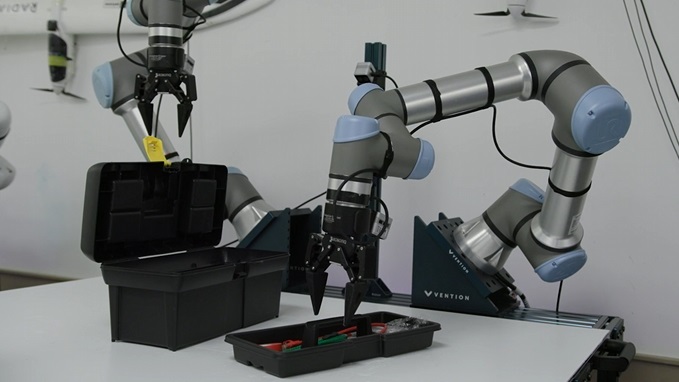}\hfill
\vlaphoto{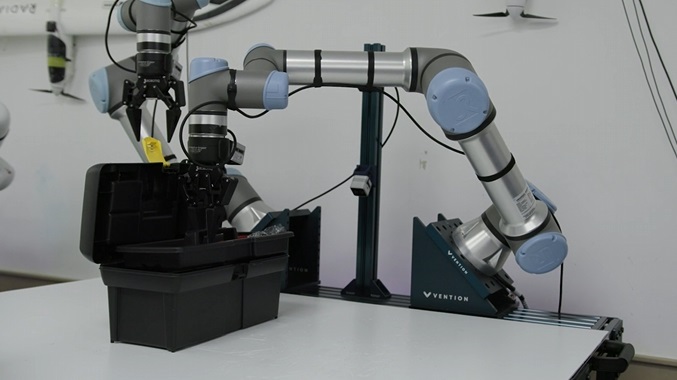}\hfill
\vlaphoto{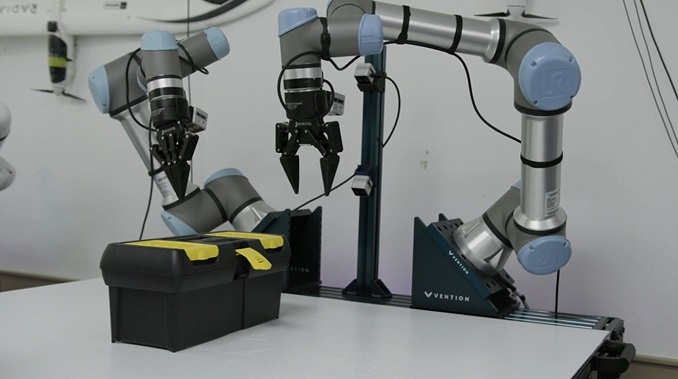}\hfill
\vlaphoto{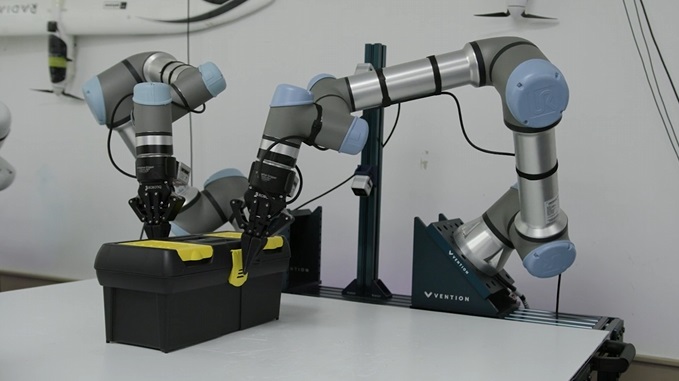}%
\vlarowlabel{Toolbox packing}%
\vspace{\vlaPhotoRowSkip}%
\vlaphoto{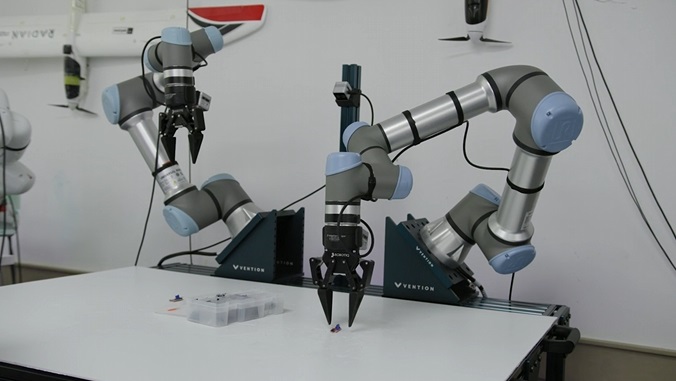}\hfill
\vlaphoto{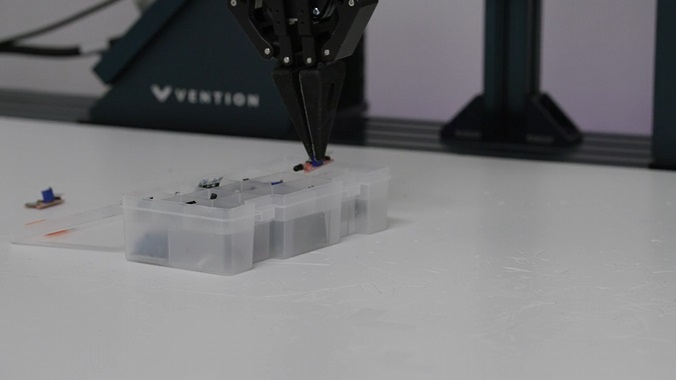}\hfill
\vlaphoto{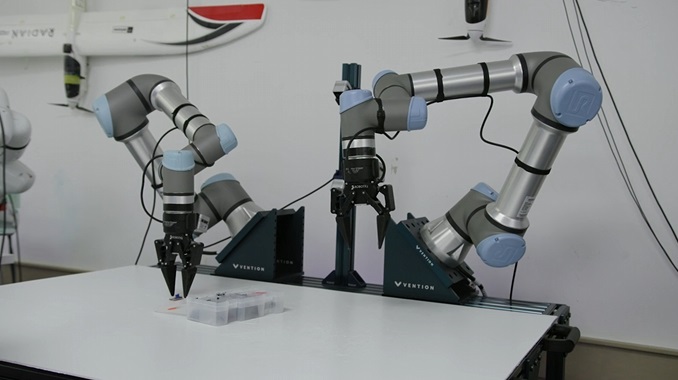}\hfill
\vlaphoto{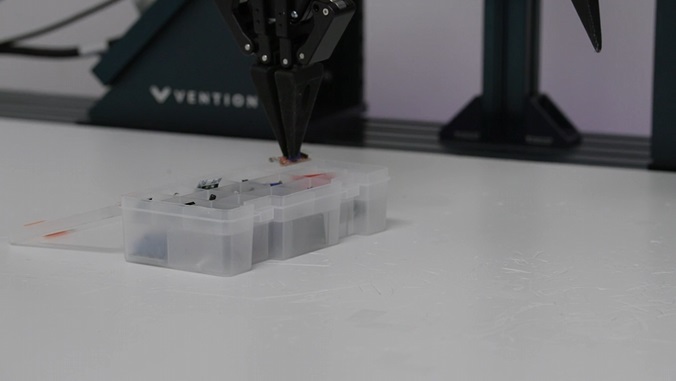}%
\vlarowlabel{Small-electronics cleanup}%

%% file: figures/fig-ur-ai-trainer.tex
%
%
%
%

\newcommand{\urRhoToolbox}{46.7}   \newcommand{\urRhoMicro}{60.0}
\newcommand{\urPiToolbox}{20.0}    \newcommand{\urPiMicro}{40.0}
\newcommand{\urGrToolbox}{16.7}    \newcommand{\urGrMicro}{13.3}

\newcommand{\urTpRhoToolbox}{57.5}  \newcommand{\urTpRhoMicro}{82.5}
\newcommand{\urTpPiToolbox}{27.5}   \newcommand{\urTpPiMicro}{80.8}
\newcommand{\urTpGrToolbox}{37.5}   \newcommand{\urTpGrMicro}{45.8}

\pgfmathsetmacro{\urRhoAll}{(\urRhoToolbox+\urRhoMicro)/2}
\pgfmathsetmacro{\urPiAll} {(\urPiToolbox +\urPiMicro )/2}
\pgfmathsetmacro{\urGrAll} {(\urGrToolbox +\urGrMicro )/2}
\pgfmathsetmacro{\urTpRhoAll}{(\urTpRhoToolbox+\urTpRhoMicro)/2}
\pgfmathsetmacro{\urTpPiAll} {(\urTpPiToolbox +\urTpPiMicro )/2}
\pgfmathsetmacro{\urTpGrAll} {(\urTpGrToolbox +\urTpGrMicro )/2}

\renewcommand{\vlaLegendCols}{3}

\pgfplotsset{
  ur ai trainer axis/.style={
    vla bar axis,
    width=0.60\vlaFigWidth,
    symbolic x coords={toolbox,micro,overall},
    xtick={toolbox,micro,overall},
    enlarge x limits=0.22,
    xticklabels={Toolbox, Micro-electronics\\cleanup, \textbf{Overall}},
  },
}

\vlalegendblock{urailegend}

\begin{minipage}{\linewidth}
\centering
\begin{tikzpicture}
\begin{axis}[
  ur ai trainer axis,
  ylabel={Success rate (\%)},
  vla legend,
  legend to name=urailegend,
]
\addplot[vla series=vlaRho] coordinates {
  (toolbox,\urRhoToolbox) (micro,\urRhoMicro) (overall,\urRhoAll)};
\addlegendentry{\RHOUAT}
\addplot[vla series=vlaPi] coordinates {
  (toolbox,\urPiToolbox) (micro,\urPiMicro) (overall,\urPiAll)};
\addlegendentry{\PI}
\addplot[vla series=vlaGroot] coordinates {
  (toolbox,\urGrToolbox) (micro,\urGrMicro) (overall,\urGrAll)};
\addlegendentry{\GR}
\vladividermid{micro}{overall}
\end{axis}
\end{tikzpicture}
\end{minipage}

\vspace{\vlaSubfigSkip}

\begin{minipage}{\linewidth}
\centering
\begin{tikzpicture}
\begin{axis}[
  ur ai trainer axis,
  ylabel={Mean task progress (\%)},
]
\addplot[vla series=vlaRho] coordinates {
  (toolbox,\urTpRhoToolbox) (micro,\urTpRhoMicro) (overall,\urTpRhoAll)};
\addplot[vla series=vlaPi] coordinates {
  (toolbox,\urTpPiToolbox) (micro,\urTpPiMicro) (overall,\urTpPiAll)};
\addplot[vla series=vlaGroot] coordinates {
  (toolbox,\urTpGrToolbox) (micro,\urTpGrMicro) (overall,\urTpGrAll)};
\vladividermid{micro}{overall}
\end{axis}
\end{tikzpicture}
\end{minipage}

%% file: figures/fig-fr3-duo-tasks.tex
%
%

\renewcommand{\vlaPhotoRowSkip}{0pt}%
\vlasetphotowidth{4}%
\vlaphoto{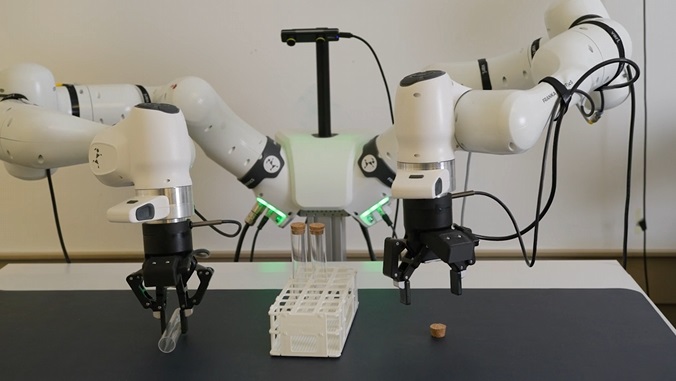}\hfill
\vlaphoto{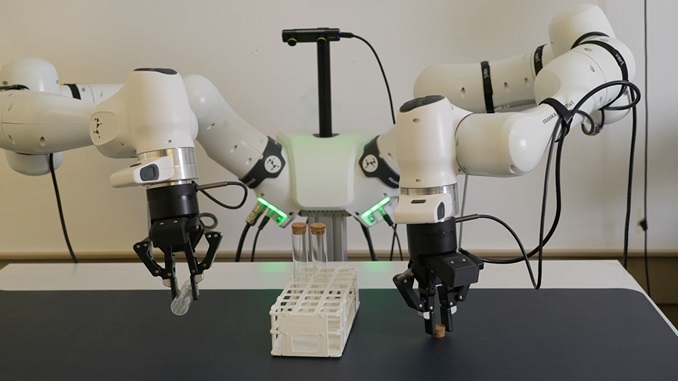}\hfill
\vlaphoto{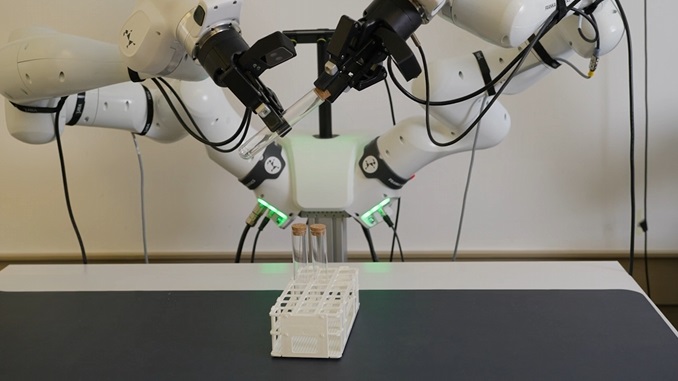}\hfill
\vlaphoto{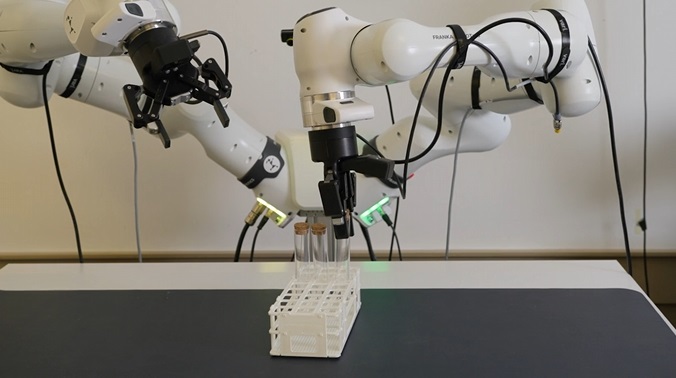}%
\vlarowarrows{Test-tube racking}{Test-tube unracking}%
\vspace{\vlaPhotoRowSkip}%
\vlaphoto{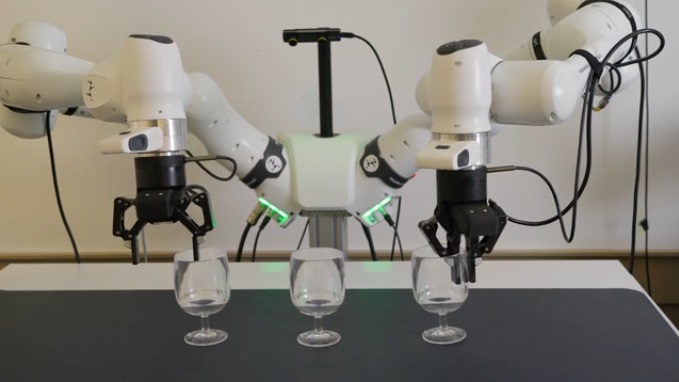}\hfill
\vlaphoto{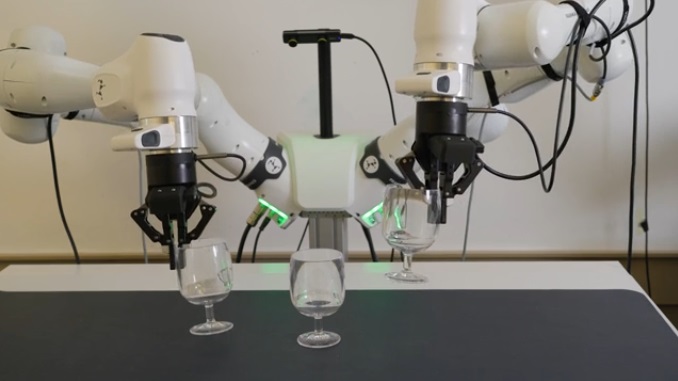}\hfill
\vlaphoto{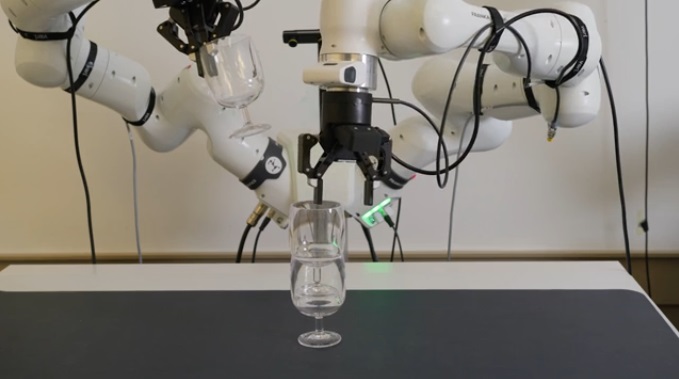}\hfill
\vlaphoto{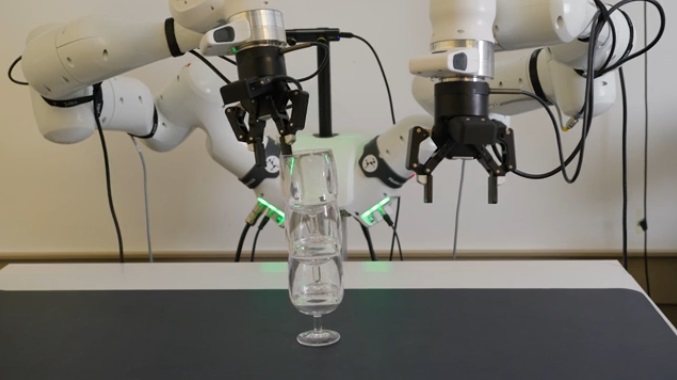}%
\vlarowarrows{Tumbler stacking}{Tumbler unstacking}%
\vspace{\vlaPhotoRowSkip}%
\hfill
\vlaphoto{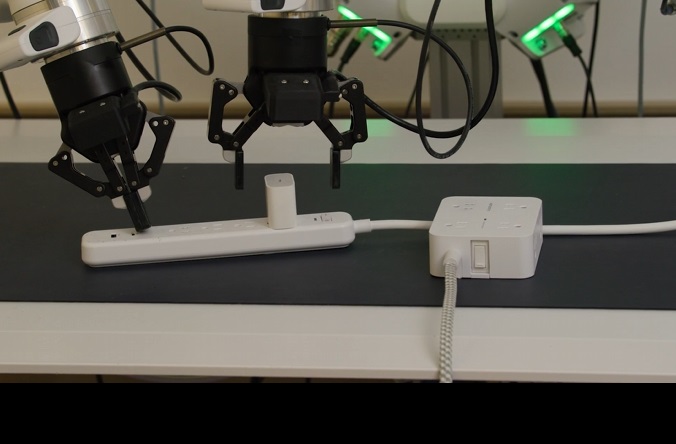}\hspace{\vlaPhotoGap}%
\vlaphoto{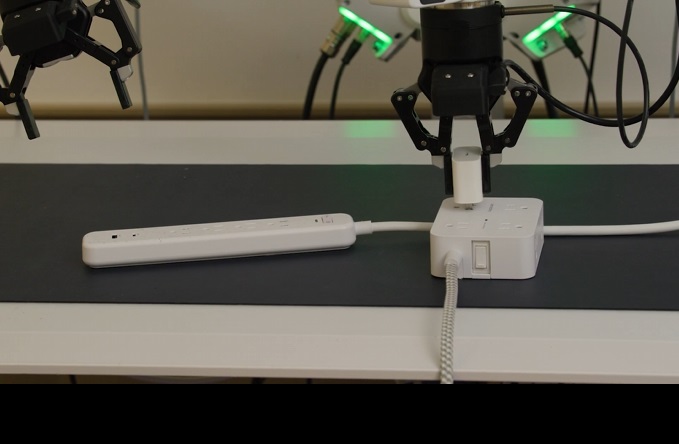}\hfill\null
\vlarowlabel{Plug insertion}%

%% file: figures/fig-fr3-duo.tex
%
%
%
%

\newcommand{\srRhoPlug}{80.0}       \newcommand{\srPiPlug}{40.0}       \newcommand{\srGrPlug}{60.0}
\newcommand{\srRhoStack}{70.0}      \newcommand{\srPiStack}{26.7}      \newcommand{\srGrStack}{36.7}
\newcommand{\srRhoUnstack}{93.3}    \newcommand{\srPiUnstack}{53.3}    \newcommand{\srGrUnstack}{56.7}
\newcommand{\srRhoRack}{63.3}       \newcommand{\srPiRack}{33.3}       \newcommand{\srGrRack}{23.3}
\newcommand{\srRhoUnrack}{93.3}     \newcommand{\srPiUnrack}{80.0}     \newcommand{\srGrUnrack}{73.3}

\newcommand{\tpRhoPlug}{90.0}       \newcommand{\tpPiPlug}{70.0}       \newcommand{\tpGrPlug}{80.0}
\newcommand{\tpRhoStack}{85.8}      \newcommand{\tpPiStack}{45.0}      \newcommand{\tpGrStack}{70.0}
\newcommand{\tpRhoUnstack}{95.0}    \newcommand{\tpPiUnstack}{60.0}    \newcommand{\tpGrUnstack}{68.3}
\newcommand{\tpRhoRack}{75.0}       \newcommand{\tpPiRack}{63.3}       \newcommand{\tpGrRack}{48.9}
\newcommand{\tpRhoUnrack}{97.8}     \newcommand{\tpPiUnrack}{84.4}     \newcommand{\tpGrUnrack}{89.4}

\pgfmathsetmacro{\srRhoAll}{(\srRhoPlug+\srRhoStack+\srRhoUnstack+\srRhoRack+\srRhoUnrack)/5}
\pgfmathsetmacro{\srPiAll} {(\srPiPlug +\srPiStack +\srPiUnstack +\srPiRack +\srPiUnrack )/5}
\pgfmathsetmacro{\srGrAll} {(\srGrPlug +\srGrStack +\srGrUnstack +\srGrRack +\srGrUnrack )/5}
\pgfmathsetmacro{\tpRhoAll}{(\tpRhoPlug+\tpRhoStack+\tpRhoUnstack+\tpRhoRack+\tpRhoUnrack)/5}
\pgfmathsetmacro{\tpPiAll} {(\tpPiPlug +\tpPiStack +\tpPiUnstack +\tpPiRack +\tpPiUnrack )/5}
\pgfmathsetmacro{\tpGrAll} {(\tpGrPlug +\tpGrStack +\tpGrUnstack +\tpGrRack +\tpGrUnrack )/5}

\renewcommand{\vlaLegendCols}{3}

\pgfplotsset{
  fr3 duo axis/.style={
    vla bar axis,
    symbolic x coords={plug,stack,unstack,rack,unrack,gap,overall},
    xmin=plug, xmax=overall,
    xtick={plug,stack,unstack,rack,unrack,overall},
    xticklabels={Plug\\insertion, Tumbler\\stacking, Tumbler\\unstacking,
                 Racking a\\test tube, Unracking a\\test tube, \textbf{Overall}},
  },
}

\vlalegendblock{fr3duolegend}

\begin{minipage}{\linewidth}
\centering
\begin{tikzpicture}
\begin{axis}[
  fr3 duo axis,
  ylabel={Success rate (\%)},
  vla legend,
  legend to name=fr3duolegend,
]
\addplot[vla series=vlaRho] coordinates {
  (plug,\srRhoPlug) (stack,\srRhoStack) (unstack,\srRhoUnstack)
  (rack,\srRhoRack) (unrack,\srRhoUnrack) (overall,\srRhoAll)};
\addlegendentry{\RHOFR}
\addplot[vla series=vlaPi] coordinates {
  (plug,\srPiPlug) (stack,\srPiStack) (unstack,\srPiUnstack)
  (rack,\srPiRack) (unrack,\srPiUnrack) (overall,\srPiAll)};
\addlegendentry{\PI}
\addplot[vla series=vlaGroot] coordinates {
  (plug,\srGrPlug) (stack,\srGrStack) (unstack,\srGrUnstack)
  (rack,\srGrRack) (unrack,\srGrUnrack) (overall,\srGrAll)};
\addlegendentry{\GR}
\vladivider{gap}
\end{axis}
\end{tikzpicture}
\end{minipage}

\vspace{\vlaSubfigSkip}

\begin{minipage}{\linewidth}
\centering
\begin{tikzpicture}
\begin{axis}[
  fr3 duo axis,
  ylabel={Mean task progress (\%)},
]
\addplot[vla series=vlaRho] coordinates {
  (plug,\tpRhoPlug) (stack,\tpRhoStack) (unstack,\tpRhoUnstack)
  (rack,\tpRhoRack) (unrack,\tpRhoUnrack) (overall,\tpRhoAll)};
\addplot[vla series=vlaPi] coordinates {
  (plug,\tpPiPlug) (stack,\tpPiStack) (unstack,\tpPiUnstack)
  (rack,\tpPiRack) (unrack,\tpPiUnrack) (overall,\tpPiAll)};
\addplot[vla series=vlaGroot] coordinates {
  (plug,\tpGrPlug) (stack,\tpGrStack) (unstack,\tpGrUnstack)
  (rack,\tpGrRack) (unrack,\tpGrUnrack) (overall,\tpGrAll)};
\vladivider{gap}
\end{axis}
\end{tikzpicture}
\end{minipage}

%% file: rel_work.tex
\section{Related work}

\subsection{Architectures of VLA models}

VLA models~\citep{zitkovich2023rt2,octo2024,kim2024openvla,shukor2025smolvla,liu2024rdt,black2024pi0,pi05,gr,wu2026pragmaticvla,im2026twinvla,fang2026molmoact2} harness pretrained vision-language representations produced by VLM models for robot control, providing a route for transferring semantic and perceptual knowledge into manipulation policies. Early systems such as RT-2 established the VLA formulation~\citep{zitkovich2023rt2} and generated actions autoregressively in a discretized space, while subsequent open models expanded the range of architectures, action representations, datasets, and deployment settings. \RHO\ belongs to this family. It combines a 4.7B Phi-family backbone distilled from Phi-4-reasoning-vision-15B~\cite{aneja2026phi4vision}, a continuous 0.5B flow-matching action expert, and a training pipeline that separates multi-embodiment pretraining, embodiment-targeted midtraining, and task adaptation.

Architectures of VLAs with a continuous action expert (AE) differ mainly in \emph{which} VLM representations condition the AE and \emph{how} they are injected. OpenVLA-OFT~\cite{kim2025finetuningvisionlanguageactionmodelsoptimizing} uses VLM's final-layer features for AE conditioning. GR00T~N1~\cite{gr} instead cross-attends to tokens from an intermediate LLM layer and discards the layers above it. SmolVLA~\cite{shukor2025smolvla} similarly keeps only the first half of the VLM but for the remaining VLM layers alternates cross- and self-attention in the AE. In $\pi_0$ and \PI~\cite{black2024pi0,pi05}, the AE is a transformer depth-matched to the VLM, and AE's whose tokens attend to the VLM's keys and values at every layer
through block-causal shared attention. For some of these architectures, several works have studied mechanisms such as knowledge insulation (KI) for limiting interference between action learning and the VLM or for introducing action-aware intermediate representations~\citep{driess2025knowledgeinsulatingvisionlanguageactionmodels,goyal2025vla0,lap}. Unfortunately, only few public studies have attempted to ablate these architectural decisions from the data recipe used for training the respective VLAs and from the amount of data used for pretraining. In one study~\cite{driess2025knowledgeinsulatingvisionlanguageactionmodels} KI is shown to be beneficial, but this study was conducted on a \PI-size dataset, likely over 10,000 hours of robot data. \citet{fang2026molmoact2}'s study, done on \MO's smaller-scale data, shows performance differences between layer-wise cross-attention-based architectures equipped with KI and single-layer-conditioning without KI to be small. Since \RHO's data scale is closer to \MO's and our experiments, too, showed little overall benefit of layer-wise cross-attention, we opted not to incorporate it into \RHO's architecture and instead optimize the shape of AE as described in Sections \ref{sec:action-expert} and \ref{sec:pretraining-ablations}.

\subsection{Other architectural paradigms in physical AI}

A growing alternative to VLM-initialized VLAs builds robot policies on pretrained video generation models instead of vision-language models. This paradigm predates capable DiT-based video generation models such as Wan~\cite{wan2025wanopenadvancedlargescale} and Cosmos~\cite{nvidia2025cosmosworldfoundationmodel}. PLEX~\cite{thomas2023plex} and UniPi~\cite{du2023unipi} generated a text-conditioned video plan and recovered actions with a separately trained inverse dynamics model (IDM). GR-1 and GR-2~\cite{wu2023unleashing,cheang2024gr2} pretrained policies on large-scale video prediction before jointly finetuning on future frames and actions. The advent of Cosmos, Wan, and other strong pretrained video foundation models gave rise to two overlapping physical AI model classes. \emph{Video-action models} (VAMs)~\cite{pai2025mimicvideo}, e.g., VPP~\cite{hu2025videopredictionpolicygeneralist}, mimic-video~\cite{pai2025mimicvideo}, and GE-Act~\cite{ge2025}, treat the video backbone mainly as a source
of predictive representations and attach a lightweight diffusion or flow-matching action decoder, which acts as an IDM, to intermediate video-model features, without necessarily rendering future frames. \emph{World action models} (WAMs)~\cite{ye2026worldactionmodelszeroshot}, including DreamZero~\cite{ye2026worldactionmodelszeroshot}, Cosmos Policy~\cite{kim2026cosmospolicy}, and UWM~\cite{zhu2025unifiedworldmodelscoupling}, instead emphasize joint generation: a single model predicts both future visual states and actions. LingBot-VA~\cite{li2026lingbotva} and Fast-WAM~\cite{yuan2026fastwamworldactionmodels} straddle the boundary between VAMs and WAMs, incorporating elements of both approaches.

Conceptually, VLAs differ from WAMs and VAMs mainly in the prior they inherit. VLAs get strong semantic and language grounding from static image-text
pretraining but must learn physical dynamics from robot data alone. Video-based policies instead inherit spatiotemporal dynamics priors and receive dense per-frame supervision, but at the cost of substantially higher training cost and inference latency from long
video-token sequences. Given only limited controlled comparisons between VAMs/WAMs and state-of-the-art VLAs~\cite{zhang2026worldactionmodelsgeneralize} and opportunities to distill a physically-aware backbone from Phi, we chose to base \RHO on a VLM. However, we envision future \RHO-series models to video-generation backbones as this technology matures.


\subsection{Physical grounding and VLA pretraining}

The knowledge encoded by a general VLM is useful for robotics but is not necessarily optimized for spatial relationships, affordances, action consequences, or the observation distributions encountered by robots. Prior work improves this starting point through robotics-relevant vision-language supervision~\citep{gemini2025robotics,fang2026molmoact2} and mixes robot demonstrations with web-scale or embodied vision-language data during VLA training~\citep{zitkovich2023rt2,pi05,gr,fang2026molmoact2,lin2026cotraining}. \RHO\ similarly grounds its Phi-family backbone before action learning and retains physical-grounding data during multi-embodiment pretraining and embodiment midtraining. However, unlike most existing works such as \citet{pi05}, \RHO\ performs the grounding by introducing the grounding data into the VLM's training mixture (RT2), rather than simply post-training the VLM backbone on a combination of robot demonstrations and VQA data.

\subsection{Data recipes, normalization and representations} 
The centerpiece of \RHO's data recipe is co-training, a concept that has been the mainstay of VLA development since at least RT-2~\citep{zitkovich2023rt2}.

One set of key questions in co-training focuses on the kinds of data that should be included at each training stage. VLAs vary greatly in mixtures they use. Although some models are trained using robot demonstrations only~\cite{kim2024openvla}, most VLAs rely on a combination of robot demonstrations with Web VQA~\citep{zitkovich2023rt2} and even videos~\cite{gr,pi07} during pretraining, followed by robot-demonstrations data during later stages~\cite{black2024pi0,kim2025finetuningvisionlanguageactionmodelsoptimizing}. \RHO's main difference from most existing approaches is the use of a multi-task multi-embodiment demonstration mixture that includes UMI-style data~\cite{umi} during pretraining, multi-task \emph{single-}embodiment mixture during midtraining, and few-task single-embodiment during finetuning. An ablation study that converges on a similar recipe, with grounding VQA data mixed in during the first two stages, is~\citet{lin2026cotraining}. This and other studies, e.g., ~\citet{chen2025internvlam1spatiallyguidedvisionlanguageaction} show that co-training on VQA data after pretraining is crucial for regularizing the VLA to prevent degradation of language understanding.

Another important data recipe consideration concerns the \emph{quality} of co-training sub-datasets. Some works suggest that demonstrations of varying quality is acceptable and even desirable at pretraining in order to teach the model recovery behaviors~\cite{black2024pi0}, but data that standard for later training stages becomes significantly higher to make the learned behaviors fluent~\cite{pi05}. Yet other approaches observe that substandard data can be broadly helpful if it is annotated accordingly~\cite{pi07}. \RHO uses a mix of public datasets~\cite{oxe2024} containing potentially suboptimal demonstrations and curated proprietary data for its target embodiments \yb, \urait, and \fr3d, where erroneous trajectory segments are deliberately removed. During online task adaptation, it gets exposed to suboptimal demonstrations as well, but uses only their corrections for training.

Effectiveness of co-training on a multi-embodiment mixture of datasets also depends on the representations these datasets use for robots' proprioception and actions. The $\pi_0$~\cite{black2024pi0} and \PI~\cite{pi05} keeps each robot's own action space, mostly joint space, and zero-pad state and action vectors to the dimensionality of the largest robot in the co-training mixture. RDT-1B~\cite{liu2024rdt} maps each robot's actions into a shared 128-dimensional vector format, where every dimension has a fixed physical meaning, filling only the slots the robot has. GR00T N1~\cite{gr} uses a separate MLP per embodiment to project states and actions of different sizes into a shared embedding for the DiT. The aforementioned co-training study in ~\citet{lin2026cotraining} converts all data to EEF pose space. The EEF pose space, in turn, is sensitive to the choice of representation for EEF rotations. \RHO is pretrained and midtrained in the EEF pose space; we use the 6D representation from ~\citet{zhou2019continuity} and ~\cite{trilbm2025behavior} due its continuity and chunk-start-relative actions, which mitigates memorization issues of absolute action spaces and avoids fast error accumulation typical of delta actions.

\subsection{Embodiment and task adaptation}

VLAs are commonly adapted to a target robot and task simultaneously through supervised fine-tuning~\citep{black2024pi0,pi05,gr,fang2026molmoact2}. \RHO\ explicitly separates these roles: embodiment midtraining uses a broad multi-task but single-robot dataset to familiarize the model with the robot platform's possible behaviors and deployment environments, while downstream adaptation uses narrower demonstration sets to learn the target tasks. The controlled RoboTwin experiment (\Cref{sec:midtraining-eval}) and evaluations of \RHO's variants midtrained for \yb, \urait, and \fr3d (\Cref{sec:physical-results}) analyze the effectiveness of this separation.

Training stages conceptually related to \RHO's embodiment midtraining have been used in building several other VLAs~\citep{pi05,lin2026cotraining,fang2026molmoact2} but with two notable differences.

First, while \RHO uses midtraining as a means of \emph{adapting} the pretrained model to a new embodiment, i.e., \emph{generally} shifting the distribution of the model's knowledge towards a particular robot and its use cases, \PI~\citep{pi05} and \MO~\citep{fang2026molmoact2} use it as a tool for \emph{specializing} the model by narrowing its training data distribution. Namely, these works consider only the case where the target robot's data is already present in the pretraining mixture, and the stage that follows pretraining focuses on the target robot data almost exclusively. Our experiment in \Cref{sec:midtraining-eval} shows that embodiment training helps significantly after pretraining even if the model isn't exposed to the target embodiment's data during pretraining.

Second, \MO~\citep{fang2026molmoact2} and \PI~\citep{pi05} view embodiment-specific training as the final step in training pipeline and accordingly refer to it as \emph{post-training} or \emph{finetuning}, with the aim of achieving zero-shot generalization on the target robot. 
\RHO's training recipe instead focuses on using embodiment midtraining as a step preparing the model for data-efficient task-specific adaptation downstream. Our experiments measure the benefits of midtraining for this purpose.

~\citet{lin2026cotraining} come closest to \RHO's use of embodiment training as a step followed by further model adaptation. Compared to their study, \RHO's evaluation in \Cref{sec:exps} shows that benefits of robot-targeted midtraining go beyond any single amount of finetuning data and hold across a wide range of data regimes (\Cref{fig:robotwin-midtraining}).

\subsection{Online robot learning}

Offline imitation policies are trained on states visited by their demonstrators, but deployment exposes them to states induced by their own errors. Online robot learning closes this distribution gap by collecting experience under the deployed policy and using it to improve the policy. Interactive imitation and reinforcement-learning systems have shown that human intervention can both keep collection safe and provide targeted learning signal at precisely these failure states~\citep{liu2023sirius,luo2024hilserl,luo2024rlif}.

Recent work extends this idea from task-specific controllers to generalist vision-language-action models. $\pi^{*}_{0.6}$ introduces RECAP, which combines demonstrations, autonomous rollouts, and teleoperated interventions, learns a value function from this heterogeneous experience, and improves the VLA through advantage-conditioned training~\citep{pi06}. This demonstrates that a large VLA can improve through on-robot practice rather than treating supervised post-training as its final stage. However, updating a full VLA online is computationally demanding and may alter capabilities inherited from pretraining.

Flow- and diffusion-based policies provide another adaptation surface: the latent initial condition from which the action-generation process begins. Latent-space reinforcement learning learns a policy over this variable from task rewards, while \FD\ recovers latent targets by inverting corrected action chunks and trains on them with supervised regression~\citep{wagenmaker2025steering,murray2026flowdagger}. Both approaches can leave the pretrained action generator fixed, concentrating online updates in a substantially smaller module, although the adapted behavior remains constrained by the frozen generator. \RHO\ incorporates this mechanism directly: its flow model can contain an internal latent policy that replaces random Gaussian sampling with an observation-conditioned initial condition and is saved as part of the same model checkpoint. Our experiments study the corrective-supervision variant, using scripted interventions in simulation and human interventions on the physical robot.

%% file: limitations.tex
\section{Limitations and future work}
\label{sec:limitations}

\RHO\ is evaluated on a finite set of simulation benchmarks, robot embodiments, and manipulation tasks. These experiments do not establish universal generalization across hardware, environments, or task distributions. In particular, embodiment midtraining is studied on three physical bimanual platforms, and its benefits may depend on the diversity and quality of the embodiment-specific data. Our strongest-model claim is therefore restricted to the baselines, training budgets, and evaluation protocols reported in \Cref{sec:exps}.

There are several important interconnected questions left open by our study. The first of them concerns \RHO's language steerability. We have not systematically evaluated how well \RHO can follow diverse natural language instructions or how sensitive it is to variations in phrasing. Despite its importance, this question has received limited systematic study in prior work on generalist robotic manipulation policies. Nonetheless, zero-shot generalization in language, vision, and action space remains the North Star of physical AI, and the research community is seeing early signs of its advent~\cite{neurips26wrl}. We believe the importance of linguistic robustness of models like \RHO and its descendants will become paramount in the near future.

The second question focuses on the effectiveness of \RHO as part of agentic AI systems driven by frontier generative AI models and harnesses. Recent results indicate that performance of VLAs can be significantly enhanced when combined with GPT-6-Astra~\citep{su2026astra} acting as a governor, enabling more sophisticated reasoning, planning, and task execution capabilities in robotic manipulation scenarios. Exploring how well \RHO performs in these workflows and, especially, how to make it stronger in this role by modifying \RHO's horizontal midtraining and vertical adaptation recipe would help inform \RHO's practical deployment strategy. Answering the aforementioned question about the extent of \RHO's steerability in the language domain, the interaction interface between \RHO and GPT-6-Astra-class models, is a crucial step in this direction.

The third direction is related to learning from feedback modalities that have remained un- or under-utilized by physical AI so far. One example of them is tactile sensing. Until recently, the lack of standardization in tactile sensing technology, its technical shortcomings, and its rarity in robot deployments have prevented the community from accumulating large amounts of data for it and, as a consequence, have been slowing its integration into physical AI models. With the increasing use of humanoid hands instead of grippers, the availability of datasets such as FTP-1~\cite{ftp1}, and the advent of simulated data generation methods including OmniReset~\cite{yin2026emergent} and TacSL~\cite{akinola2025tacsl}, adapting \RHO to accept and process tactile feedback hold great potential for increasing its manipulation robustness and dexterity.

%% file: release.tex
\section{Public release and responsible deployment}
\label{sec:release}

To support reproducibility and downstream use, we release the pretrained \RHO\ model together with the embodiment-midtrained \RHOFR, \RHOUAT, and \RHOYAM\ checkpoints. Alongside the model weights, we release code to fine-tune the checkpoints on downstream tasks, adapt them online using the latent-space methods studied in \Cref{sec:online-adaptation}, and evaluate the released checkpoints. The release is intended to give practitioners a choice between adapting the general multi-embodiment checkpoint and starting from a checkpoint that has already learned the sensing and control conventions of a target platform. Last but not least, we release the dataset for BusyBox tasks collected on \yb at \url{https://huggingface.co/datasets/microsoft/BusyBox}

The released models are available on Hugging Face at \url{https://huggingface.co/collections/microsoft/\Rho}, and the code for downstream finetuning and online adaptation is available on GitHub at \url{https://github.com/microsoft/rhobotics}. For an overview of the \RHO project, refer to \url{https://microsoft.github.io/rhobotics}.

When experimenting with \RHO, we urge researchers and practitioners to follow responsible deployment practices. Offline task adaptation remains dependent on representative demonstrations, while online adaptation introduces additional operational and safety considerations. Interactive imitation learning requires timely and consistent corrections, and reinforcement learning depends on reward design and safe exploration. Online updates should be performed under human supervision with platform-specific safeguards, workspace constraints, and evaluation gates before deployment.

%% file: conclusion.tex
\section{Conclusion}

We presented \RHO, an open-weights 5B-parameter vision-language-action model family for dual-arm robot manipulation built from physically grounded Phi-family backbone and a flow-matching action expert. In addition to the base \RHO model, we release embodiment-midtrained checkpoints for I$^2$RT \yb, Franka \fr3d, and Universal Robots AI Trainer -- three representative dual-arm robot platforms for bimanual robotic manipulation.

The report studies the architectural choices that connect these components, the recipe used for multi-embodiment pretraining, and embodiment midtraining as an intermediate stage between general pretraining and task specialization. Across simulation and physical-robot experiments, we evaluate both the resulting base model and robot-specific checkpoints against strong open-weights baselines, establishing \RHO as a strong contender to state-of-the-art VLAs across a broad range of tasks and embodiments. In particular, our experiments demonstrate the power of embodiment midtraining, which reduces the amount of offline finetuning data that needs to be collected on the target robot by half or, viewed differently, routinely improves task success rates after finetuning on a given small dataset by 20\% or more.  We further demonstrate that task-specialized \RHO policies can continue adapting online in latent action space through interactive learning. Together with the released pretrained and embodiment-midtrained checkpoints, these results provide a practical foundation for studying and deploying adaptable robot policies.

%% file: acks.tex
\section{Acknowledgments}
We thank Microsoft Research Hardware Lab members Christopher O'Dowd, Teresa LaScala, Ibrahim Saeed, Mike Sheppard, and Lex Story for manufacturing robot parts and other experiment equipment; Vivan Amin and Ade Famoti for help in establishing collaborations with industrial partners; Ashley Llorens for sponsorship and guidance; Sriram Siva, K Wang, Xu Liu, Tim Chung, and Dan Rosenstein for sharing their experiences on dataset and model evaluations; and Abhishek Gupta and Patrick Yin from the University of Washington, Seattle for stimulating research conversations and insights on simulated data generation. We also thank Jyoti Aneja, Tyler LaBonte, and John Langford for their hard work on the Phi-4-reasoning-vision model that was used as the basis for this work.

We express our gratitude to Microsoft Research Early Access Program participant Teradyne Robotics for their technical support of our experiments on the UR AI Trainer robot, and thank Scale AI for collecting UR AI Trainer teleoperation data and iterating with us on the data collection protocol.

%% file: appendix.tex
\appendix
\crefalias{section}{appendix}
\crefalias{subsection}{appendix}

\section*{Appendix}
\section{Model architecture details}
\label{app:model}

This section records the complete model configuration used for \RHO's full
multi-embodiment pretraining run.  \RHO{} couples the \phiphy{} backbone to a
12-block continuous action expert and conditions the expert on the backbone's
decoder-block-14 representation.  Prediction and execution horizons are
configurable properties of the training and deployment recipes; the widths,
attention layout, and conditioning interface remain fixed.

\begin{table}[H]
  \centering
  \small
  \setlength{\tabcolsep}{5pt}
  \begin{tabularx}{\textwidth}{@{}lX@{}}
    \toprule
    Component & Configuration \\
    \midrule
    Vision-language backbone & \phiphy{}; 4.68B parameters \\
    Language decoder & 32 causal-transformer blocks; width 3,072; MLP width
      8,192; 32 query and 32 key/value heads \\
    Text interface & Vocabulary size 100,352; configured context length 16,384 \\
    Vision encoder & SigLIP~2 SO400M NaFlex; 428M parameters; $16\!\times\!16$
      patches; penultimate-layer visual features \\
    Visual-token budget & 256--3,600 tokens per image; robot images are resized
      with padding to $256\!\times\!256$, yielding 256 visual patches per view \\
    Cross-modal projector & Two-layer MLP,
      $1{,}152\!\rightarrow\!3{,}072\!\rightarrow\!3{,}072$, with GELU \\
    Robot observation & One current observation, a language instruction,
      proprioceptive state, and up to three RGB camera views \\
    Backbone output used by control & Joint image--language sequence after
      decoder block 14, before the remaining decoder blocks \\
    \bottomrule
  \end{tabularx}
  \caption{\phiphy{} backbone configuration used by \RHO.}
  \label{tab:app-backbone-architecture}
\end{table}

\begin{table}[H]
  \centering
  \small
  \setlength{\tabcolsep}{5pt}
  \begin{tabularx}{\textwidth}{@{}lX@{}}
    \toprule
    Component & Configuration \\
    \midrule
    Action stream & One projected state token followed by one noisy-action
      token for every position in the action chunk \\
    Transformer & 12 blocks; width 2,048; feed-forward width 4,096; GELU
      activation; dropout 0.1 during pretraining \\
    Block structure & Action-stream self-attention, cross-attention to the fixed
      VLM context, then a feed-forward update, each with a gated residual \\
    Attention & 16 attention heads, 4 key/value heads, head dimension 128;
      grouped-query attention is used in both attention modules \\
    Normalization & DiT-style adaLN-Zero; shared adaLN-single base maps with
      zero-initialized block- and site-specific offsets \\
    Position encoding & Fixed sinusoidal encoding; configured maximum sequence
      length 6,144 \\
    State/action width & State and action vectors are padded to 32 dimensions;
      masks exclude padded dimensions and invalid chunk positions from the loss \\
    Flow input and output & Linear $32\!\rightarrow\!2{,}048$ noisy-action
      projection and $2{,}048\!\rightarrow\!32$ velocity head \\
    Prediction horizon & 50 positions during multi-embodiment pretraining; configurable
      during embodiment adaptation without changing transformer parameters \\
    \bottomrule
  \end{tabularx}
  \caption{Configuration of the \RHO{} flow-matching action expert.}
  \label{tab:app-action-expert-architecture}
\end{table}

\begin{table}[H]
  \centering
  \small
  \setlength{\tabcolsep}{4.5pt}
  \begin{tabularx}{\textwidth}{@{}p{0.18\textwidth}p{0.25\textwidth}p{0.18\textwidth}X@{}}
    \toprule
    Signal & Source representation & Mapping & Use in action expert \\
    \midrule
    Visual-language context & Block-14 \phiphy{} hidden sequence, width 3,072
      & Learned linear $3{,}072\!\rightarrow\!2{,}048$
      & Shared as the key/value sequence for cross-attention in every expert block \\
    Robot state & One state vector padded to width 32
      & Learned linear $32\!\rightarrow\!2{,}048$
      & Prepended to the action-query sequence as one token \\
    Noisy action & $H\!\times\!32$ flow state
      & Linear action projection followed by a
        $4{,}096\!\rightarrow\!2{,}048\!\rightarrow\!2{,}048$ action--time MLP
      & Supplies the remaining $H$ query tokens \\
    Flow time & Scalar $t$ with a 2,048-dimensional sinusoidal embedding
      & $2{,}048\!\rightarrow\!2{,}048\!\rightarrow\!2{,}048$ SiLU MLP
      & Conditions the shared adaLN maps in every block \\
    Padding masks & Valid image/text tokens, state/action dimensions, and chunk positions
      & No learned mapping
      & Masks attention or loss terms associated with padding \\
    \bottomrule
  \end{tabularx}
  \caption{The interface between \phiphy{} and the continuous action expert.
  The backbone context is computed once per observation and reused by all flow
  integration steps.}
  \label{tab:app-conditioning-interface}
\end{table}

\begin{table}[H]
  \centering
  \small
  \setlength{\tabcolsep}{5pt}
  \begin{tabular}{@{}lrl@{}}
    \toprule
    Component & Parameters & Pretraining status \\
    \midrule
    \phiphy{} vision--language backbone & 4.68B & Trainable except token embeddings \\
    \quad SigLIP~2 vision encoder & 428M & Trainable; included above \\
    \quad Token-embedding table & 308M & Frozen; included above \\
    Action expert & 542M & Trainable from random initialization \\
    \midrule
    Complete \RHO{} model & 5.22B & Approximately 4.91B trainable \\
    \bottomrule
  \end{tabular}
  \caption{Parameter accounting for \RHO. Counts are rounded to the precision
  used throughout the paper.}
  \label{tab:app-parameter-counts}
\end{table}

\begin{table}[H]
  \centering
  \small
  \setlength{\tabcolsep}{5pt}
  \begin{tabularx}{\textwidth}{@{}lX@{}}
    \toprule
    Inference stage & Operation \\
    \midrule
    Observation encoding & Encode all camera views and the instruction with
      \phiphy{}, extract decoder block 14, and project the valid context tokens
      to width 2,048. This backbone pass occurs once per policy query. \\
    Latent initialization & Draw an independent Gaussian tensor
      $x_1\in\mathbb{R}^{H\times32}$; an internal learned latent policy can
      replace this draw during online adaptation (\Cref{app:online}). \\
    Flow integration & Apply 10 uniform explicit-Euler steps from $t=1$ to
      $t=0$. Each step re-embeds the current action latent and queries the
      12-block action expert while reusing the cached VLM context. \\
    Action decoding & Project the final action tokens to 32 dimensions, remove
      padded dimensions, and invert the embodiment-specific normalization. No
      autoregressive text decoding is used to produce robot actions. \\
    Receding-horizon execution & Execute a configurable prefix and then replan.
      Multi-embodiment pretraining uses $H=50$ and a 25-action execution horizon;
      RoboEval uses 32/16 and LIBERO uses 16/8. \\
    Numerics & Bfloat16 model execution with FlashAttention~2; the Euler state
      update is accumulated in float32 before conversion back to bfloat16. \\
    \bottomrule
  \end{tabularx}
  \caption{\RHO{} inference path. Prediction and execution horizons
  are adapted to the control frequency and temporal extent of each embodiment.}
  \label{tab:app-inference}
\end{table}

\section{Training data and implementation details}
\label{app:training}

\subsection{Multi-embodiment pretraining}

\begin{table}[H]
  \centering
  \small
  \setlength{\tabcolsep}{5pt}
  \begin{tabular}{@{}lp{0.69\textwidth}@{}}
    \toprule
    Component & Configuration \\
    \midrule
    Initialization & \phiphy\ backbone; randomly initialized state/action projections and action expert \\
    Trainable parameters & Vision encoder, cross-modal projector, language decoder, and action modules; token embeddings frozen \\
    Robot inputs & One observation; up to three RGB views; padded $256\!\times\!256$ images; task instruction and proprioceptive state \\
    Action targets & Chunk-relative EEF translation and 6D-rotation deltas; absolute grippers; approximately 1\,s at the native rate; at most 50 steps; per-position 1st/99th-percentile normalization \\
    Training objectives & Flow matching on robot batches; autoregressive cross-entropy on VQA and pointing batches, weighted by $0.02$ \\
    Batch sampling & Robot/VL update probability $0.9/0.1$; global batch 3,072/384 \\
    Optimizer & AdamW; $\beta_1=0.9$, $\beta_2=0.95$; $\epsilon=10^{-6}$; weight decay $0.1$ \\
    Peak learning rates & $1\!\times\!10^{-6}$ (vision-language modules); $1\!\times\!10^{-4}$ (action modules) \\
    Schedule & 1,000-update linear warmup; stable phase; 45,000-update cosine decay to $0.1\times$ peak \\
    Regularization & Dropout $0.1$; global gradient-norm clipping at $1.0$ \\
    Numerics and compute & Bfloat16; 48 NVIDIA B200 GPUs \\
    \bottomrule
  \end{tabular}
  \caption{Multi-embodiment pretraining recipe for \RHO. Batch sizes count examples per modality-homogeneous optimization update.}
  \label{tab:pretraining-recipe}
\end{table}

\subsection{Embodiment midtraining}

For each target platform, embodiment midtraining continues from the
multi-embodiment checkpoint using a broad, platform-specific robot dataset while
retaining a small fraction of vision--language supervision.  \Cref{tab:fr3-midtraining-recipe}
gives the \fr3d recipe used to produce the checkpoint from which the FR3 task
policies are initialized.  Robot and vision--language batches are sampled in a
9:1 ratio.  This stage preserves the pretraining action-expert architecture and
optimizes the complete model, so the backbone, interface projections, and action
expert can jointly acquire the target platform's visual and control conventions.

\begin{table}[H]
  \centering
  \small
  \setlength{\tabcolsep}{5pt}
  \begin{tabular}{@{}lp{0.69\textwidth}@{}}
    \toprule
    Component & Configuration \\
    \midrule
    Initialization & First-epoch multi-embodiment \RHO{} checkpoint \\
    Data mixture & \fr3d multi-task robot data and retained vision--language
      data, sampled at a 9:1 robot/VL ratio \\
    Trainable parameters & Vision encoder, cross-modal projector, language
      decoder, and action modules; token embeddings frozen \\
    Action representation & One-second chunk of chunk-relative Cartesian
      end-effector translation and 6D-rotation deltas with absolute gripper
      commands, padded to 50 positions \\
    Batch and updates & 128 examples per GPU on 8 B200 GPUs (global batch
      1,024); 165,000 optimizer updates \\
    Optimizer & AdamW; $\beta_1=0.9$, $\beta_2=0.95$; $\epsilon=10^{-6}$;
      weight decay $0.1$ \\
    Learning rates & $1\!\times\!10^{-6}$ for vision--language modules and
      $1\!\times\!10^{-4}$ for the action modules \\
    Schedule & 1,000-update warmup, stable phase, then 15,000-update decay
      from $10^{-4}$ to $10^{-5}$ \\
    Regularization and numerics & Dropout $0.1$; global gradient-norm clipping
      at $1.0$; bfloat16 \\
    \bottomrule
  \end{tabular}
  \caption{\fr3d embodiment-midtraining recipe.}
  \label{tab:fr3-midtraining-recipe}
\end{table}

\subsection{Task finetuning}

Task finetuning starts from the embodiment-midtrained checkpoint and uses only
the demonstrations for the target task family.  \Cref{tab:fr3-finetuning-recipe}
summarizes the FR3 test-tube configuration, which is representative of the FR3
task-adaptation runs.  Each model is adapted independently on the corresponding
task demonstrations; the evaluation-specific data budgets are given in
\Cref{sec:physical-results}.

\begin{table}[H]
  \centering
  \small
  \setlength{\tabcolsep}{5pt}
  \begin{tabular}{@{}lp{0.69\textwidth}@{}}
    \toprule
    Component & Configuration \\
    \midrule
    Initialization & \fr3d embodiment-midtrained \RHO{} checkpoint \\
    Data & Target-task demonstrations only; three RGB views, language, and
      proprioception \\
    Trainable parameters & Vision encoder, cross-modal projector, language
      decoder, and action modules; token embeddings frozen \\
    Action representation & One-second chunk of chunk-relative Cartesian
      end-effector translation and 6D-rotation deltas with absolute gripper
      commands, padded to a 50-position model chunk \\
    Normalization & Per-task action-chunk mean and standard deviation \\
    Batch and updates & 16 examples per GPU on 4 B200 GPUs (global batch 64);
      50,000 optimizer updates \\
    Flow samples & One sample per training example \\
    Optimizer & AdamW; $\beta_1=0.9$, $\beta_2=0.95$; $\epsilon=10^{-6}$;
      weight decay $0.1$ \\
    Learning rate & $1\!\times\!10^{-4}$ for all unfrozen parameters \\
    Schedule & 1,000-update warmup, stable phase, then 15,000-update decay
      from $10^{-4}$ to $10^{-5}$ \\
    Regularization and numerics & Dropout $0.1$; global gradient-norm clipping
      at $1.0$; bfloat16 \\
    \bottomrule
  \end{tabular}
  \caption{Representative \fr3d task-finetuning recipe.}
  \label{tab:fr3-finetuning-recipe}
\end{table}

\section{Simulation evaluation protocols}
\label{app:sim-eval}

\Cref{tab:simulation-protocol-summary} summarizes the LIBERO, RoboEval, and RoboTwin evaluation
budgets.  Unless stated otherwise, success is binary and a rollout is
terminated as soon as the benchmark's task predicate is satisfied or its time limit
is reached.  We average tasks uniformly, so tasks with longer horizons or more
demonstrations do not receive greater weight.

\begin{table}[H]
  \centering
  \small
  \setlength{\tabcolsep}{4.5pt}
  \begin{tabular}{@{}lcccc@{}}
    \toprule
    Benchmark & Tasks & Adaptation & Trials / task / pass & Evaluation passes \\
    \midrule
    LIBERO & 40 (four suites) & 40k updates & 50 & 3 \\
    RoboEval & 8 & 40k updates & 100 & 3 \\
    RoboTwin & 5 (each under Easy \& Hard settings) & 40k updates & 50 & 3 (Hard)/4 (Easy) \\
    \bottomrule
  \end{tabular}
  \caption{LIBERO, RoboEval, and RoboTwin task-adaptation and evaluation budgets for the
  reported \RHO{} results. Each pass is a complete evaluation over the
  corresponding task set.}
  \label{tab:simulation-protocol-summary}
\end{table}

\paragraph{LIBERO.}
We use all 40 tasks in the four standard ten-task suites.  For completeness, the
task inventory is listed below; semicolon-separated entries denote distinct tasks.

\begin{table}[H]
  \centering
  \scriptsize
  \setlength{\tabcolsep}{5pt}
  \begin{tabularx}{\textwidth}{@{}lX@{}}
    \toprule
    Suite & Tasks \\
    \midrule
    Spatial &
    Move the black bowl to the plate from: between the plate and ramekin; next to
    the ramekin; the table center; atop the cookie box; the cabinet's top drawer;
    atop the ramekin; next to the cookie box; the stove; next to the plate; and
    atop the wooden cabinet. \\
    \addlinespace
    Object &
    Place in the basket: alphabet soup; cream cheese; salad dressing; BBQ sauce;
    ketchup; tomato sauce; butter; milk; chocolate pudding; and orange juice. \\
    \addlinespace
    Goal &
    Open the cabinet's middle drawer; put the bowl on the stove; put the wine bottle
    atop the cabinet; open the top drawer and put the bowl inside; put the bowl atop
    the cabinet; push the plate in front of the stove; put the cream cheese in the
    bowl; turn on the stove; put the bowl on the plate; put the wine bottle on the
    rack. \\
    \addlinespace
    Long &
    Put alphabet soup and tomato sauce in the basket; put cream cheese and butter in
    the basket; turn on the stove and put the moka pot on it; put the black bowl in
    the bottom drawer and close it; put the white mug on the left plate and the
    yellow-and-white mug on the right plate; put the book in the caddy's rear
    compartment; put the white mug on the plate and the chocolate pudding to its
    right; put alphabet soup and cream cheese in the basket; put both moka pots on
    the stove; put the yellow-and-white mug in the microwave and close it. \\
    \bottomrule
  \end{tabularx}
  \caption{LIBERO task inventory used in our simulation evaluation. Spatial and
  Object hold the goal fixed while varying the relevant spatial relation or object;
  Long contains the benchmark's ten multi-stage tasks.}
  \label{tab:libero-task-list}
\end{table}

We first adapt \RHO{} offline on the official demonstrations from all four suites for 40k
optimizer updates with global batch size 128 and eight flow samples per training
example. We then freeze the backbone and action expert and online-adapt only the
internal latent policy using rollouts from the benchmark's training initial states
and corrective supervision from a scripted LIBERO expert, following
\Cref{sec:online-adaptation}. At evaluation, each task is run for 50 trials using the benchmark-provided
initial states.  The horizon is 220 steps for Spatial, 280 for Object, 300 for Goal,
and 520 for Long.  A rollout succeeds when the official LIBERO task predicate is
satisfied before its horizon.  One pass therefore contains 500 trials per suite and
2,000 across all suites; the reported mean over evaluation seeds 42--44 contains
6,000 trials.
Suite scores average their ten task success rates, and the overall score averages
the four suite scores.

\paragraph{RoboEval.}
We evaluate the Position+Orientation variation of each of RoboEval's eight
bimanual tasks: \emph{Lift Pot} (grasp both handles and
raise the pot), \emph{Lift Tray} (grasp and lift the tray with both arms),
\emph{Pack Box} (close both box flaps), \emph{Pick Single Book from Table},
\emph{Rotate Valve} (rotate both valves counterclockwise), \emph{Stack Single Book
on Shelf}, \emph{Stack Two Blocks}, and \emph{Cube Handover} (transfer a cube
between arms).  We jointly adapt on all eight task datasets, sampling tasks
uniformly.  The reported \RHO{} model uses bimanual 6D end-effector actions, global
batch size 128, eight flow samples per training example, and the checkpoint after
40k optimizer updates.  State and action features use a single set of means and
standard deviations pooled over all eight task datasets (89,765 frames); the same
statistics are shared by every task during training and evaluation.

A RoboEval rollout succeeds when the environment's binary task-success predicate is
met within 250 environment steps.  The policy uses an action-chunk size of 32 and a
16-action execution horizon.  Each evaluation pass contains 100 trials per task
(800 total), and we report the mean over three complete passes: 300 trials per task
and 2,400 trials per model.  Task progression and the behavioral
measures in \Cref{tab:roboeval-behavioral,app:roboeval-behavioral} are computed from
the same rollouts; they do not alter the binary success definition.

\paragraph{RoboTwin 2.0.}

The RoboTwin 2.0 study of midtraining effectiveness in ~\Cref{sec:robotwin-midtraining} entails midtraining \RHO for one of RoboTwin 2.0's simulated robot embodiments using a multitask set of demonstrations provided by RoboTwin, finetuning the pretrained-only and midtrained \RHO variants on RoboTwin tasks whose data is excluded from the midtraining mixture, and rolling out the resulting policies on these held-out tasks.

We choose the dual-arm UR5 robot with WSG grippers as our target RoboTwin 2.0 embodiment. RoboTwin provides demonstration data for 50 tasks. We use 45 tasks for midtraining and hold out 5 tasks for finetuning and evaluation: \emph{Hanging Mug}, \emph{Move Stapler to Pad},
\emph{Place Burger and Fries on Tray}, \emph{Microphone Handover}, and \emph{Stack Three Bowls}.

For midtraining, we merge the demonstration data for 45 tasks, yielding roughly 23,000 demonstrations. We run midtraining for 2 epochs of 40K updates each, with a global batch size of 128 and 8 flow samples per training example.

For finetuning, the combined 5-task demonstration dataset contains 2,750 demonstrations. 
For measuring the effect of midtraining on finetuning datasets of different sizes, we carve out 12.5\%, 25\%, 50\%, and
100\% subsets of this combined dataset, where the 50\% dataset contains the 25\% dataset, and the 25\% dataset contains the 12.5\% dataset. We finetune the midtrained and pretrained-only \RHO variants on each of these datasets, and compare their performance on 3 seeds.

Each of the reported Rho models uses bimanual 6D end-effector actions.
Actions and states use a single set of mean and standard-deviation normalization statistics pooled over
the 100\% finetuning dataset and shared across all tasks during training and evaluation.

For each of the finetuning datasets and for each of the 3 seeds, we do one evaluation pass for each of the finetuned models. Each evaluation pass contains 50 trials per task, or 250 trials total; we measure the success rate of each pass and report the success rates averaged over 3 seeds. A RoboTwin 2.0 rollout succeeds when the environment’s native binary task-success predicate
is met within 300 policy steps. Trials are matched across models: episode (i) for a given task uses scene and
instruction seed (S+i), ensuring that checkpoints evaluated with the same base seed
encounter identical scenarios. The policy predicts chunks of 32 actions and executes 16
actions before replanning. Each evaluation pass contains 50 trials per task, or 250 trials
total.

RoboTwin 2.0 has two evaluation modes, Easy and Hard, and we report the results for each of them. In the Easy mode, the environment has the same fixed backgrounds, lighting, and table height as in the finetuning datasets, with no additional clutter. The Hard protocol randomizes backgrounds
and lighting, adds table clutter, and varies table height by up to 3 cm. Hard-protocol results are averaged over base seeds 10000, 11000, and 13000 (these seeds yielded valid scenarios, unlike many others). Easy-protocol results use seeds 42-45.

\section{Physical-robot evaluation protocols}
\label{app:real-eval}

We use a common evaluation protocol across \fr3d, \urait, and \yb, the parallel A/B evaluation~\cite{kressgazit2024robotlearningempiricalscience, fortier2026busybox}. The evaluation proceeds in ``blocks'', where every block consists in picking an initial state of the environment and sequentially running one trial for each model being compared starting from that initial state. Before each
trial, an operator resets the robot and all manipulated objects to the initial state chosen for that block. Each initial state specifies
the robot's starting pose and the initial object arrangement. The evaluation is run for $N$ initial states/blocks, and we report the success rate of each model across the $N$ blocks. In essence, parallel A/B evaluation reduces the variance of physical-robot experiments by ensuring that each model is evaluated from the same set of initial conditions and that, for each initial condition, all models are evaluated on it at roughly the same time, minimizing environment drift.

Across initial states/blocks, we vary object position, orientation, and, where the task permits,
object identity or arrangement. Camera poses, controller settings, language instructions,
and the physical workspace remain fixed within each comparison.

Each rollout begins after the reset is complete and runs autonomously until the task-specific time
horizon, measured by the number of action execution steps, is reached or until an unrecoverable failure occurs.  A trial is successful if and only if the robot completes every
required task stage and all relevant objects are in their task-specified terminal
configuration without human assistance when the task horizon is reached. This means that, e.g., an insertion must remain
seated, a stacked object must remain stable, and articulated BusyBox controls
must be at their target state at the time horizon, not just at some earlier point. Dropping a required object outside the reachable
workspace, reaching an unrecoverable configuration, or triggering a safety stop
counts as failure.  

Human involvement during evaluation is limited to starting a trial, resetting
the robot and scene between trials, and supervising safety; no teleoperation or
corrective intervention is allowed during a scored rollout.  An operator remains
within reach of the emergency stop throughout execution.  Low-level controllers
enforce joint, reachability, action, and gripper limits, and communication
watchdogs stop motion if commands are interrupted.  The operator terminates a
rollout if continued motion could damage the robot, task apparatus, or nearby
objects.  A trial interrupted by policy behavior is scored as a failure; a trial
invalidated before meaningful policy execution by an unrelated camera,
communication, or hardware fault is reset and repeated.

For multi-stage tasks, task progress is the fraction of
predefined stages completed; it is reported as a diagnostic measure and does not
change the binary success label.

The experiments on \urait (\Cref{sec:urait-exps}) and \fr3d (\Cref{sec:fr3d-exps}) are done on multi-stage tasks. We finetune \RHOUAT and \RHOFR on each task's data separately and use $N = 30$ evaluation blocks per task, i.e., each model's success rate on each task in these experiments is computed over 30 rollouts. The \yb experiments (\Cref{sec:yb-exps}) use single-stage BusyBox tasks. We pool their \yb demonstration data, available at \url{https://huggingface.co/datasets/microsoft/BusyBox}, finetune a single policy for all BusyBox task classes from each base model, and carry out $N=60$ evaluation blocks, 10 blocks for each task class. The overall scores on the right of Figures~\ref{fig:yam-box}, \ref{fig:ur-ai-trainer}, and \ref{fig:fr3-duo} weight task variants equally.

The online-adaptation evaluation uses a separate hard-initialization protocol,
described in \Cref{sec:online-results,app:online}.

\section{Target embodiment descriptions}
\label{app:robots}

\subsection{\yb}
\label{sec:yb-descr}

The \yb is a portable bimanual platform containing two six-degree-of-freedom
I$^2$RT YAM follower arms with motorized parallel-jaw grippers.  We use three
Intel RealSense D405 cameras: one fixed scene camera and one wrist camera per
arm.  The arms and cameras are calibrated into a common frame, and the two arm
bases in our setup are separated by approximately 589\,mm.  \RHO{} receives
the three RGB views and the two end-effector poses and predicts bimanual
Cartesian targets with position, 6D orientation, and absolute gripper commands.
These targets are converted by the YAM control stack to joint-position commands
and dispatched at 30\,Hz.  The workspace is the shared tabletop directly in
front of the portable base; the adjustable arm spacing produces a central
overlap in which either arm can reach the BusyBox and in which the box can be
rotated or translated during the repositioning tasks.

\subsection{\urait}
\label{sec:urait-descr}

Our Universal Robots AI Trainer instance has two UR5e follower arms with Robotiq 2F-85
parallel-jaw grippers.  The policy observes three Orbbec RGB streams: one fixed
high scene view and one wrist-mounted view per arm.  Cameras acquire at
60\,Hz, and the latest synchronized images are consumed by the robot's 18-Hz 
control loop.  Both arms' poses are transformed from their tilted base frames
into a common world frame.  The policy commands bimanual Cartesian
end-effector poses and absolute gripper positions in this frame; the deployment
controller maps each target into the appropriate arm frame and tracks it with
Cartesian impedance control.  The physical workspace is the shared tabletop
in front of the two inward-facing arms.  Objects can be placed in either
arm's reachable region or in the central overlap used for coordinated lifting,
packing, and handoff behaviors.

The physical midtraining data is augmented with successful trajectories from
an OmniReset simulation of \urait in NVIDIA Isaac Sim.  The
simulation runs at 50 Hz with absolute Cartesian operational-space control and
uses the same three-view policy interface as the physical platform.  It
contributes cube stacking, left- and right-arm placement, and left- and
right-arm insertion tasks.  We randomize object colors, camera poses, and focal
lengths during collection, as described in \Cref{sec:midtraining}.

\subsection{\fr3d}
\label{sec:fr3d-descr}

The \fr3d platform comprises two seven-degree-of-freedom Franka Research~3
arms fitted with Robotiq parallel-jaw grippers.  Visual observations come from
one fixed ZED scene camera and one Intel RealSense camera mounted at each
wrist.  All three RGB streams are captured at 30\,Hz.  The two arms are
expressed in a common world frame centered on the dual-arm mount.  \RHO{}
predicts bimanual Cartesian end-effector targets---position, a continuous 6D
orientation representation, and an absolute gripper command for each arm---at
30\,Hz.  At deployment, orientations are converted to quaternions and the pose
targets are executed by the platform's Cartesian impedance controllers; target
poses that would violate joint limits are projected back to reachable poses.
The task workspace is the shared tabletop region in front of the mount,
including the overlap of the two arms' reachable sets for bimanual handoffs,
assembly, and object manipulation.  Unimanual tasks use the corresponding
arm's portion of the same table and omit the unused wrist-camera stream.

\section{Online-adaptation details}
\label{app:online}

\paragraph{Latent-space adaptation.}
Let $G_\theta(o,z)$ denote \RHO's flow solver, which maps an observation $o$ and latent initial condition $z$ to an action chunk $a$. Ordinarily, $z$ is sampled from a Gaussian distribution. For online adaptation, we enable \RHO's optional internal latent policy $\pi_\phi(o)=z$, implemented as a learned noise-sampler submodule of the flow model. The resulting policy executes
\begin{equation}
    a = G_\theta\!\left(o,\pi_\phi(o)\right).
\end{equation}
We freeze all pretrained parameters $\theta$ and train only $\phi$. Because the latent policy is a member of \RHO's flow model, it is trained in place, saved in the normal model state, and served as part of a single \RHO\ checkpoint rather than as an external policy wrapper.

For an executed action chunk $a^\star$, we invert the frozen flow to recover a corresponding latent target,
\begin{equation}
    z^\star \approx G_\theta^{-1}(o,a^\star).
\end{equation}
The latent policy is then trained by supervised regression,
\begin{equation}
    \mathcal{L}_{\mathrm{latent}}(\phi)
    =
    \mathbb{E}_{(o,z^\star)\sim\mathcal{B}}
    \left[\left\|\pi_\phi(o)-z^\star\right\|_2^2\right].
\end{equation}
Only the internal latent-policy parameters are updated; \RHO's vision-language backbone and flow-matching action expert remain frozen throughout online adaptation.

\paragraph{Noise inversion and refinement.}
We invert the same discretized flow used for action generation. Starting from the executed action chunk, we traverse the integration grid in reverse. Each inverse Euler step is implicit because the velocity must be evaluated at the unknown state. We solve this equation using fixed-point iteration: an explicit reverse-Euler estimate initializes the step, after which the velocity field is repeatedly reevaluated at the current estimate. We use five fixed-point iterations per solver step in Meta-World and three on \fr3d. We invert the complete chunk executed during a query period, including both autonomous and supervisor-controlled portions when a takeover occurs. Because the flow need not be injective, this procedure recovers a preimage that reproduces the executed chunk rather than the unique noise originally sampled. We decode each recovered target through the forward flow and discard it if its round-trip mean-squared error exceeds $10^{-3}$. Thus, refinement refers to the per-step fixed-point solution of the inverse trajectory; it does not update \RHO's pretrained parameters.

\paragraph{Internal latent policy.}
The latent policy reuses \RHO's own observation representation rather than introducing a separate visual encoder. It masked-mean-pools the vision-language prefix features, concatenates the current proprioceptive state, and applies a three-layer MLP with 1,024 hidden units per layer, layer normalization, and $\tanh$ activations. We calibrate the per-dimension mean and standard deviation of the pooled prefix features from the initial expert episodes before optimization. The MLP outputs a $16\!\times\!32$ latent initial condition, bounded by a final $\tanh$ to magnitude 3. We optimize the mean-squared latent-regression loss with Adam.

\paragraph{Correction interfaces.}
In Meta-World, a privileged scripted expert takes over at a randomly selected point in an episode and remains in control through its end. The policy is queried every eight environment steps and is adapted for 50 interaction episodes per task. On \fr3d, a human operator initiates a takeover when correction is needed and supplies end-effector corrections with a SpaceMouse. We collect 15 supervised episodes per task. In both cases, the actions actually executed over each query period are assembled into a complete chunk and inverted to produce the latent training target. Training and evaluation are run separately, with a fixed checkpoint used for evaluation rollouts.

\paragraph{Safety.}
The operator retains takeover and emergency-stop authority throughout physical data collection. The learned policy can change only the bounded latent initial condition supplied to the frozen action generator. Generated actions remain subject to the robot controller's action, joint, and workspace limits and communication watchdogs. We save the adapted model through \RHO's normal checkpoint path after optimization and use a fixed checkpoint for evaluation.

\begin{table}[H]
  \centering
  \small
  \setlength{\tabcolsep}{5pt}
  \begin{tabular}{@{}lcc@{}}
    \toprule
    Hyperparameter & Meta-World & \fr3d \\
    \midrule
    Correction source & Scripted expert & Human (SpaceMouse) \\
    Online episodes per task & 50 & 15 \\
    Action/latent horizon & 16 & 16 \\
    Latent dimension per step & 32 & 32 \\
    Latent bound & 3.0 & 3.0 \\
    Inversion method & Per-step fixed point & Per-step fixed point \\
    Fixed-point iterations & 5 & 3 \\
    Inversion-error threshold & $10^{-3}$ & $10^{-3}$ \\
    Latent-policy input & VLM prefix + state & VLM prefix + state \\
    Latent-policy MLP & $3\!\times\!1{,}024$ & $3\!\times\!1{,}024$ \\
    Optimizer & Adam & Adam \\
    Learning rate & $10^{-4}$ & $5\!\times\!10^{-5}$ \\
    Training batch size & 256 & 256 \\
    Updates per optimization cycle & 100 & 100 \\
    \bottomrule
  \end{tabular}
  \caption{Online latent-space adaptation hyperparameters. \RHO's vision-language backbone and flow-matching action expert are frozen in both settings; only its lightweight internal latent policy is optimized.}
  \label{tab:online-adaptation-hparams}
\end{table}

\section{RoboEval behavioral metrics}
\label{app:roboeval-behavioral}

\Cref{tab:roboeval-per-task-behavioral} gives the task-level breakdown of all RoboEval metrics, making task-specific performance differences visible.

\begin{table}[H]
  \centering
  \scriptsize
  \setlength{\tabcolsep}{3.25pt}
  \begin{tabular}{@{}llrrrrrrrrr@{}}
    \toprule
    Task & Model & Succ. $\uparrow$ & TP $\uparrow$ & BGVD $\downarrow$ & CPL $\downarrow$ & ECC $\downarrow$ & JPL $\downarrow$ & OPL $\downarrow$ & SCC $\downarrow$ & SC $\downarrow$ \\
    \midrule
    \multirow{4}{*}{Cube handover}
      & GR00T N1.7 & 0.64 & 0.93 & 0.043 & 1.35 & 0.16 & 6.81 & 4.14 & 0.373 & 0.477 \\
      & $\pi_{0.5}$ & 0.79 & 0.92 & 0.038 & 1.18 & 0.10 & 5.44 & 3.37 & 0.210 & 0.237 \\
      & MolmoAct2 & 0.86 & 0.95 & 0.036 & 1.11 & 0.04 & 5.12 & 3.30 & 0.107 & 0.193 \\
      & \RHO{} & 0.80 & 0.93 & 0.037 & 1.18 & 0.09 & 5.15 & 3.14 & 0.207 & 0.217 \\
    \addlinespace
    \multirow{4}{*}{Lift pot}
      & GR00T N1.7 & 0.67 & 0.81 & 0.058 & 2.57 & 0.13 & 16.83 & 9.70 & 0.120 & 0.603 \\
      & $\pi_{0.5}$ & 0.72 & 0.84 & 0.029 & 1.48 & 0.00 & 9.39 & 6.80 & 0.017 & 0.417 \\
      & MolmoAct2 & 0.57 & 0.77 & 0.026 & 1.61 & 0.00 & 9.32 & 6.83 & 0.003 & 0.570 \\
      & \RHO{} & 0.71 & 0.85 & 0.027 & 1.45 & 0.00 & 9.06 & 6.60 & 0.000 & 0.390 \\
    \addlinespace
    \multirow{4}{*}{Lift tray}
      & GR00T N1.7 & 0.96 & 0.99 & 0.027 & 0.85 & 0.00 & 5.84 & 4.62 & 0.083 & 0.083 \\
      & $\pi_{0.5}$ & 1.00 & 1.00 & 0.026 & 0.66 & 0.00 & 4.75 & 4.04 & 0.000 & 0.017 \\
      & MolmoAct2 & 0.96 & 0.99 & 0.029 & 0.71 & 0.00 & 4.81 & 3.94 & 0.007 & 0.057 \\
      & \RHO{} & 0.99 & 1.00 & 0.017 & 0.68 & 0.01 & 4.94 & 4.24 & 0.000 & 0.020 \\
    \addlinespace
    \multirow{4}{*}{Pack box}
      & GR00T N1.7 & 0.44 & 0.67 & 0.065 & 2.91 & 1.05 & 14.00 & 8.20 & 0.043 & 1.313 \\
      & $\pi_{0.5}$ & 0.53 & 0.74 & 0.064 & 3.23 & 0.92 & 14.99 & 9.17 & 0.067 & 1.080 \\
      & MolmoAct2 & 0.28 & 0.63 & 0.045 & 2.76 & 1.04 & 12.65 & 8.44 & 0.017 & 1.030 \\
      & \RHO{} & 0.50 & 0.75 & 0.055 & 2.99 & 1.34 & 13.90 & 8.77 & 0.057 & 1.367 \\
    \addlinespace
    \multirow{4}{*}{Pick single book}
      & GR00T N1.7 & 0.63 & 0.98 & 0.202 & 1.79 & 0.72 & 11.66 & 5.13 & 0.030 & 0.537 \\
      & $\pi_{0.5}$ & 0.72 & 0.99 & 0.197 & 1.40 & 0.63 & 8.80 & 4.14 & 0.000 & 0.613 \\
      & MolmoAct2 & 0.69 & 0.98 & 0.213 & 1.54 & 0.47 & 9.49 & 4.25 & 0.000 & 0.527 \\
      & \RHO{} & 0.85 & 1.00 & 0.214 & 1.33 & 0.46 & 8.44 & 3.84 & 0.000 & 0.480 \\
    \addlinespace
    \multirow{4}{*}{Rotate valve}
      & GR00T N1.7 & 0.59 & 0.84 & 0.015 & 0.45 & 6.79 & 3.48 & 2.55 & 0.000 & 0.000 \\
      & $\pi_{0.5}$ & 0.57 & 0.81 & 0.012 & 0.39 & 5.25 & 2.66 & 2.06 & 0.003 & 0.000 \\
      & MolmoAct2 & 0.60 & 0.86 & 0.014 & 0.41 & 5.83 & 2.98 & 2.28 & 0.000 & 0.000 \\
      & \RHO{} & 0.69 & 0.88 & 0.012 & 0.41 & 5.62 & 3.06 & 2.50 & 0.020 & 0.000 \\
    \addlinespace
    \multirow{4}{*}{Stack single book shelf}
      & GR00T N1.7 & 0.37 & 0.61 & 0.165 & 2.20 & 1.03 & 14.06 & 6.36 & 0.013 & 0.767 \\
      & $\pi_{0.5}$ & 0.61 & 0.73 & 0.178 & 2.12 & 0.86 & 13.28 & 6.14 & 0.000 & 0.923 \\
      & MolmoAct2 & 0.42 & 0.61 & 0.161 & 2.42 & 0.63 & 14.70 & 6.56 & 0.000 & 0.797 \\
      & \RHO{} & 0.77 & 0.85 & 0.171 & 2.05 & 0.56 & 12.73 & 5.78 & 0.000 & 0.767 \\
    \addlinespace
    \multirow{4}{*}{Stack two blocks}
      & GR00T N1.7 & 0.56 & 0.61 & 0.052 & 1.17 & 0.15 & 4.47 & 2.70 & 0.093 & 0.543 \\
      & $\pi_{0.5}$ & 0.43 & 0.53 & 0.049 & 1.17 & 0.18 & 4.55 & 2.83 & 0.003 & 0.567 \\
      & MolmoAct2 & 0.47 & 0.56 & 0.050 & 1.28 & 0.22 & 4.67 & 3.10 & 0.000 & 0.550 \\
      & \RHO{} & 0.55 & 0.59 & 0.054 & 1.16 & 0.06 & 4.28 & 2.63 & 0.000 & 0.577 \\
    \bottomrule
  \end{tabular}

  \vspace{4pt}
  \begin{minipage}{0.98\linewidth}
  \scriptsize
  Succ.: success rate; TP: task progression; BGVD: bimanual gripper vertical difference (m); CPL: Cartesian path length (m); ECC: environment collision count; JPL: joint path length (rad); OPL: orientation path length (rad); SCC: self-collision count; SC: slip count.
  \end{minipage}
  \caption{Task-level RoboEval outcome and behavioral metrics after 40k task-adaptation steps. Each cell is pooled over three 100-episode evaluations for the corresponding task. Path-length measures sum the two arms. Arrows indicate the preferred direction.}
  \label{tab:roboeval-per-task-behavioral}
\end{table}

\section{Authors' contributions}
\label{app:authors}

\textbf{Simran Bagaria} designed and implemented synthetic data generation in NVIDIA Isaac Sim for midtraining \RHO for \urait. She implemented key components of the server and simulation infrastructure, including the integration of RoboEval. She contributed to maintaining the physical \urait\ robot, collecting teleoperated task demonstrations on it, and finetuning \RHO{} for various experiments. \\

\noindent
\textbf{Daphne Chen} assisted with integrating online latent-space adaptation into \RHO and running the online-adaptation experiments. \\

\noindent
\textbf{Dean Fortier} made significant contributions to setting up and maintaining the project’s robot hardware, particularly \fr3d. He also established the initial teleoperation and data-collection pipeline, collected data for various experiments, and fine-tuned baseline models. \\

\noindent
\textbf{Jianlong Fu} worked on WAM-based \RHO variants, which are still undergoing evaluation at the time of writing. \\

\noindent
\textbf{Michael Harrison} assisted with curation and evaluation of training datasets for \RHO's VLM backbone \phiphy relevant to \phiphy's physical grounding. \\

\noindent
\textbf{Tess Hellebrekers} designed and implemented early versions of \RHO's architecture, simulation experiments, and data processing pipelines. She set up data cleaning and conversion for pretraining. She experimented with several methods of incorporating tactile sensing into \RHO that aren't described in this report. She also set up and maintained \urait, including teleoperation and tactile sensing hardware, its software control stack, and controller tuning. \\

\noindent
\textbf{Neel Joshi} led the development and evaluation of \RHO's VLM backbone \phiphy, including its architecture design, data curation, training, and evaluation. \\

{\noindent\looseness=-1 \spaceskip=0.97\fontdimen2\font plus \fontdimen3\font minus \fontdimen4=\spaceskip
\textbf{Andrey Kolobov} was the overall lead of \RHO. He set the project's goals and strategy, including \RHO's midtraining approach focusing on common dual-arm embodiments, oversaw its research and implementation across several teams,  designed the evaluation protocol and the experiments in Sections \ref{sec:midtraining-eval}-\ref{sec:physical-results}, contributed to \RHO's architecture design, collected a significant fraction of the finetuning data for \urait and \yb's experiments, and carried out the final \yb evaluations. He was also \RHO's data lead, defining \RHO's data recipe and mixtures for all training stages, procuring the proprietary data, and implementing and optimizing \RHO's data processing pipelines.\par}\vspace{\baselineskip} 

\noindent
\textbf{Dalton Moore} made major contributions to setting up the software stack for \yb, collected much of teleoperation data on it and on \urait, and finetuned \RHO and baseline models on it. \\

\noindent
\textbf{Galen Mullins} led the engineering of \RHO's software framework and end-to-end MLOps pipeline. He developed and iteratively optimized the distributed training and evaluation infrastructure, engineered data loading pipelines for system scaling, and set up the project's continuous integration and reproducibility tooling. He built the core simulation and server infrastructure, integrating environments such as RoboTwin 2.0 and LIBERO, and the software stack for \yb. He was responsible for the \urait physical experiments and significantly contributed to those on \yb, including model finetuning, implementing evaluation protocols, and collecting finetuning data.\\

\noindent
\textbf{Michael Murray} led the design of \RHO's training recipe and architecture, including implementing shared AdaLN modulation, grouped-query attention, VL cotraining, knowledge insulation, and online latent-space adaptation. He designed and ran experiments to determine the role of these features in \RHO's final action-expert design and the values of pretraining and finetuning hyperparameters. He ran \RHO's pretraining and set up its midtraining for \fr3d and \urait. He built the software stack for \fr3d, collected data for offline finetuning experiments on \fr3d, and ran them. He also ran online-adaptation experiments on \fr3d and in Meta-World. \\ 

\noindent
\textbf{Eduardo Salinas} contributed to engineering \phiphy's training pipeline and inference. \\

\noindent
\textbf{Reuben Tan} consulted the project team on the design of \RHO training recipe and data augmentation strategies, and helped debug the \RHO training pipeline. \\